\pdfoutput=1
\documentclass{article} %
\usepackage{iclr2026_conference,times}

\usepackage{amsmath,amsfonts,bm}

\def\eqref#1{equation~\ref{#1}}

\def\1{\bm{1}}

\DeclareMathAlphabet{\mathsfit}{\encodingdefault}{\sfdefault}{m}{sl}
\SetMathAlphabet{\mathsfit}{bold}{\encodingdefault}{\sfdefault}{bx}{n}

\usepackage[hyperfootnotes=false]{hyperref}
\usepackage{url}
\usepackage{times}
\usepackage{microtype}
\usepackage{epsfig}
\usepackage[skip=5pt]{caption}
\usepackage{float}
\usepackage{placeins}
\usepackage{color, colortbl}
\usepackage{stfloats}
\usepackage{enumitem}
\usepackage{tabularx}
\usepackage{xstring}
\usepackage{cuted} %
\usepackage{multirow}
\usepackage{xspace}
\usepackage{url}
\usepackage{mathtools}
\usepackage[hang,flushmargin]{footmisc}
\usepackage[normalem]{ulem}  %
\usepackage{graphicx}
\usepackage{amsmath}
\usepackage{amssymb}
\usepackage{comment}
\usepackage[title]{appendix}
\usepackage{cancel}
\usepackage{subcaption}
\usepackage{afterpage}
\usepackage{algorithm}
\usepackage{algorithmic}
\usepackage{bm}
\usepackage{wrapfig}

\newcommand*{\rom}[1]{\expandafter\@slowromancap\romannumeral #1@}

\newfloat{sepfigure}{p}{lop}
\floatname{sepfigure}{Figure}

\usepackage{algorithm}
\usepackage{algorithmic}

\usepackage{bm}
\renewcommand{\vec}[1]{\bm{#1}}

\AtEndPreamble{
    \usepackage[capitalize]{cleveref}
    \crefname{section}{Sec.}{Secs.}
    \Crefname{section}{Section}{Sections}
    \Crefname{table}{Table}{Tables}
    \crefname{table}{Tab.}{Tabs.}
}
\usepackage{amsmath}
\usepackage{amssymb}
\usepackage{booktabs}

\hypersetup{
  colorlinks=true,
  linkcolor=red,     
  urlcolor=magenta,
  citecolor=teal!80!black
}

\title{Target-Aware Video Diffusion Models}

\author{Taeksoo Kim \\
Seoul National University\\
\texttt{taeksu98@snu.ac.kr}
\And
Hanbyul Joo\setcounter{footnote}{1}\thanks{Corresponding author}\\
Seoul National University \& RLWRLD \\
\texttt{hbjoo@snu.ac.kr}
}

\iclrfinalcopy %
\begin{document}

\maketitle
\begingroup
\renewcommand\thefootnote{}%
\footnotetext{\hspace*{1.8em}Project page: \url{https://taeksuu.github.io/tavid/}}%
\addtocounter{footnote}{-1}%
\endgroup

\begin{abstract}
We present a target-aware video diffusion model that generates videos from an input image, in which an actor interacts with a specified target while performing a desired action. The target is defined by a segmentation mask, and the action is described through a text prompt.
Our key motivation is to incorporate target awareness into video generation, enabling actors to perform directed actions on designated objects. This enables video diffusion models to act as motion planners, producing plausible predictions of human-object interactions by leveraging the priors of large-scale video generative models.
We build our target-aware model by extending a baseline model to incorporate the target mask as an additional input. To enforce target awareness, we introduce a special token that encodes the target's spatial information within the text prompt. We then fine-tune the model with our curated dataset using an additional cross-attention loss that aligns the cross-attention maps associated with this token with the input target mask. To further improve performance, we selectively apply this loss to the most semantically relevant attention regions and transformer blocks.
Experimental results show that our target-aware model outperforms existing solutions in generating videos where actors interact accurately with the specified targets. We further demonstrate its efficacy in two downstream applications: zero-shot 3D HOI motion synthesis with physical plausibility and long-term video content creation.

\end{abstract}

\section{Introduction}
\label{sec:intro}

Video diffusion models have demonstrated remarkable capabilities in simulating complex real-world scenes.
Ideally, such models can serve as motion planners, in line with the concept of world models~\citep{ha2018world, bar2024navigationworldmodels, kim2025dwm}, by producing plausible predictions of interactions between an actor (human or robot) and target objects using priors learned from large-scale video datasets.
However, existing image-to-video diffusion models~\citep{yang2024cogvideox, kong2024hunyuanvideo, HaCohen2024LTXVideo}, which generate videos from an input image guided by text prompts, are target-unaware.
An alternative line of work attempts to explicitly control actor-target interactions using dense structural or motion cues, such as depth maps~\citep{esser2023structure, zhang2023controlvideo}, edges~\citep{chen2023controlavideo, khachatryan2023text2video}, optical flow~\citep{ni2023conditional, burgert2025gowiththeflowmotioncontrollablevideodiffusion, gu2025diffusion}, motion trajectories~\citep{yan2023motion, shi2024motion}, or drag-based manipulation~\citep{deng2023dragvideo, teng2023drag, shi2024dragdiffusion}.
While effective for certain tasks, these approaches do not meet our needs. Our goal is to use video diffusion models to infer plausible actor-target interactions, where action guidance for the actor is not available in advance and thus cannot be provided as input. Ultimately, we aim to leverage video generative models for high-level action planning, inferring realistic interaction cues for the actor within the current scene, as explored in recent robotics research~\citep{black2023zero, du2023learning, ajay2023compositional, ni2024generate}.

In this paper, we present a target-aware video diffusion model that generates videos from an input image, where an actor performs a desired action directed at a specified target. The target is defined by a segmentation mask, and the action is described with a text prompt.
We use the mask as a means to specify the target object in the scene, which can be obtained with minimal effort (e.g., a single click), or automatically from text input using off-the-shelf tools~\citep{ren2024grounded}, and we show that our method is robust to variations in mask quality. By providing an explicit way to designate the target, our model can serve as an effective motion planner, enabling the actor to perform diverse interactions with the specified object. While our training uses videos with human actors, we also demonstrate that the model generalizes seamlessly to other agents, including animals and robotic hands.

To integrate spatial information of the target mask, we extend a base image-to-video diffusion model~\citep{yang2024cogvideox} to take the mask as an additional input, and fine-tune it on our newly curated dataset. However, simply fine-tuning the model with the extra mask input does not ensure the target awareness of the model. To address this, we introduce a special token, [TGT], into the text prompt to describe the target and enforce an alignment between the [TGT] token's cross-attention maps and the input target mask {by applying a loss on the model's cross-attention. This} cross-attention loss enables the model to associate the [TGT] token with the spatial information of the target, improving the precision of the generated interactions with the target {by injecting spatial grounding into the text-conditioning mechanism of the model}. We selectively apply this loss to specific attention regions and transformer blocks that are most semantically relevant for effective supervision.

Experimental results demonstrate that our target-aware video diffusion model outperforms existing solutions in synthesizing videos where actors precisely engage with designated targets.
To further demonstrate the strength of our target-aware model, we apply our model to two downstream applications: (1) zero-shot 3D human-object interaction (HOI) motion synthesis with physical plausibility, simulating physical agents performing plausible actions in a given environment, and (2) video content creation, generating long-term videos covering navigations and interactions with minimal user input.

Our contributions can be summarized as follows: (1) We present a target-aware video diffusion model that generates videos of interactions between the actor and the target using a segmentation mask and a text prompt; (2) {We propose to utilize a cross-attention loss to enable the base model to effectively incorporate the mask input and achieve target awareness}, and provide a comprehensive analysis of its effects across different parts of the model; (3) We present a newly curated dataset specifically designed to train and evaluate our target-aware model; and (4) We demonstrate two real-world applications of our target-aware model: zero-shot 3D HOI motion synthesis for controlling physical agents and video content creation.

\section{Related Work}
\label{sec:related}

\noindent \textbf{Controllable Video Generation}
Building on early work~\citep{ho2022video, esser2023structure, guo2023animatediff} that extends text-to-image diffusion models to video generation, the community has shown significant interest in producing videos with enhanced controls.
Several methods, inspired by ControlNet~\citep{zhang2023controlnet}, have been adapted for videos, where structural cues, such as depth maps~\citep{esser2023structure, zhang2023controlvideo}, edge information~\citep{chen2023controlavideo, khachatryan2023text2video}, optical flow~\citep{ni2023conditional, gu2025diffusion}, or motion~\citep{yan2023motion, shi2024motion, cha2025durian}, are integrated into the generation process via additional modules to produce structure-consistent outputs.
Another line of research focuses on manipulating the internal representations of diffusion models to achieve controls without extra modules. Attention modulation approaches~\citep{yang2024direct, wu2024motionbooth, jain2024peekaboo} adjust cross-attention maps of predefined regions to steer subject movements. Inversion-based feature injection methods~\citep{liu2023videop2p, vid2vid-zero, jeong2023ground} edit videos in a zero-shot manner by leveraging cross-attention control, allowing for content editing without training.
Other work~\citep{teng2023drag, deng2023dragvideo, wu2024draganything, yin2023dragnuwa} extends drag-based image editing techniques~\citep{pan2023drag, shi2024dragdiffusion, shin2024instantdrag} to video, by either adding an extra drag-embedding module or optimizing video latents to align with drag inputs.
While existing approaches focus on generating videos that faithfully follow the dense input cues, either from source video or user inputs, we aim to extract those motion cues from video diffusion models with minimal extra input, a mask of the target.

\noindent \textbf{Human-Scene Interaction Synthesis.}
Synthesizing natural human motions within a given scene remains challenging, requiring a high-level semantic understanding of human-scene interactions and affordances. Early work primarily focuses on posing a static 3D human in a 3D environment~\citep{kim2014shape2pose, savva2016pigraphs, li2019putting, hassan2019resolving, zhang2020generating, zhang2020place, hassan2021populating, huang2022capturing, Zhao:ECCV:2022} or generating short-term, predefined motions, such as reaching or sitting, given a static object~\citep{starke2019neural, taheri2020grab, zhang2021manipnet, zhang2022couch,  taheri2022goal}. More recent work~\citep{wang2021synthesizing, wang2021scene, hassan2021stochastic, lee2023locomotion, jiang2024autonomous, Cha_2025_CVPR, kim2025david} has extended this to synthesizing human motions in 3D scenes with multiple objects, while other work has explored human motion generation involving dynamic objects~\citep{li2023controllable, xu2023interdiff, ghosh2023imos, li2023object, xu2024interdreamer, jiang2024scaling, xu2025intermimic}. These methods leverage 3D motion-scene paired datasets~\citep{savva2016pigraphs, monszpart2019imapper, hassan2019resolving, yi2024generating, kim2024parahome} or 3D human-object interaction datasets~\citep{taheri2020grab, bhatnagar2022behave, jiang2023full, wang2023physhoi, kim2023ncho, xie2024template, oor} to enable such scene-conditioned motion generation. To address the scarcity of large-scale 3D datasets, some approaches~\citep{han2023chorus, ComA, li2024genzi, kim2024gala} leverage the knowledge of 2D generative or vision language models, yet are limited to static human-scene interactions. In this work, we synthesize 3D HOI motions from the 2D scene, utilizing our target-aware video diffusion model.

\section{Preliminaries: Video Diffusion Models}
\label{sec:preliminaries}
Building on the success of text-to-image latent diffusion models~\citep{rombach2022latent, Flux, esser2024scaling}, recent text-to-video (T2V) diffusion models~\citep{blattmann2023stable, yang2024cogvideox, HaCohen2024LTXVideo} generate videos in a latent space. Given a video $\vec{x}$, an encoder $\mathcal{E}$ maps it to its latent representation $\vec{z}$. During the forward process, Gaussian noise is added to $\vec{z}$ at each timestep $t$, as $\vec{z}_t = \alpha_t \vec{z} + \sigma_t \vec{\epsilon}$, where $\vec{\epsilon} \sim \mathcal{N}(0, \vec{I})$ and $\alpha_t$, $\sigma_t$ are noise scheduling coefficients. In the reverse process, the model is trained to predict the noise added to the video latent, guided by input conditions such as text prompt, by minimizing the following objective:
\begin{equation}
\label{eq:vdm}
\begin{split}
    \mathcal{L}_{\text{VDM}}
    = \mathop{\mathbb{E}} \left[ \lVert \vec{\epsilon} - \vec{\epsilon}_{\theta} (\vec{z}_t; \vec{y}, t) \rVert^2_2 \right],
\end{split}
\end{equation}
where $\vec{y}$ denotes the text prompt. T2V diffusion models can be fine-tuned for image-to-video (I2V) tasks by conditioning the model on an extra input image, allowing the video to start from the given image~\citep{xing2023dynamicrafter, yang2024cogvideox, kong2024hunyuanvideo, HaCohen2024LTXVideo}. In this work, we use CogVideoX~\citep{yang2024cogvideox}, one of the SOTA open-sourced video diffusion models based on diffusion transformers~\citep{peebles2023scalable}.

\section{Target-Aware Video Diffusion Models}
\label{sec:method}

Given an image, a mask of the target, and a text prompt describing an action, our target-aware video diffusion model generates videos where an actor accurately interacts with the specified target. We first extend our base video diffusion model to accept the mask as an additional input (\cref{sec:mask_inject}). To make the model utilize the extra mask information, we augment the text prompt by adding a sentence, ``The person interacts with [TGT] object.'', where the [TGT] token is used to encode the spatial information of the target. We {then apply a cross-attention loss} to align the cross-attention maps of the [TGT] token with the mask input and make our model target-aware (\cref{sec:training}). This loss is selectively applied to specific cross-attention regions and transformer blocks to maximize its effectiveness (\cref{sec:selective}). Finally, we curate a dedicated dataset tailored to train our target-aware model (\cref{sec:dataset}).

\subsection{Specifying the Target with a Mask}
\label{sec:mask_inject}

\begin{figure}[t]
\centering
\vspace{-5pt}
\begin{subfigure}[t]{0.46\linewidth}
    \centering
    \includegraphics[width=\linewidth, trim={0cm 0cm 0cm 0cm},clip]{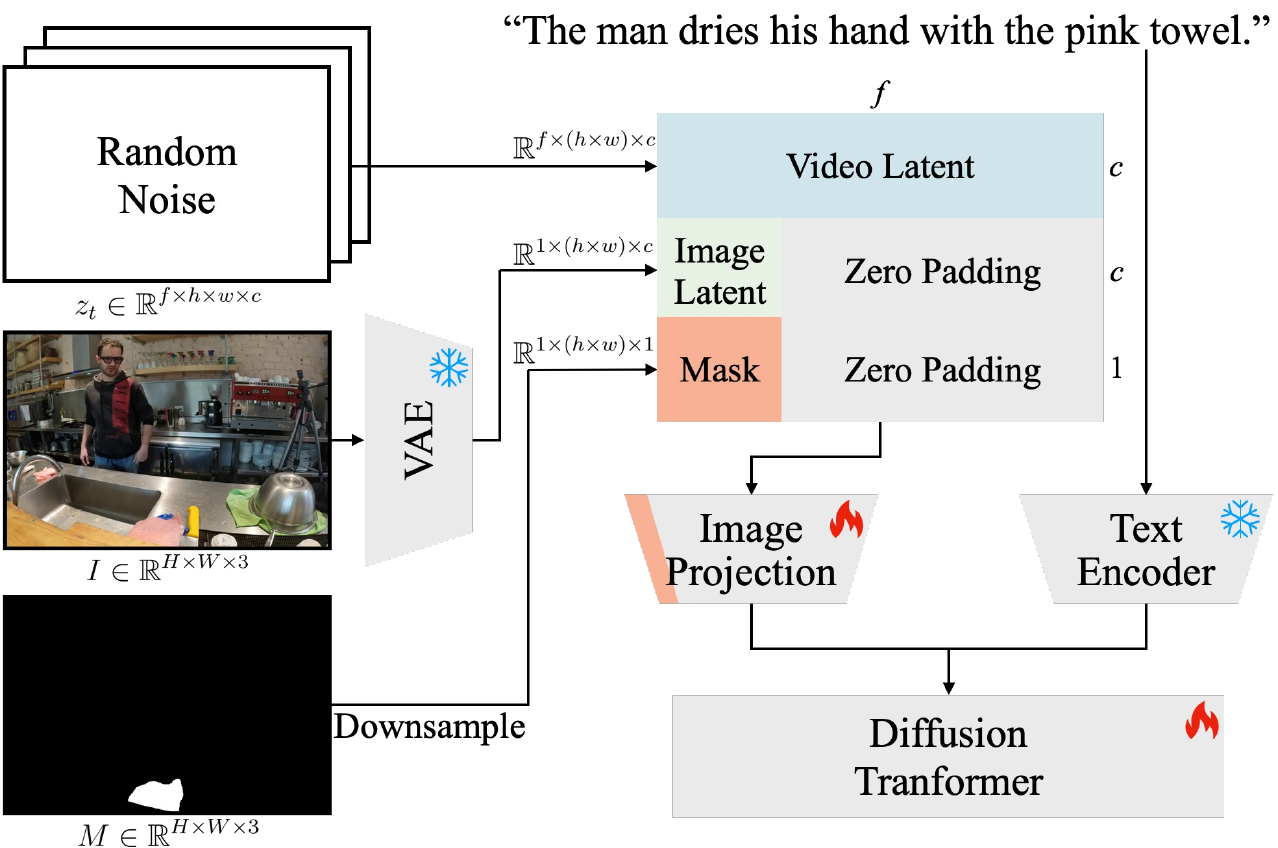}
    \vspace{-15pt}
    \caption{{Injecting the extra mask condition}}
    \label{fig:arch}
\end{subfigure}\hfill
\begin{subfigure}[t]{0.50\linewidth}
    \centering
    \includegraphics[width=\linewidth, trim={0.8cm 0cm 0.8cm 0.8cm},clip]{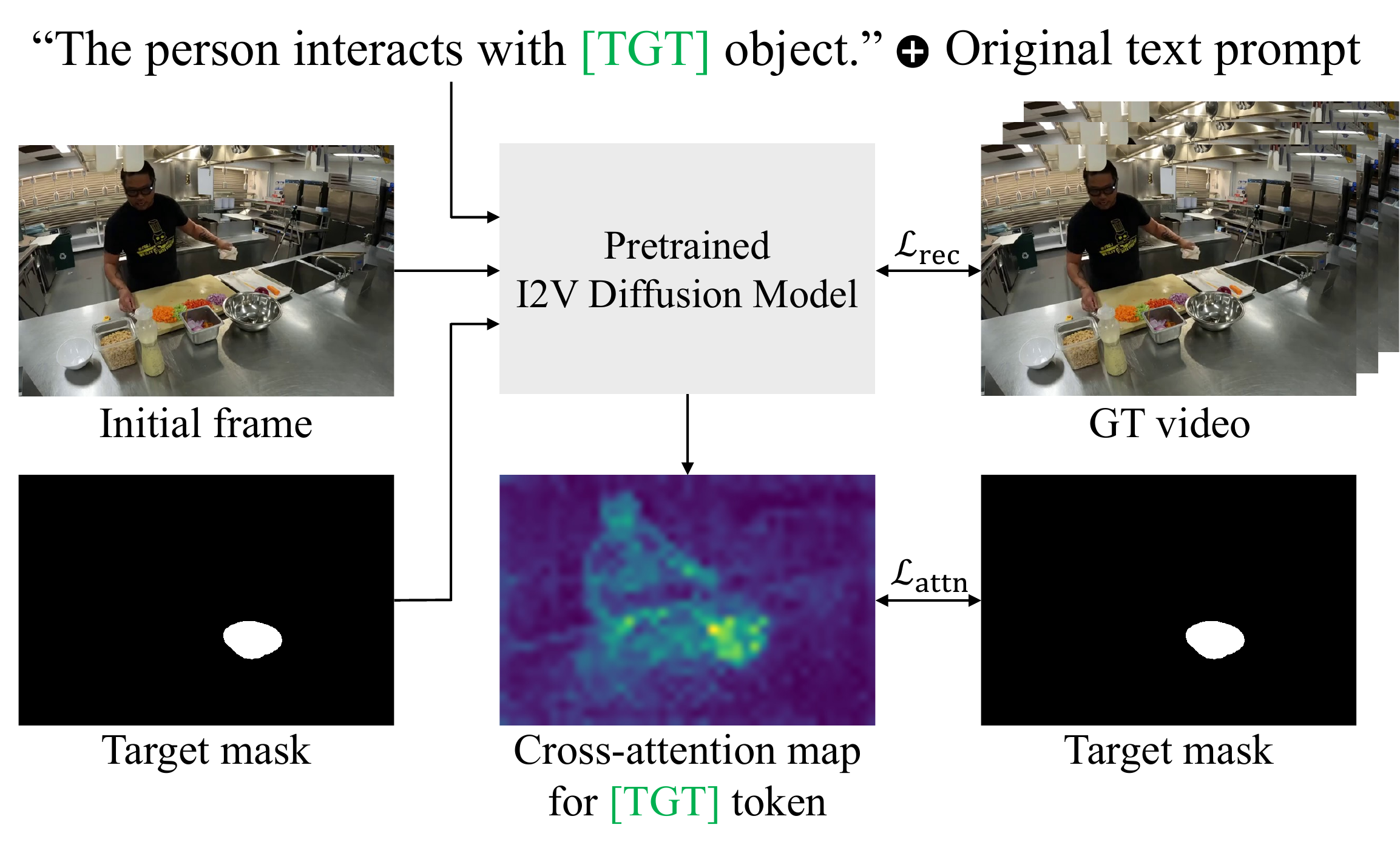}
    \vspace{-15pt}
    \caption{{Target awareness via cross-attention loss}}
    \label{fig:tavdm}
\end{subfigure}

\caption{\textbf{Target-aware video diffusion models.} 
(a) We condition the noisy video latent with a segmentation mask of the target to incorporate spatial information during generation.
(b) We fine-tune the pretrained video diffusion model to utilize the mask input via additional cross-attention loss.}
\label{fig:merged}
\vspace{-15pt}
\end{figure}

To encode the spatial information of the target, we extend our base I2V diffusion model~\citep{yang2024cogvideox} to incorporate a binary segmentation mask of the target. Our base model takes an input image $\vec{I} \in \mathbb{R}^{H\times W\times 3}$ and generates an output video $\vec{V} \in \mathbb{R}^{F\times H\times W\times 3}$, where $F$ denotes the number of frames. To enforce the input image as the first frame of the video, the latent noise $\vec{z}_t \in \mathbb{R}^{f\times h\times w\times c}$ is concatenated channel-wise with the latent encoding of the input image $\mathcal{E}(\vec{I}) \in \mathbb{R}^{1\times h\times w\times c}$ for the first frame, with zero-padding applied for the remaining frames. The concatenated representation is then projected through an image projection layer to align with the text embedding dimension.

To integrate the target mask $\vec{M} \in \mathbb{R}^{H\times W\times 1}$ into this process, we downsample it to $\vec{\tilde{M}} \in \mathbb{R}^{h\times w\times 1}$ and concatenate it alongside the input image condition, again applying zero-padding for the other frames. To support the extra mask channel, we extend the original image projection layer by adding an input channel, and initialize the new weights to zero while preserving the pretrained parameters, following InstructPix2Pix~\citep{brooks2022instructpix2pix}. The overall pipeline is shown in ~\cref{fig:arch}.

\subsection{Target Awareness via Cross-Attention Loss}
\label{sec:training}

We make our model target-aware {by applying a} cross-attention loss that aligns the model's attention on the target with the additional input mask during fine-tuning. For every text prompt in our training dataset, we append a general sentence ``The person interacts with [TGT] object.'', where the [TGT] token is intended to encode the target's spatial information. We then encourage the cross-attention weights between the latent noise corresponding to the first frame of the video and the [TGT] token to align with the provided target mask $\vec{M}$ as demonstrated in ~\cref{fig:tavdm}. Specifically, we minimize the following loss:
\begin{equation}
\label{eq:attn}
\begin{split}
    \mathcal{L}_{\text{attn}}
    = \mathop{\mathbb{E}} \left[ \lVert A(\vec{z}_t^0, \text{[TGT]}) - \vec{\tilde{M}} \rVert^2_2 \right],
\end{split}
\end{equation}
where $A(\vec{z}_t^0, \text{[TGT]})$ denotes the cross-attention weights between the latent noise for the first frame of the video and the [TGT] token. In addition to {the} cross-attention loss, we employ the standard diffusion objective:
\begin{equation}
\label{eq:recon}
\begin{split}
    \mathcal{L}_{\text{rec}}
    = \mathop{\mathbb{E}} \left[ \lVert \vec{\epsilon} - \vec{\epsilon}_{\theta} (\vec{z}_t; \vec{y}, \vec{I}, \vec{\tilde{M}}, t) \rVert^2_2 \right],
\end{split}
\end{equation}
where notations remain consistent with those used in ~\cref{eq:vdm}. Our overall objective is defined as:
\begin{equation}
\label{eq:total}
\begin{split}
    \mathcal{L}_{\text{total}}
    = \mathcal{L}_{\text{rec}} + \lambda_{attn} \mathcal{L}_{\text{attn}},
\end{split}
\end{equation}
where $\lambda_{attn}$ balances the two loss terms. During inference, we prepend the [TGT] token to words referring to the target, enabling the model to leverage the spatial cue provided by the segmentation mask. As presented in ~\cref{fig:attn_loss}, {the} cross-attention loss effectively guides the [TGT] token to focus on the target region, enabling precise interactions with it.

\subsection{Selective Cross-Attention Loss}
\label{sec:selective}
For effective and efficient supervision, we selectively apply the cross-attention loss to the model by identifying (1) the cross-attention regions that most influence target awareness of the model and (2) the transformer blocks that best capture semantics. We validate these design choices via ablations.

\begin{figure}[t]
\centering
\vspace{-15pt}
\begin{subfigure}[t]{0.49\linewidth}
    \centering
    \includegraphics[width=\linewidth, trim={0cm 0cm 1cm 0cm}, clip]{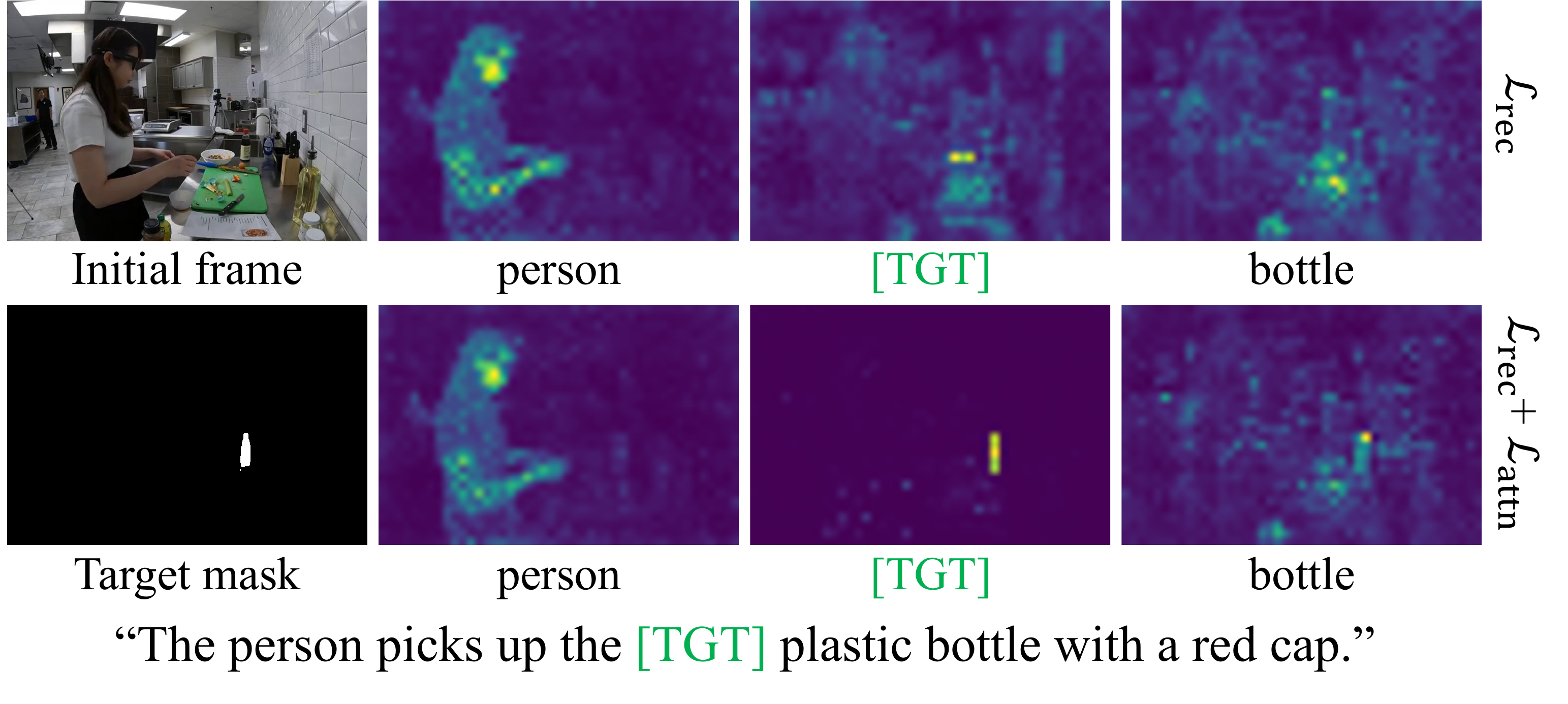}
    \vspace{-15pt}
    \caption{{Effect of the cross-attention loss}}
    \label{fig:attn_loss}
\end{subfigure}\hfill
\begin{subfigure}[t]{0.47\linewidth}
    \centering
    \includegraphics[width=\linewidth, trim={0cm 0cm 0cm 0cm}, clip]{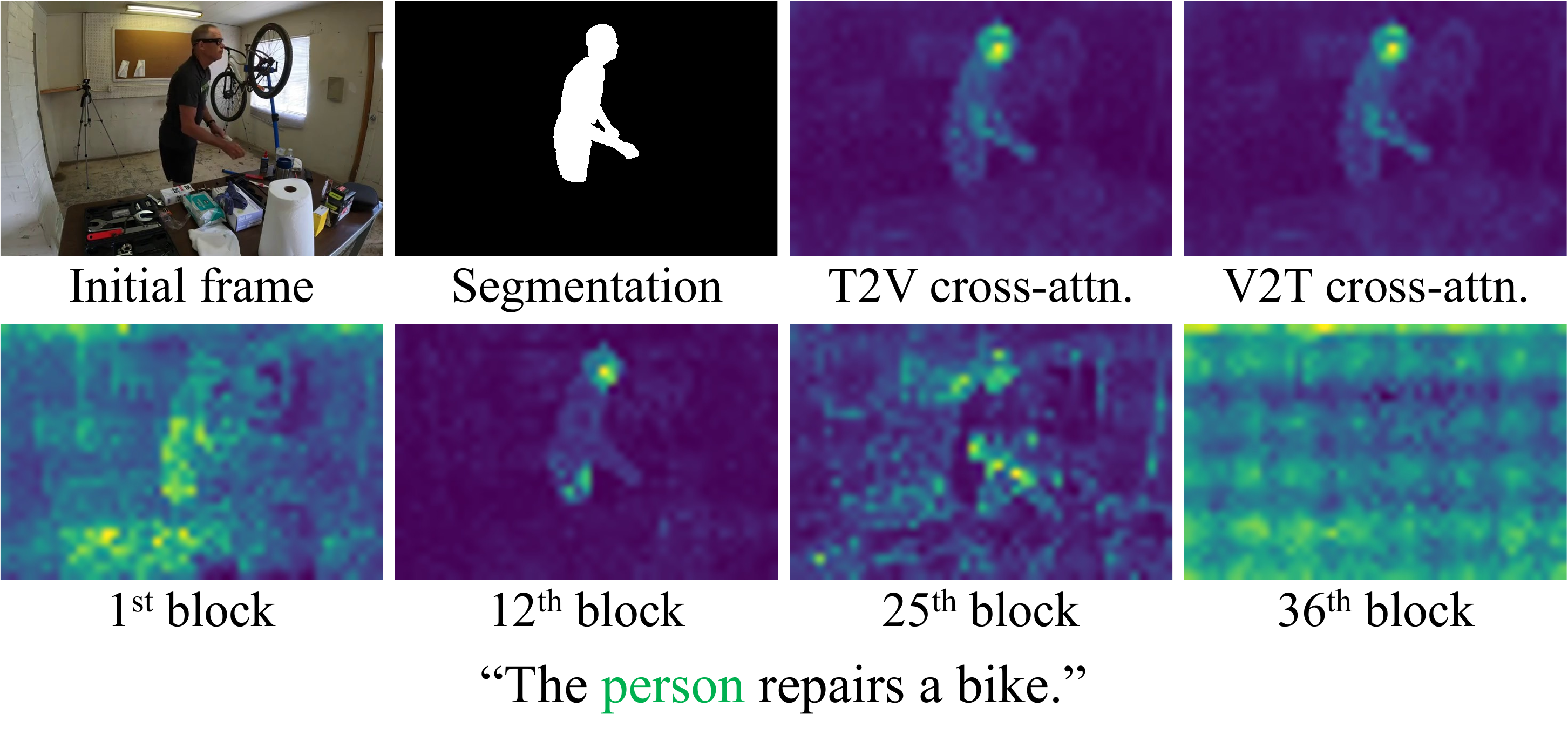}
    \vspace{-15pt}
    \caption{{Selective cross-attention loss}}
    \label{fig:layers}
\end{subfigure}

\caption{\textbf{Cross-attention visualization.}
(a) The cross-attention loss successfully guides the model to focus on the target region. 
(b) We apply the loss on transformer blocks and cross-attention areas that largely impact target awareness of the model.}
\label{fig:attn_loss_layers}
\vspace{-12pt}
\end{figure}

\noindent \textbf{Selective Cross-Attention Regions.}
The multi-modal diffusion transformer architecture, employed by state-of-the-art image and video diffusion models~\citep{Flux, esser2024scaling, kong2024hunyuanvideo}, including our base model~\citep{yang2024cogvideox}, concatenates text and video embeddings into a unified sequence and computes attention over the combined representation. This process yields four distinct attention regions: text-to-text self-attention, text-to-video (T2V) cross-attention, video-to-text (V2T) cross-attention, and video-to-video self-attention. While both T2V and V2T cross-attention maps encode semantic information (\cref{fig:layers} top row), we apply {the} cross-attention loss on V2T cross-attention regions to maximize its impact since V2T cross-attention directly influences the video latent representations during the dot product computation of the attention weights and value features. In contrast, T2V cross-attention primarily affects the text latents, offering less direct influence on the video content. Details on attention mechanisms are provided in Appendix~\cref{sec:attention_supp}.

\noindent \textbf{Selective Transformer Blocks.}
Motivated by prior work~\citep{hertz2022prompt}, we observe that certain transformer blocks capture richer semantic details than others, as shown in the bottom row of ~\cref{fig:layers}. We therefore apply {the} cross-attention loss to those blocks whose cross-attention maps closely resemble the segmentation masks of the corresponding token. To empirically identify these blocks, we evaluate the semantic alignment of each transformer block by first generating 100 videos from a subset of our training images. We then compute the mean squared error between each block's cross-attention map and the segmentation mask for a predefined token.
Our analysis shows that blocks 5 through 23 of the base model~\citep{yang2024cogvideox} yield the smallest errors, and we consequently apply {the} loss to uniformly sampled blocks (every $5^{th}$) within this range, constrained by GPU VRAM limitations.

\subsection{Dataset Curation for Training the Target-Aware Model}
\label{sec:dataset}
To train our target-aware model, we require videos that satisfy two conditions: (1) the initial frame should depict a scene where an actor is present but not yet interacting with the target, and (2) subsequent frames must capture the actor engaging with the target. Directly collecting such data through one's own captures is infeasible, since it would not provide sufficient scale or diversity.
To this end, we curate a dedicated dataset for target-aware video generation, designed to cover diverse interaction scenarios and align with our training pipeline. Each video, sourced from BEHAVE~\citep{bhatnagar2022behave} and Ego-Exo4D~\citep{grauman2024ego} datasets, is annotated with a segmentation mask of the target in the initial frame and paired with text prompts describing the action.
BEHAVE dataset features videos where a single person interacts with a clearly defined target object in a relatively simple setting, whereas Ego-Exo4D contains more complex scenarios, such as cooking or bike repairing, where multiple objects, including those of the same type, may be present. In total, we extract 1290 clips that meet our criteria.
We obtain the mask for the target object in the initial frame using an off-the-shelf segmentation model~\citep{kirillov2023sam} and generate text prompts with CogVLM2-Caption~\citep{yang2024cogvideox}, the same captioning tool used for training our base model. While it is ideal to prepend [TGT] tokens to the target object nouns during caption generation, we find that current video captioning tools cannot reliably identify the target object in complex scenes. Therefore, we add a general sentence, ``The person interacts with [TGT] object." to the generated captions as described in ~\cref{sec:training}, which we find sufficient to train our target-aware model.

\section{Practical Applications of Our Target-Aware Model}

The core strength of our target-aware model lies in its ability to generate plausible and diverse interaction motions between actors and specified target objects, without requiring additional guidance. Leveraging this capability, we present two practical pipelines: (1) zero-shot 3D HOI motion synthesis for a target object with physical plausibility and (2) long-term video content creation with minimal user input. See the detailed pipelines in Appendix~\cref{sec:app_supp}.

\begin{figure}[t]
    \vspace{-15pt}
    \centering
    \begin{minipage}{0.476\linewidth}
        \centering
        \includegraphics[width=\linewidth, trim={0cm 0cm 0cm 0cm}]{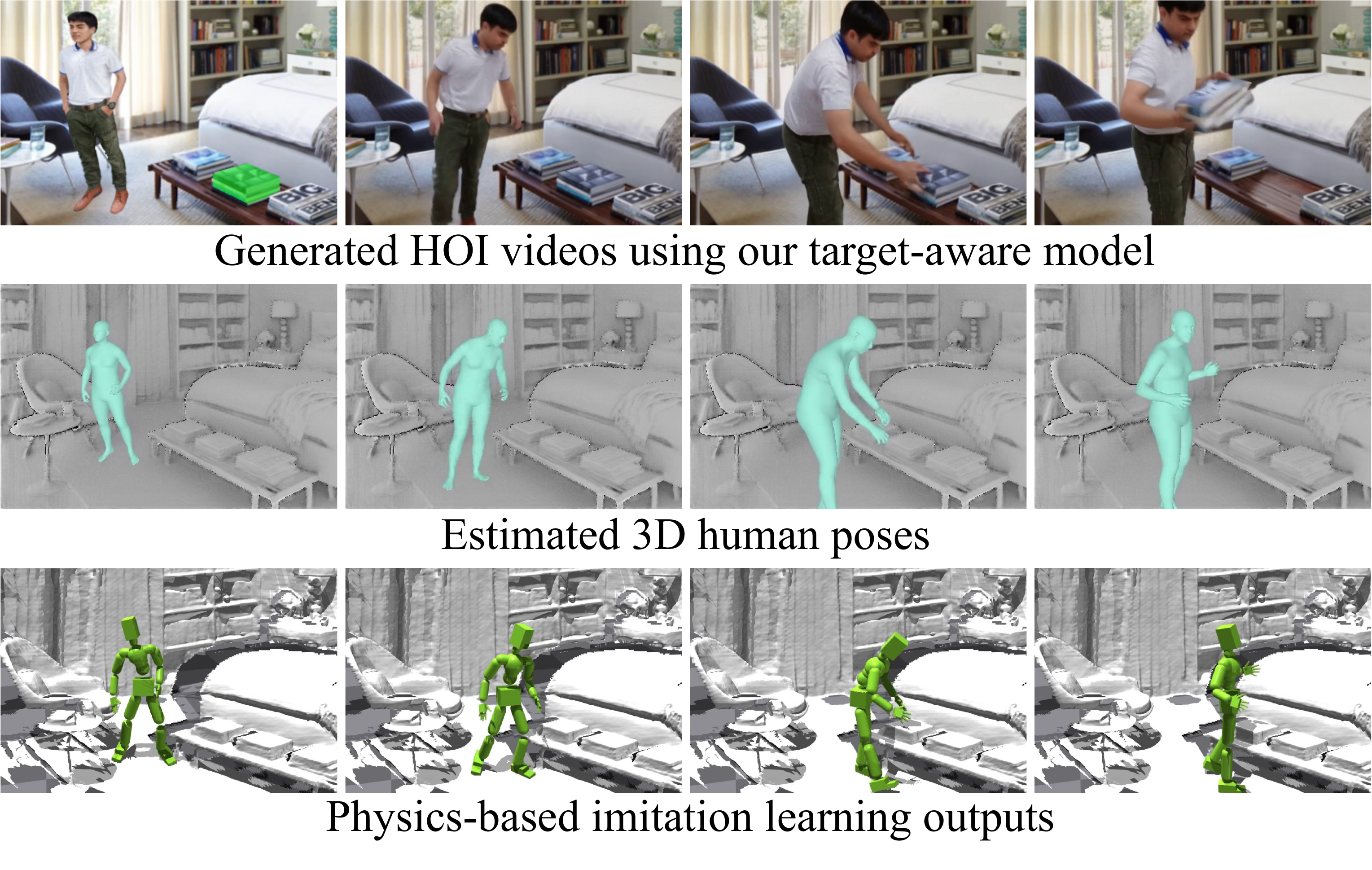}
        \caption{
            \textbf{Zero-shot 3D HOI motion synthesis.}
            We perform imitation learning on 3D poses of a person interacting with a target in the scene, obtained from videos generated with our model.
          }
        \label{fig:app2}
    \end{minipage}\hfill
    \raisebox{4pt}[0pt][0pt]{
    \begin{minipage}{0.484\linewidth}
        \centering
        \includegraphics[width=\linewidth, trim={2cm 0cm 2.2cm 0cm}]{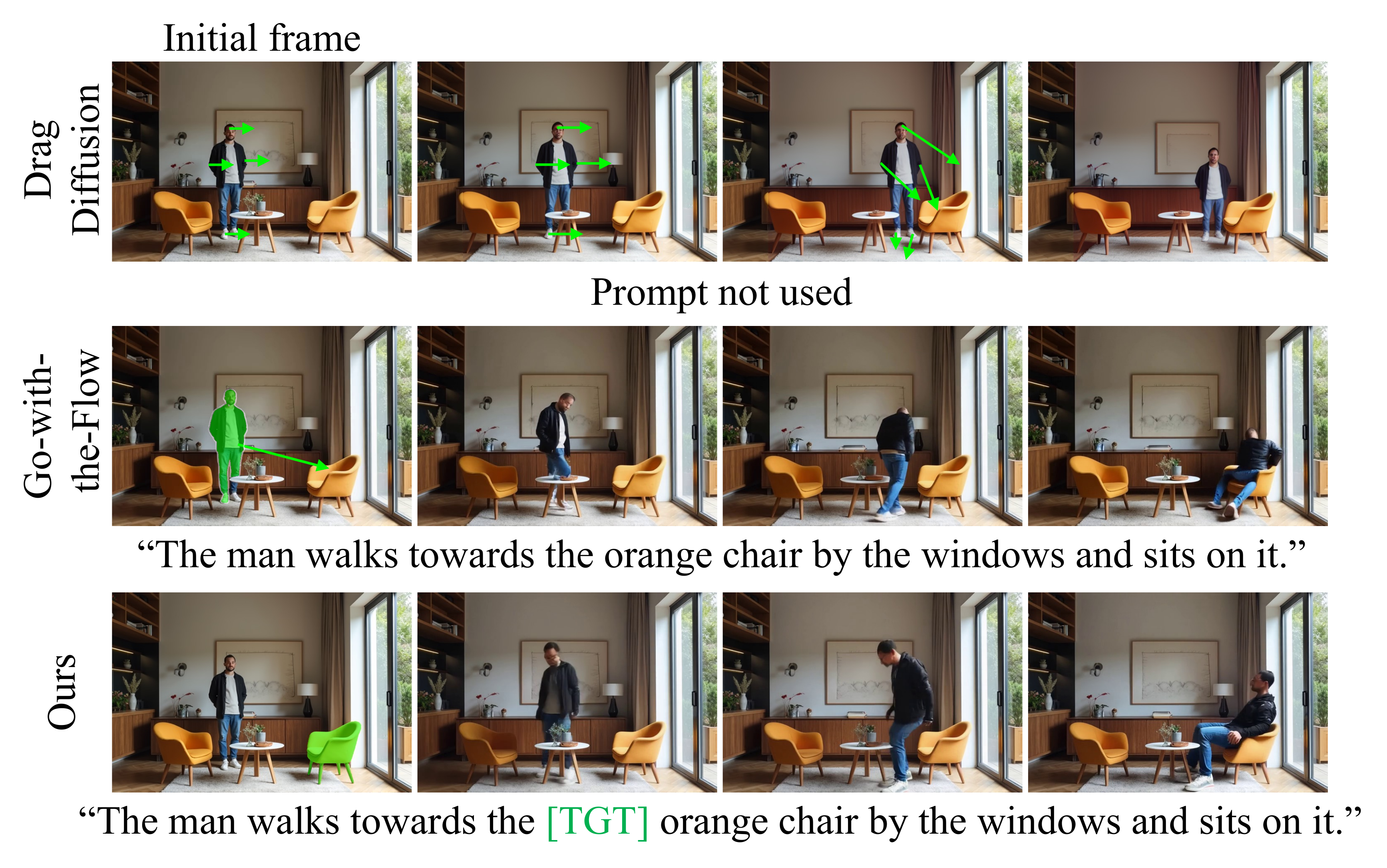}
        \caption{\textbf{Comparison over drag-based methods.} Ours with less extensive inputs outperforms drag-based editing methods~\citep{shi2024dragdiffusion, burgert2025gowiththeflowmotioncontrollablevideodiffusion}.}
        \label{fig:drag}
    \end{minipage}}
    \vspace{-10pt}
\end{figure}

\subsection{Zero-Shot 3D HOI Motion Synthesis with Physical Plausibility}
Given an actor in a 2D scene and a desired target object, our model produces realistic HOI actions aligned with text prompts, providing strong planning cues for robotics control~\citep{du2023learning, ajay2023compositional, black2023zero, ni2024generate, park2025demodiffusiononeshothumanimitation}. To validate this connection, we present a pipeline that first applies an off-the-shelf 3D human pose estimator~\citep{shen2024gvhmr} to videos generated by our model, extracting 3D human motion sequences. We then perform a physics-based imitation learning~\citep{wang2023physhoi} to train a policy that mimics these motions in the Isaac Gym simulator~\citep{makoviychuk2021isaac}. As shown in ~\cref{fig:app2}, the resulting agents reproduce human-object interactions in a physically plausible manner, demonstrating the potential of our approach to bridge between video generation and robotics.

\subsection{Long-term Video Generation with Target-Aware Interactions}
Our target-aware model serves as a key component for video content creation, enabling effective control over an actor's actions in a scene without extensive manual effort.
We introduce a simple yet robust pipeline that combines a video interpolation technique between keyframes and our target-aware video diffusion model to support two types of actions: navigating the scene and interacting with target objects.
For navigation, we interpolate between two keyframes using an off-the-shelf frame interpolation model~\citep{CogvideoXInterpolation}. Each keyframe is constructed by placing the actor at a desired location in the scene via our depth-aware 3D insertion method. To generate HOI actions with a specified target object, we first position the actor using the same insertion method, specify the target with an off-the-shelf segmentation tool~\citep{kirillov2023sam}, and finally employ our target-aware model to synthesize realistic interactions.
Importantly, our target-aware model provides a convenient way to produce plausible HOI scenes by simply specifying a target, which can then be connected through interpolation-based models for long-form video synthesis. The overall pipeline is illustrated in ~\cref{fig:overview_app} of Appendix~\cref{sec:app_supp}.

\section{Experiments}
\label{sec:experiments}
We are the first to introduce a target-aware video diffusion framework that explicitly models actor-target interactions. To rigorously evaluate this new task, we construct a dedicated benchmark, introduce metrics, and establish strong baselines for comparison.

\subsection{Experimental Setup}
\label{sec:exp_setup}

\noindent \textbf{Dataset.}
We construct a benchmark set of 80 images depicting scenes with a person, where each image is paired with a text prompt describing an interaction between the person and a target object. For all pairs, we ensure that the target can be clearly distinguished with text prompts using a noun, a color descriptor, or a spatial detail (e.g., soda bottle on the table, blue box at the center). Text prompts follow the format ``The person \{action\} with \{object\}.'' for baselines and ``The person \{action\} with [TGT] \{object\}.'' for ours. These prompts are further refined using GPT-4o, following the prompt enhancement procedure of CogVideoX~\citep{yang2024cogvideox}. For each test image, we generate 5 videos with random seeds, comparing results with a total of 400 samples.

\noindent \textbf{Metric.}
We evaluate our approach along two dimensions: target alignment and generation quality. To measure target alignment, we assess whether the generated video captures an accurate interaction between the person and the target object. Specifically, we employ an off-the-shelf contact detector~\citep{contacthands_2020} to identify human-object contact in each frame of the generated video and consider the interaction accurate if the detected contact regions overlap the target object's mask in at least one frame. We report the rate of accurate interactions over all generated videos (\emph{Contact Score}). In addition, we perform two types of user studies (\emph{Hum. Eval.} and \emph{User Pref.}) to further assess the target alignment of each method.
For generation quality, we adopt the evaluation metrics from VBench~\citep{huang2023vbench}, which break down video quality into subject consistency (\emph{SS}), background consistency (\emph{BC}), dynamic degree (\emph{DD}), motion smoothness (\emph{MS}), aesthetic quality (\emph{AQ}), and imaging quality (\emph{IQ}). The final score (\emph{Avg.}) is computed by averaging them. %

\noindent \textbf{Baselines.}
Since our method uses a single mask to specify the target, direct comparisons with state-of-the-art approaches that rely on heavy temporal conditioning on the actors are not appropriate. Instead, we evaluate against three representative baselines. First, we evaluate against our base image-to-video diffusion model, vanilla CogVideoX~\citep{yang2024cogvideox}. Second, we assess against a version of CogVideoX fine-tuned on videos of our dataset to isolate the effect of our method from that of the additional data (\emph{CogVideoX w. data}). Finally, we compare with the attention modulation method from Direct-a-video~\citep{yang2024direct}, which enforces a subject's trajectory by amplifying cross-attention weights within predefined bounding box regions (\emph{Attn. Mod.}). Since we assume that trajectory annotations for actors and targets are unavailable in our evaluation setting, we adapt this method by prepending the keyword ``target'' to the object description in the prompt and amplifying cross-attention weights in the target object mask region for that keyword. %

\subsection{Qualitative Evaluation}
\label{sec:qualitative}

\begin{figure}
\centering
\vspace{-15pt}
\includegraphics[width=1.0\linewidth, trim={3cm 0cm 0cm 1cm}]{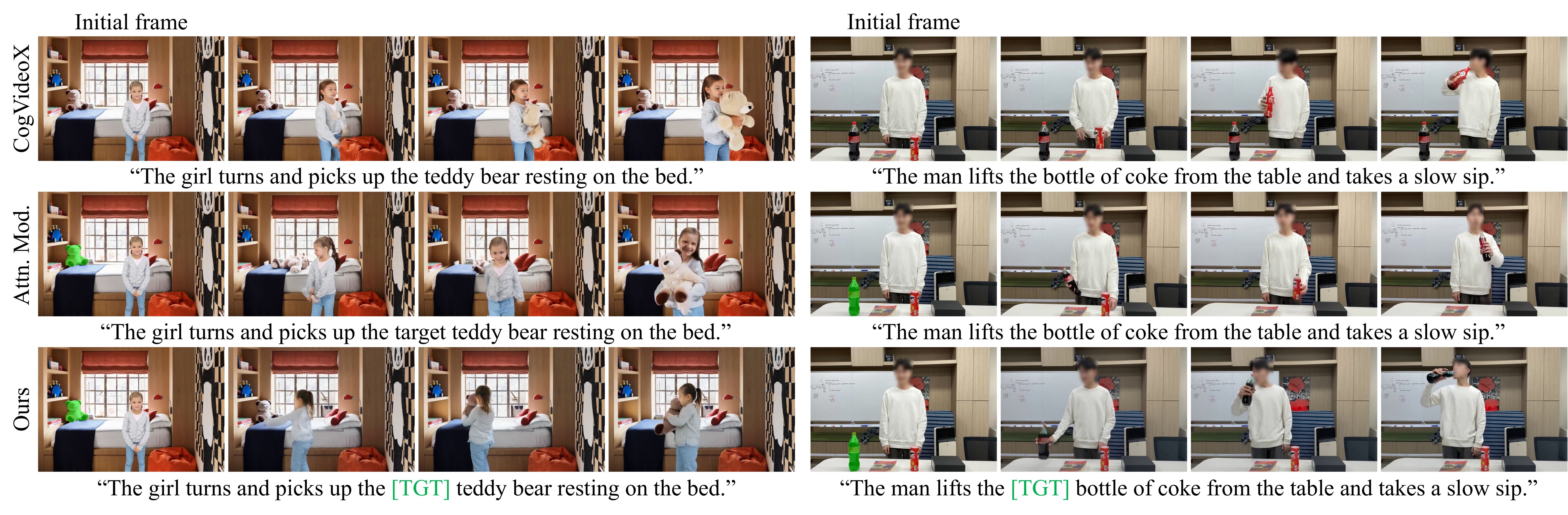}
\vspace{-15pt}
\caption{\textbf{Qualitative comparison on target alignment.} Each set displays generated videos using different methods. While baselines tend to hallucinate the target, our target-aware model produces videos where the actor interacts accurately with the actual target in the scene.}
\label{fig:target}
\end{figure}

\noindent \textbf{Target Alignment.}
\cref{fig:target} compares our method with baselines in terms of targeting accuracy. In rows 1 and 2, baseline methods occasionally hallucinate the target described in the prompt, rather than incorporating the actual target from the input image. In contrast, our approach generates videos where the actor accurately interacts with the specified target.

\begin{figure}[t]
    \centering
    \begin{minipage}{0.476\linewidth}
        \centering
        \includegraphics[width=\linewidth, trim={0cm 0cm 0cm 0cm}]{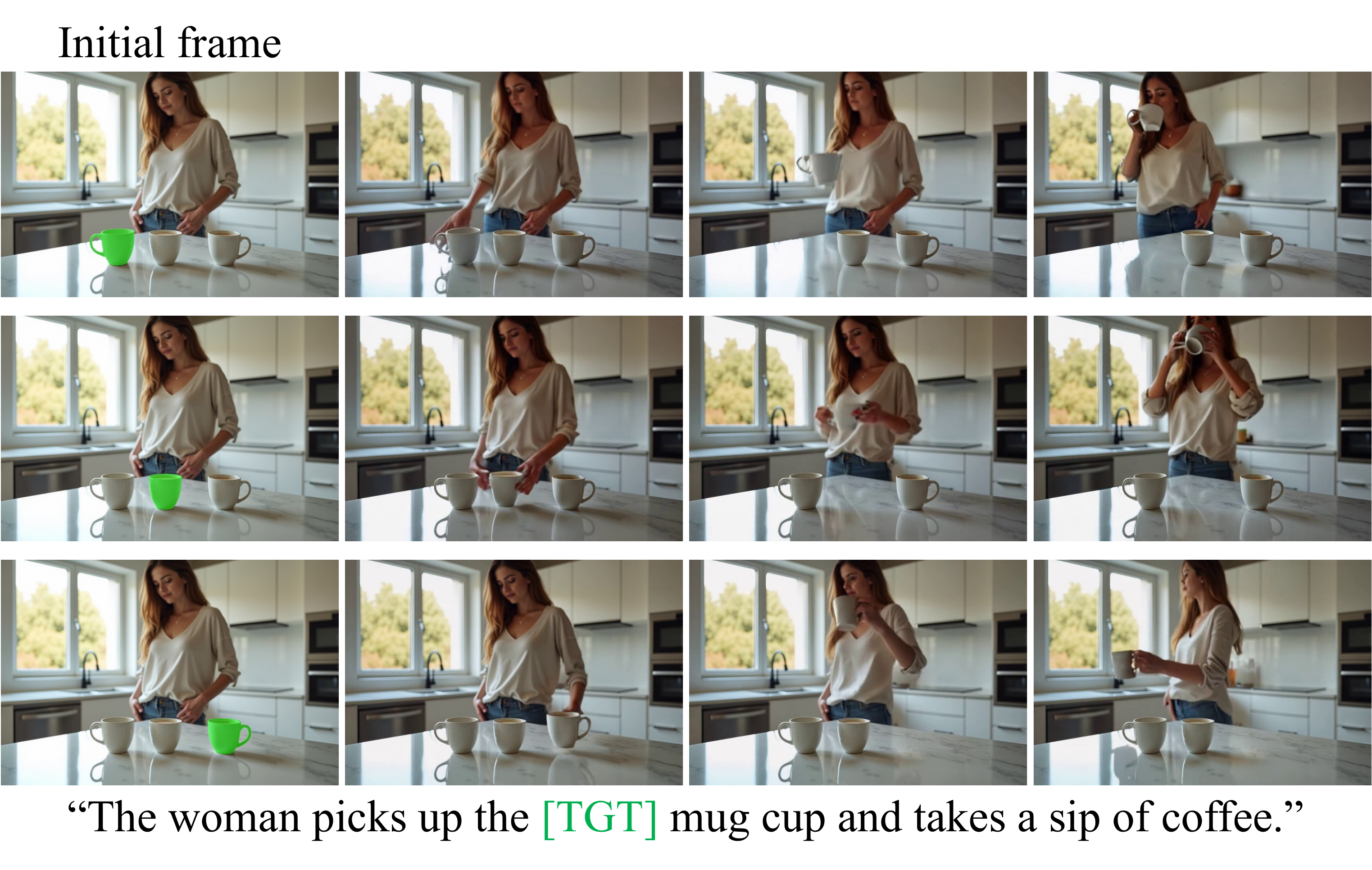}
        \caption{\textbf{Multiple objects of the same type.} 
        Our method ensures accurate interaction with the intended target by leveraging its mask.}
        \label{fig:multi_obj}
    \end{minipage}\hfill
    \begin{minipage}{0.484\linewidth}
        \centering
        \includegraphics[width=\linewidth, trim={0cm 0cm 0cm 0cm}]{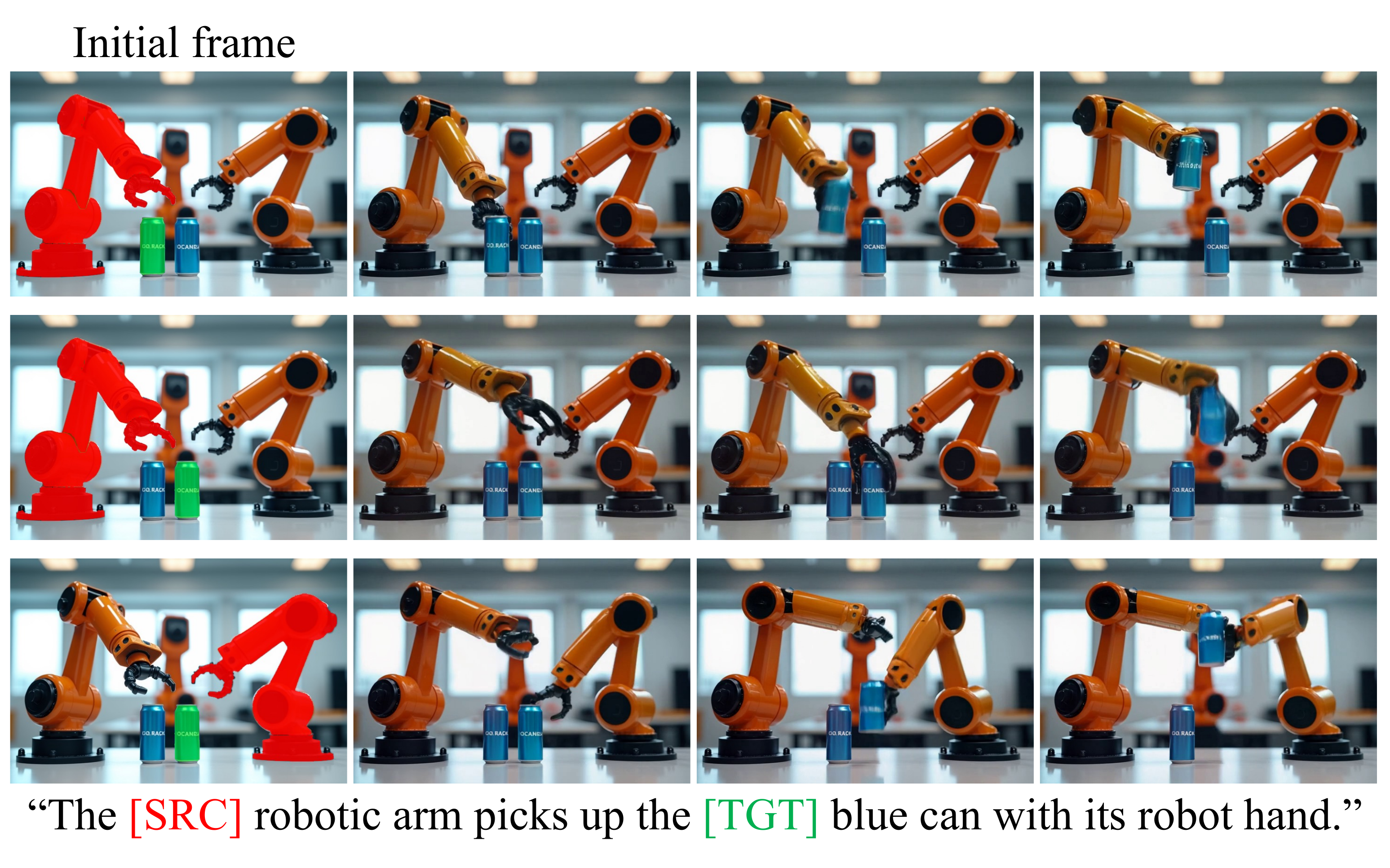}
        \caption{\textbf{Control over multiple entities.} 
        Our model can be extended to specify both the source actor and the target object using two masks.}
        \label{fig:two_masks}
    \end{minipage}
    \vspace{-20pt}
\end{figure}

\noindent \textbf{Multiple Objects of the Same Type.}
\cref{fig:multi_obj} highlights our key advantage where the scene contains multiple target objects of the same type. By enabling the usage of an explicit segmentation mask to identify the target, our method ensures precise selection and manipulation of the intended target.

\begin{figure}[t]
\vspace{-10pt}
\centering
\includegraphics[width=1.0\columnwidth, trim={0cm 0cm 0cm 0cm}]{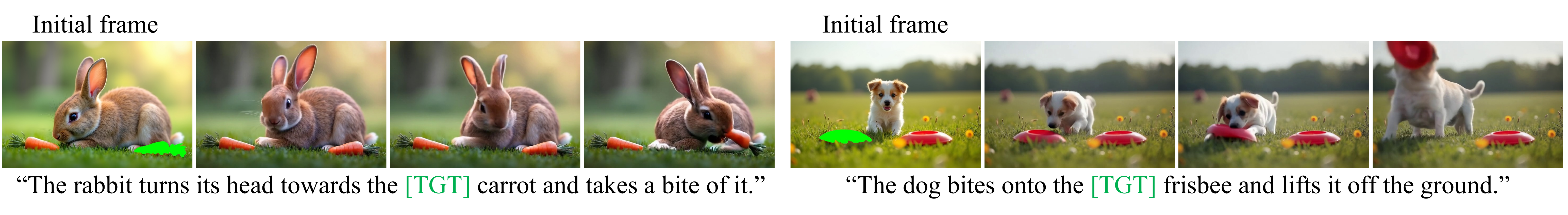}
\vspace{-15pt}
\caption{\textbf{Non-human interactions.}  Our target-aware model generalizes to non-human cases despite being fine-tuned on human-scene interaction videos.}
\label{fig:non_hum}
\vspace{-15pt}
\end{figure}
\noindent \textbf{Non-Human Interactions.}
While our model is fine-tuned on human-scene interactions, it generalizes well to interactions involving non-human subjects, as shown in \cref{fig:non_hum}.

\begin{table}[t]
\centering
\resizebox{\columnwidth}{!}{  
\begin{tabular}{lcccccccccc}
\toprule
& \multicolumn{3}{c}{\textbf{Targeting Quality}} & \multicolumn{7}{c}{\textbf{Video Quality}} \\
\cmidrule(lr){2-4} \cmidrule(lr){5-11}
& Contact Score$\uparrow$ & Hum. Eval.$\uparrow$ & User Pref.$\uparrow$ & SC$\uparrow$ & BC$\uparrow$ & DD$\uparrow$ & MS$\uparrow$ & AQ$\uparrow$ & IQ$\uparrow$ & Avg.$\uparrow$\\
\midrule
CogVideoX & 0.560 & 0.456 & 28.4\% & 0.893 & 0.898 & 0.883 & 0.988 & 0.502 & 0.694 & 0.810 \\
CogVideoX w.data & 0.638 & 0.596 & 36.2\% & 0.914 & 0.907 & 0.907 & 0.990 & 0.492 & 0.653 & 0.810 \\
Attn. Mod. & 0.546 & 0.508 & 22.2\% & 0.872 & 0.889 & 0.786 & 0.986 & 0.499 & 0.687 & 0.786 \\
Ours & \textbf{0.878} & \textbf{0.892} & (100\%-above) & 0.933 & 0.919 & 0.899 & 0.937 & 0.496 & 0.656 & 0.807 \\

\bottomrule
\end{tabular}
}
\caption{\textbf{Quantitative comparison.} Our method enables the generation of videos containing accurate interactions with the specified targets. We also report generation quality with measures from VBench~\citep{huang2023vbench}, confirming that our approach does not compromise video quality.}
\label{tab:main}
\end{table}

\noindent \textbf{Specifying Both the Actor and the Target.}
Our approach enables simultaneous control over both the source actor and the target object, as demonstrated in \cref{fig:two_masks}. We extend our model to accept two separate segmentation masks as additional inputs and introduce two tokens, [SRC] and [TGT]. Each token is encouraged to attend to each mask with our cross-attention loss during fine-tuning. During inference, we prepend each token to the actor and target descriptions, respectively.

\noindent \textbf{Comparison with Drag-Based Methods.}
~\cref{fig:drag} shows a qualitative comparison with two drag-based image/video editing methods, which require additional user inputs to directly control the actor. DragDiffusion~\citep{shi2024dragdiffusion} adjusts image latents based on drag operations. Since a single large drag is ineffective, we gradually move the actor toward the target using multiple small drags. While DragDiffusion produces reasonable results for small translations, it fails for larger adjustments. Go-with-the-Flow~\citep{burgert2025gowiththeflowmotioncontrollablevideodiffusion} controls motion by warping the initial noise sequence to follow a desired flow, which we implement by dragging the actor's segmentation mask toward the target. Although this method enables the actor to make contact with the target, the output video lacks plausible motion due to its coarse conditioning. In contrast, our approach produces realistic interactions even without explicit motion guidance.%

\subsection{Quantitative Evaluation}
\label{sec:quantitative}

\noindent \textbf{Target Alignment and Video Quality.}
As presented in ~\cref{tab:main} \emph{Contact Score}, our method substantially outperforms all baselines in generating accurate interactions with target objects. 
At the same time, ours maintains video generation quality, achieving comparable scores to baselines across the \emph{Video Quality} metrics in ~\cref{tab:main}.
The attention modulation approach fails to maintain the temporal consistency of videos since the amplified cross-attention values adversely affect the self-attention values, resulting in low contact scores. Additional details are provided in Appendix~\cref{sec:attention_supp}.

\noindent \textbf{User Study.}
We conduct two types of user studies via CloudResearch Connect.
In (\emph{Hum. Eval.}), each generated video is presented together with the corresponding input image and target object specification. Participants then make a binary judgment on whether the actor interacts accurately with the specified target. A total of 50 participants evaluate 10 videos per method, and we report the overall rate of accurate interactions in ~\cref{tab:main}.
In (\emph{User Pref.}), each input image is presented alongside two generated videos: one produced by our method and the other by a baseline. Participants are asked to choose which video better reflects accurate interaction with the target. Again, 50 participants answer 10 questions per baseline, and we report in ~\cref{tab:main} the proportion of times each baseline is preferred over ours.
In both studies, participants consistently favor our outputs by a large margin.

\begin{minipage}[t]{0.448\linewidth}
  \centering

\centering
\begingroup
\resizebox{\columnwidth}{!}{
\begin{tabular}{lcc}
\toprule
       & Contact Score$\uparrow$ & Video Quality$\uparrow$ \\
\midrule
Random          & 0.819 & 0.807 \\
Equally-Spaced  & 0.816 & 0.800 \\
Ours            & \textbf{0.878} & 0.807 \\
\bottomrule
\end{tabular}}
\endgroup

  \captionof{table}{\textbf{Cross-attention loss on selective transformer blocks.} 
  We apply the loss to blocks that best capture semantics.}
  \label{tab:ablation_layer}
\end{minipage}\hfill
\begin{minipage}[t]{0.512\linewidth}
  \centering

\centering
\begingroup
\resizebox{\columnwidth}{!}{
\begin{tabular}{lcc}
\toprule
       & Contact Score$\uparrow$ & Video Quality$\uparrow$ \\
\midrule
T2V Cross-Attn.       & 0.740 & 0.806 \\
Both Cross-Attn.      & 0.860 & 0.810 \\
Ours (V2T Cross-Attn.) & \textbf{0.878} & 0.807 \\
\bottomrule
\end{tabular}}
\endgroup

  \captionof{table}{\textbf{Cross-attention loss on selective attention regions.} 
  We apply the loss to the attention regions that most influence target awareness.}
  \label{tab:ablation_area}
\end{minipage}

\noindent
\begin{minipage}[t]{0.48\linewidth}
  \centering

\centering
\small
\begin{tabular}{lcc}
    \toprule
           & Contact Score$\uparrow$ & Video Quality$\uparrow$ \\
    \midrule
    $\lambda_{attn}=0.0$ & 0.647 & 0.815 \\
    $\lambda_{attn}=0.05$ & 0.727 & 0.811 \\
    $\lambda_{attn}=0.1$ & 0.878 & 0.807 \\
    $\lambda_{attn}=0.25$ & \textbf{0.890} & 0.806 \\
    $\lambda_{attn}=0.5$ & 0.888 & 0.807 \\
    $\lambda_{attn}=1.0$ & 0.888 & 0.804 \\
    \bottomrule
\end{tabular}

  \captionof{table}{\textbf{Effects of different cross-attention loss weights.} 
  Incorporating our cross-attention loss is crucial for achieving target awareness.}
  \label{tab:ablation_lambda}
\end{minipage}\hfill
\raisebox{5pt}[0pt][0pt]{
\begin{minipage}[t]{0.48\linewidth}
  \centering

\centering
\small{

\begin{tabular}{lcc}
\toprule
       & Contact Score$\uparrow$ & Video Quality$\uparrow$ \\
\midrule
original & 0.896 & 0.812  \\
dilate-3 & 0.884 & 0.808 \\
dilate-5 & 0.872 & 0.815 \\
erode-3 & \textbf{0.904} & 0.815 \\
erode-5 & 0.880 & 0.813 \\

\bottomrule
\end{tabular}
}

  \captionof{table}{\textbf{Effect of the mask quality.} Even when masks are expanded or shrunk, the Contact Score remains stable, supporting that our method is not sensitive to precise segmentation.}
  \label{tab:mask_acc}
\end{minipage}}

\subsection{Ablation Studies}
\label{sec:ablation}

\noindent \textbf{Cross-Attention Loss on Selective Blocks.}
We evaluate the impact of applying cross-attention loss on different transformer blocks. For all experiments, we fix the number of blocks receiving the loss per training step to seven. We compare three strategies: (1) random seven blocks at each training step, (2) seven equally spaced blocks, and (3) blocks chosen using our proposed method. As shown in ~\cref{tab:ablation_layer}, our approach shows improved target alignments.

\noindent \textbf{Cross-Attention Loss on Selective Regions.}
In ~\cref{tab:ablation_area}, we examine how applying cross-attention loss on various regions of the cross-attention influences performance. The results indicate that the loss should be applied on the V2T cross-attention for better target alignment.

\noindent \textbf{Cross-Attention Loss Weight.}
In \cref{tab:ablation_lambda}, we analyze the impact of cross-attention loss coefficient $\lambda_{attn}$. 
When $\lambda_{attn}=0.0$, meaning the model is trained solely with the reconstruction loss, the target alignment performance is nearly identical to that of the CogVideoX fine-tuned on our dataset (\cref{tab:main}, second row). 
This demonstrates that simply introducing the mask does not, by itself, improve the target awareness of the model, and incorporating the cross-attention loss is essential. 
As we increase the loss weight, we observe a saturation of the contact score and set $\lambda_{attn}=0.1$.

\noindent \textbf{Quality of Masks.} To evaluate the effect of segmentation quality on target alignment, we dilate and erode the masks at varying levels. As shown in~\cref{tab:mask_acc}, our method remains robust to these perturbations, as even coarse masks effectively guide the model by narrowing the region of interest.

\begin{wraptable}{r}{0.52\linewidth} %
  \centering
  \vspace{-12pt}

\centering
\small{

\begin{tabular}{lcc}
\toprule
       & Contact Score$\uparrow$ & Video Quality$\uparrow$ \\
\midrule
original & \textbf{0.896} & 0.812  \\
circular-15 & 0.838 & 0.813 \\
circular-30 & 0.888 & 0.809 \\

\bottomrule
\end{tabular}
}

  \caption{\textbf{Effect of the mask shape.} Our method does not depend on the exact mask shape.}
  \label{tab:mask_circle}
  \vspace{-10pt}
\end{wraptable}
\noindent \textbf{Shape of Masks.} We further assess the model's robustness to mask shape by replacing the original mask with a circular mask centered at the original mask's bounding box center, with a radius of 15 or 30 pixels. As demonstrated in ~\cref{tab:mask_circle}, our model allows abstract spatial cues as input.

\section{Conclusion}
\label{sec:discussion}

We presented a target-aware video diffusion model that generates videos where an actor plausibly interacts with a specified target, defined by a segmentation mask in the first frame.
Our goal is to establish target awareness as a core capability of video generative models. Under this formulation, our model naturally guides the actor to interact with the designated targets in realistic and semantically consistent ways.
Experiments demonstrate the strengths and advantages of our model compared to baseline methods and alternative solutions.
Finally, we present key applications of zero-shot 3D HOI motion synthesis and video content creation using our target-aware video diffusion model.

\section*{Acknowledgments}
This work was supported by RLWRLD, NRF grant funded by the Korean government (MSIT) [No. RS-2022-NR070498, No. RS-2025-25396144, and No. PJT-25-122310], and IITP grant funded by the Korea government (MSIT) [No. RS-2024-00439854, No. RS-2025-25442338, No. RS-2021-II211343, and No. RS-2025-02653113]. H. Joo is the corresponding author.

We thank Jeonghwan Kim for implementing physics-based imitation learning and rendering those results. We also thank Inhee Lee for proofreading the paper. H. Joo is the corresponding author.


\newpage
\appendix
\section{Implementation Details}
\label{sec:implementation}

\subsection{Training and Inference Details}
\label{sec:training_details}
We use the CogVideoX-5B-I2V model~\citep{yang2024cogvideox} as our base image-to-video diffusion model, producing output videos at a resolution of 720 $\times$ 480 with a total of 49 frames. For model fine-tuning, we employ LoRA~\citep{hu2021lora} with a rank of 128 and $\alpha=64$ to the diffusion transformer and designate the word ``target'' as our [TGT] token. We optimize the added LoRA layers and the extended image projection layer while keeping the other parts of the model frozen and train for 2,000 steps using an AdamW optimizer with a learning rate of $1\times10^{-4}$, and an effective batch size of 4. The addition of a single channel to the image projection layer for incorporating the target mask introduces 15,360 additional parameters to the 5B-parameter base model, resulting in negligible extra computational overhead. We set the cross-attention loss coefficient $\lambda_{attn}=0.1$ and apply the loss to the video-to-text (V2T) cross-attention regions of every third transformer block from the 5th block to the 23rd block of the total 42 blocks. This selective application reduces VRAM usage by 71\% compared to applying the loss across all blocks. We average the attention maps of these selected blocks and regions and normalize them to $[0, 1]$. The overall training takes approximately 6 hours on 4 NVIDIA A100 GPUs.
For inference, we employ a DPM sampler~\citep{lu2022dpm} with $T=50$ sampling steps and set the classifier-free guidance scale~\citep{ho2022classifier} to 6 with the same dynamic guidance strategy as the original work~\citep{yang2024cogvideox}. The inference for a single video approximately takes 249.8 seconds on a single NVIDIA A100 GPU.

\subsection{Details on Applications}
\label{sec:app_supp}

\noindent \textbf{Application 1: Physics-Based Imitation Learning.}
We use the official code of PhysHOI~\citep{wang2023physhoi} to implement physics-based imitation learning on 3D human poses extracted from our output videos. Since our goal is to learn a policy for human motion, we disable modules related to object motions during training. Joint training of full modules by obtaining paired data of 3D human pose and 3D object pose via an off-the-shelf object 6D pose estimator~\citep{zhang2023genpose, wen2024foundationpose} could be a possible extension. Also, our current imitation learning outputs, as shown in ~\cref{fig:app_supp}, are manually aligned with the 3D scene due to different scales between estimated 3D human pose translations and the scene. Since the 3D location of the initial pose is given through 3D insertion of humans, we may adjust the scale of the subsequent translations, leveraging the depth information, which we leave as future work.

\begin{figure}
\centering
\includegraphics[width=0.7\columnwidth, trim={2cm 0cm 0cm 0cm}]{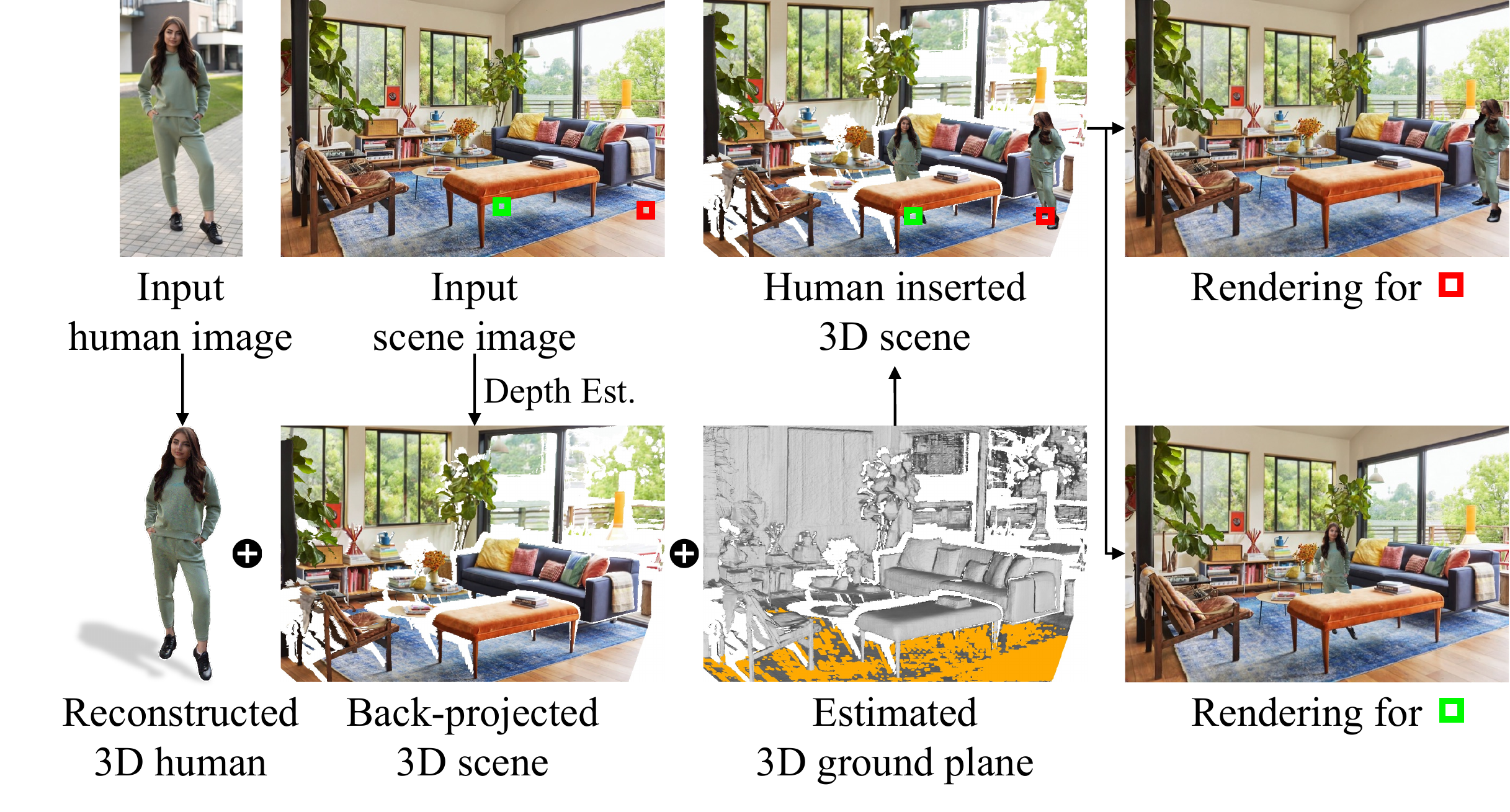}
\caption{\textbf{Inserting humans into scenes.} We perform a 3D depth-based insertion of the human into the scene by performing single-view 3D reconstruction of the human and estimating the depth of the scene.}
\label{fig:insertion}
\end{figure}

\noindent \textbf{Application 2: Inserting Humans into Scenes.}
\label{sec:app1}
As demonstrated in ~\cref{fig:insertion}, we perform human insertion in 3D space rather than in 2D pixel space to handle depth ordering and occlusions between the human and objects in the scene. Given an input human image, we use a single-view 3D human reconstruction method~\citep{albahar2023humansgd} to obtain a 3D reconstruction of the person. For the input scene image, we first apply a segmentation tool~\citep{kirillov2023sam} to identify and segment the ground. We then use a metric-depth estimation method~\citep{Bochkovskii2024:arxiv} to generate the real-scale 3D pointcloud. From the 3D pointcloud, we extract the points that belong to the ground when projected to the image and perform RANSAC-based plane fitting on these points to derive a 3D ground plane. Using the mapping between pixels and the 3D pointcloud, we obtain the 3D coordinates of the pixel to place the human.
The reconstructed 3D human is positioned at its 3D point, perpendicular to the ground plane. To manage occlusions between the 3D human and the 3D scene pointcloud, we discard cases where significant overlap occurs and ask the user for alternative input coordinates. Once occlusion handling is complete, we render them together to obtain the human-inserted scene images.

\begin{figure}
\centering
\includegraphics[width=0.75\columnwidth, trim={0cm 0cm 0cm 0cm}]{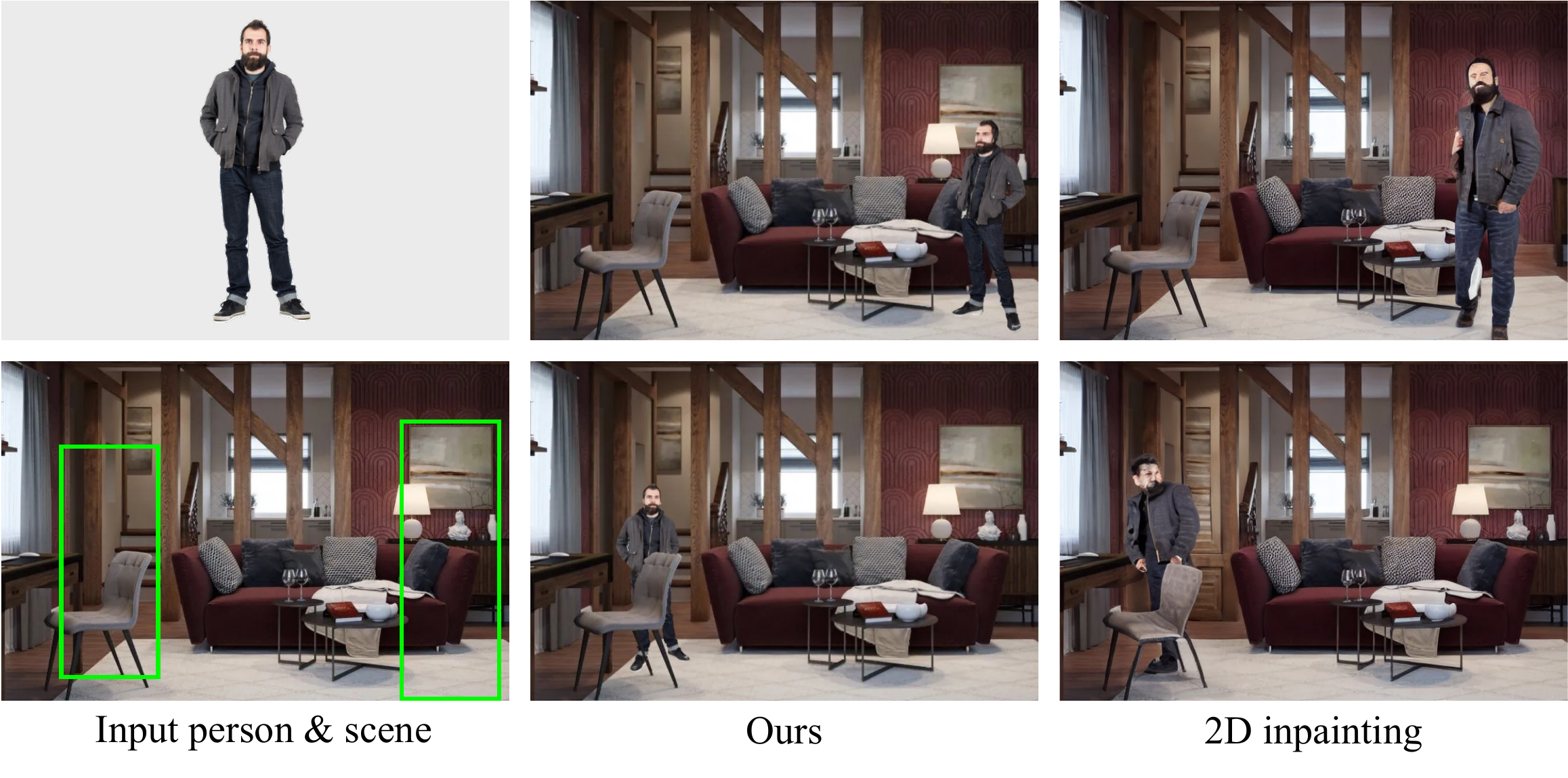}
\caption{\textbf{Comparison with 2D inpainting.} Our 3D-based human insertion effectively inserts the human into the scene while preserving both identities and handling occlusion.}
\label{fig:inpaint}
\end{figure}

Compared to 2D-based solutions using inpainting, where a specific region of the scene is masked and the person is inserted via personalized diffusion models~\citep{rombach2022high, ye2023ip}, our 3D approach better preserves the appearance of the original inputs. As shown in ~\cref{fig:inpaint}, inpainting-based methods often fail to maintain consistency with the original scene, resulting in undesired removal of objects in the scene. Additionally, inpainting can generate random details for occluded parts of the person in the input image, leading to inconsistencies between frames.

\begin{figure}
\includegraphics[width=1.0\linewidth, trim={0cm 0cm 0cm 1cm}]{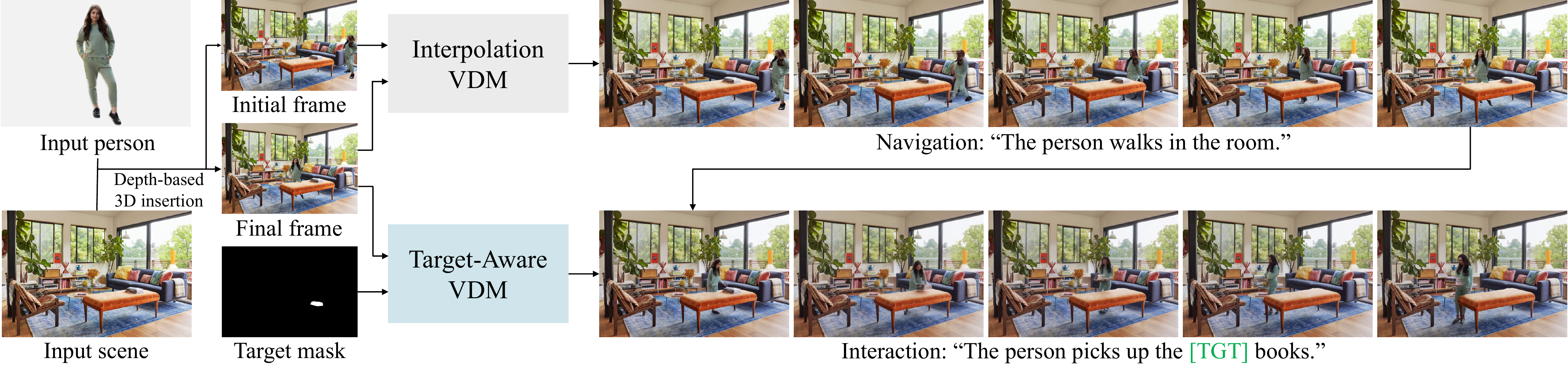}
\caption{\textbf{Video content creation.} Given images of a person and a scene, we perform depth-based 3D insertion of the person into the scene and render them together to produce frames for video diffusion input. We interpolate generated initial and final frames to synthesize navigation contents, and utilize our target-aware video diffusion model to synthesize interaction contents.}
\label{fig:overview_app}
\end{figure}

\subsection{Evaluation Details}
\begin{figure}
\centering
\includegraphics[width=1.0\columnwidth, trim={0cm 0cm 0cm 0cm}]{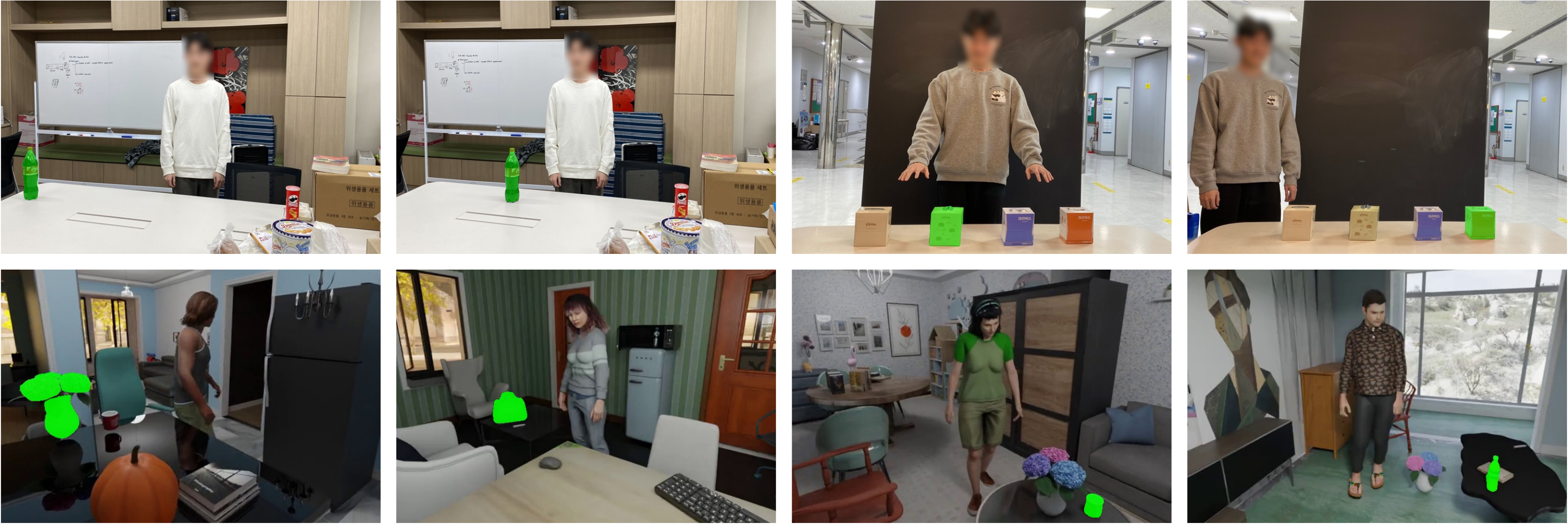}
\caption{\textbf{Evaluation data example.} We show a subset of our evaluation dataset, where the target is indicated with a green mask. We confirm that the target is fully distinguishable with the input text prompt.}
\label{fig:eval_data}
\end{figure}

\noindent \textbf{Evaluation Dataset.}
We construct a set of images of a scene containing a person paired with the prompt depicting an interaction between the person and the target. In the prompt, the person is described to interact with (1) an object placed at different locations of the scene relative to the person's position, or (2) a specific object among several different objects in the scene. For all images, we ensure that the target can be precisely determined with text prompts by a noun, a color description, or a spatial description. To get the target mask, we first perform instance segmentation on the image using SAM~\citep{kirillov2023sam}, and manually select the masks that belong to the target. Some image and mask samples of our evaluation dataset are presented in ~\cref{fig:eval_data}.

\begin{figure}
\centering
\includegraphics[width=0.75\columnwidth, trim={0cm 0cm 0cm 0cm}]{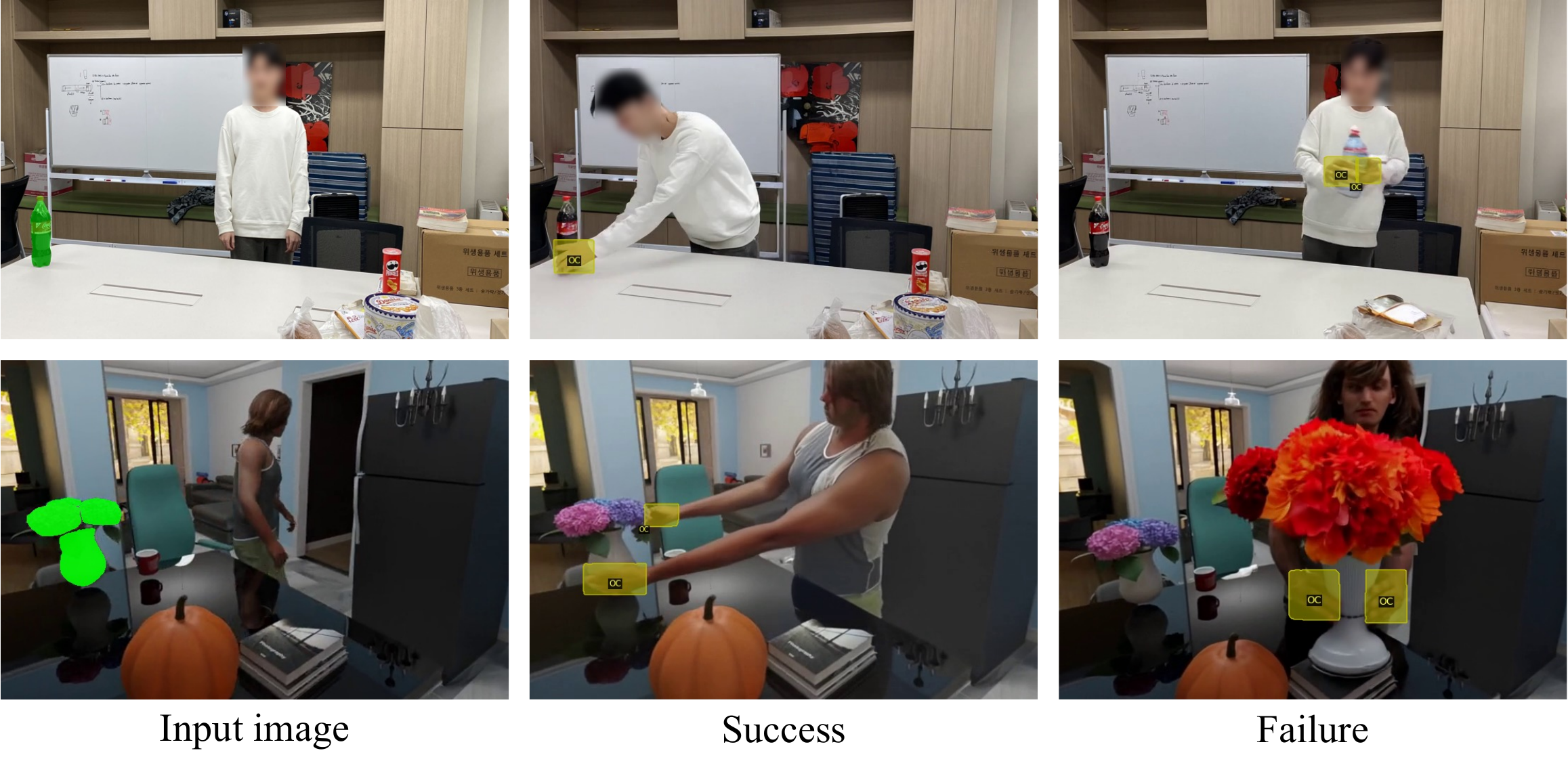}
\caption{\textbf{Contact score based on detection.} Green masks in the input image indicate the target, and yellow masks in output video frames indicate the detected object contact regions. We consider the interaction with the target accurate if the detected object contact region overlaps with the target mask.}
\label{fig:contact}
\end{figure}

\noindent \textbf{Contact Score.}
To detect physical contact between the actor and objects, we use the official code of ContactHands~\citep{contacthands_2020}. We set the hand detection threshold to 0.5 and the object contact detection threshold to 0.5. We consider the interaction between the actor and the target to be successful when the detected object contact region overlaps with the segmentation mask of the target. Detection results for success and failure cases are presented in ~\cref{fig:contact}, where the yellow masks indicate the detected object contact regions.

\begin{figure}[t]
\centering
\begin{subfigure}[t]{0.48\linewidth}
  \centering
  \includegraphics[width=\linewidth, trim={0cm 0cm 0cm 0cm},clip]{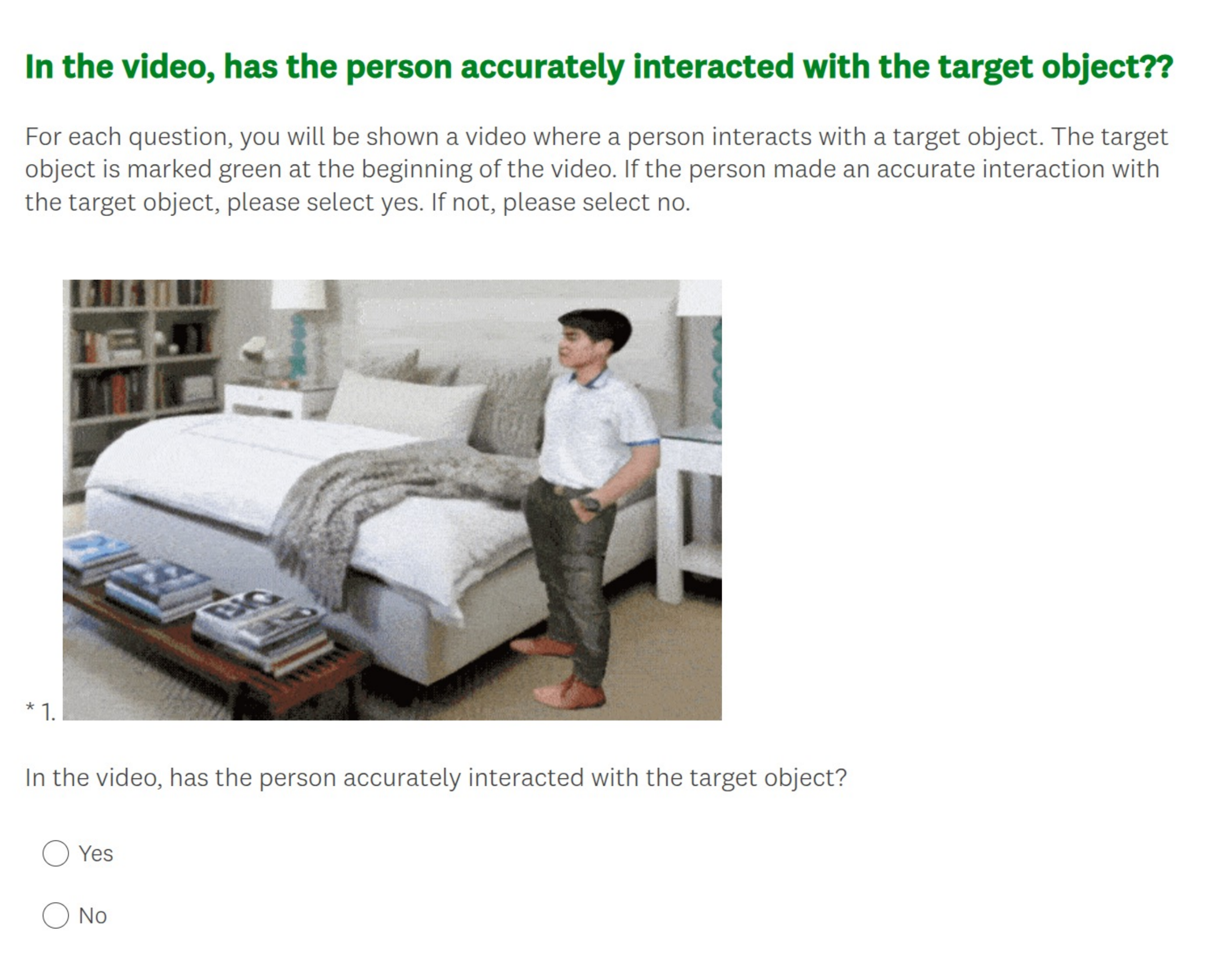}
  \caption{Human evaluation on target alignment.}
  \label{fig:user1}
\end{subfigure}\hfill
\begin{subfigure}[t]{0.48\linewidth}
  \centering
  \includegraphics[width=\linewidth, trim={0cm 0cm 0cm 0cm},clip]{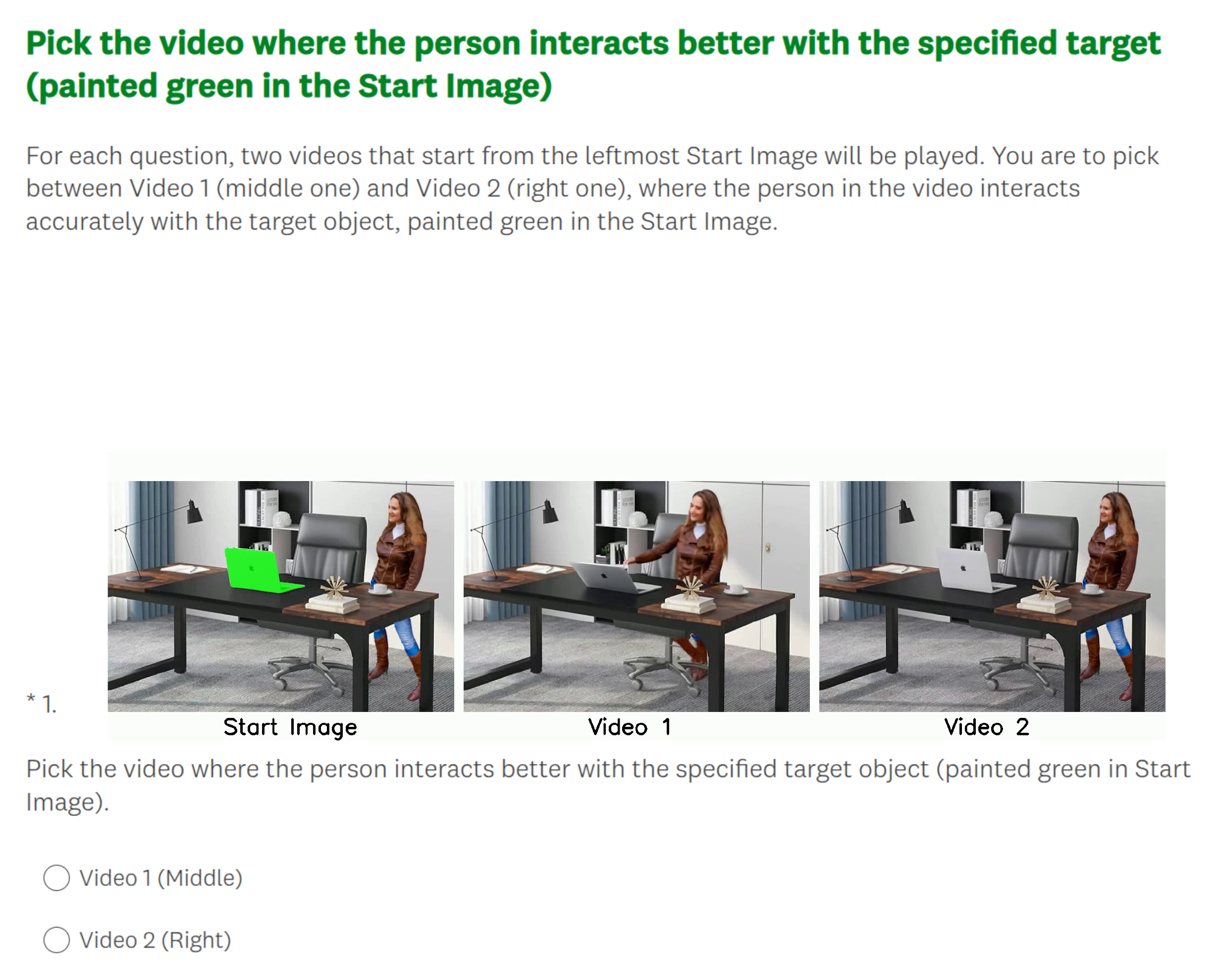}
  \caption{User preference study on target alignment.}
  \label{fig:user2}
\end{subfigure}
\caption{\textbf{User studies on target alignment.} (a) We ask participants to assess whether the person in the video accurately interacts with the target. (b) We ask participants to select the video that better demonstrates accurate target alignment.}
\label{fig:user_study}
\end{figure}

\noindent \textbf{User Study.}
We conduct two types of user studies to validate our comparisons. First, we perform human evaluation on videos generated by each method to assess whether the depicted person accurately interacts with the target. Fifty participants are presented with 10 videos per method. In each video, we show an input image with the target object highlighted with a green mask for 2 seconds, followed by the generated video. A screenshot of the human evaluation interface is shown in ~\cref{fig:user1}. We also perform an A/B test to measure user preference of our method over each baseline method in terms of target alignment. Fifty participants are asked 10 questions per baseline method, where each question displays an input image with the target in a green mask, alongside the generated videos from our method and a baseline in a random order. The screenshot of the user preference study interface is shown in ~\cref{fig:user2}.

\section{Additional Qualitative Results}
\label{sec:qual_add}

\begin{figure}
\centering
\includegraphics[width=1.0\columnwidth, trim={0cm 0cm 0cm 2cm}]{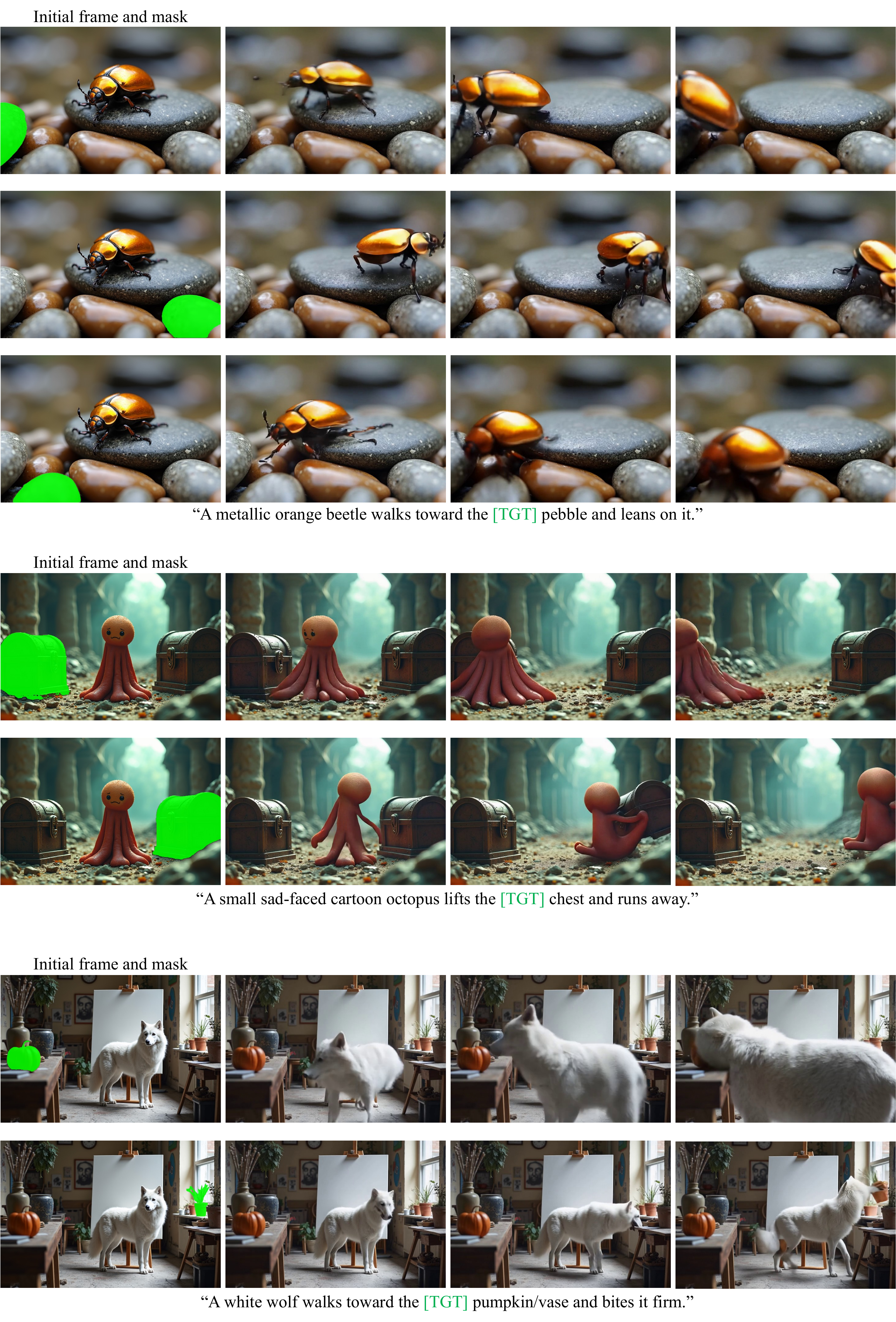}
\caption{\textbf{Non-human interactions.} Our target-aware model generalizes well to non-human interactions despite being fine-tuned solely on 1K+ human-scene interaction videos.}
\label{fig:supp_nonhum}
\end{figure}

\noindent \textbf{Non-human Interactions.}
\cref{fig:supp_nonhum} is an extended figure of \cref{fig:non_hum} in the main paper, demonstrating that our model generalizes to non-human interactions. Although the model is fine-tuned solely on a relatively small set of human-scene interaction videos and has never been exposed to actors other than humans during training, it can generate coherent and plausible non-human interactions. This generalization ability arises from the strong generative priors of the base video diffusion model, enabling the itself to adapt to novel agent types with target awareness.

\begin{figure}
\centering
\includegraphics[width=1.0\columnwidth, trim={0cm 0cm 0cm 2.5cm}]{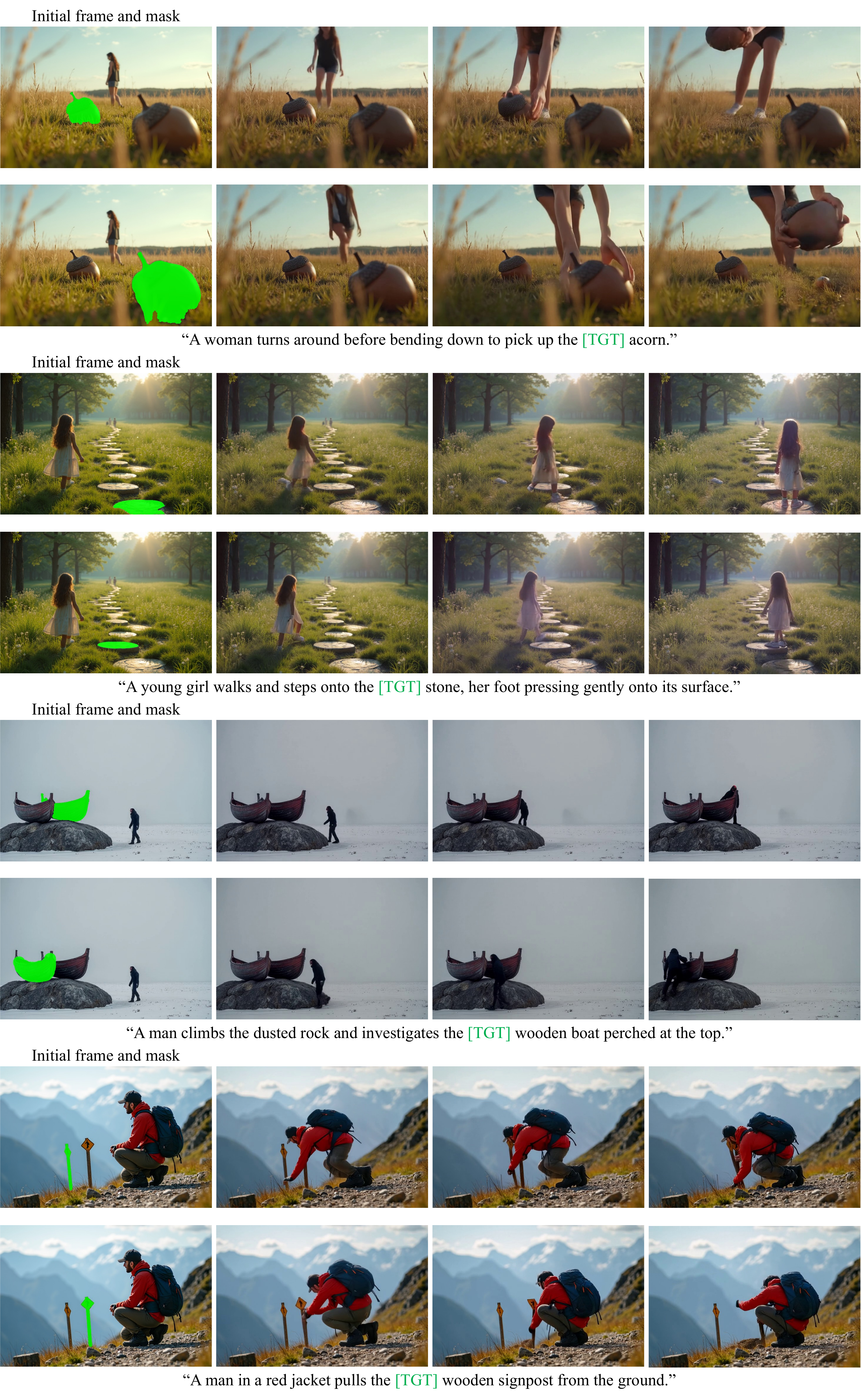}
\caption{\textbf{Target-aware generation in outdoor scenes.}
Despite being fine-tuned only on indoor interaction videos, our model generalizes to interactions in diverse outdoor scenes.}
\label{fig:supp_outdoors}
\end{figure}

\noindent \textbf{Targeting Objects in Outdoor Scenes.}
\cref{fig:supp_outdoors} presents additional results of our target-aware model applied to outdoor scenes. Although the model is trained exclusively on indoor interaction videos, it generalizes well to diverse outdoor environments, successfully grounding the target and producing coherent actor-target interactions.

\begin{figure}
\centering
\includegraphics[width=1.0\columnwidth, trim={1cm 0cm 1cm 0cm}]{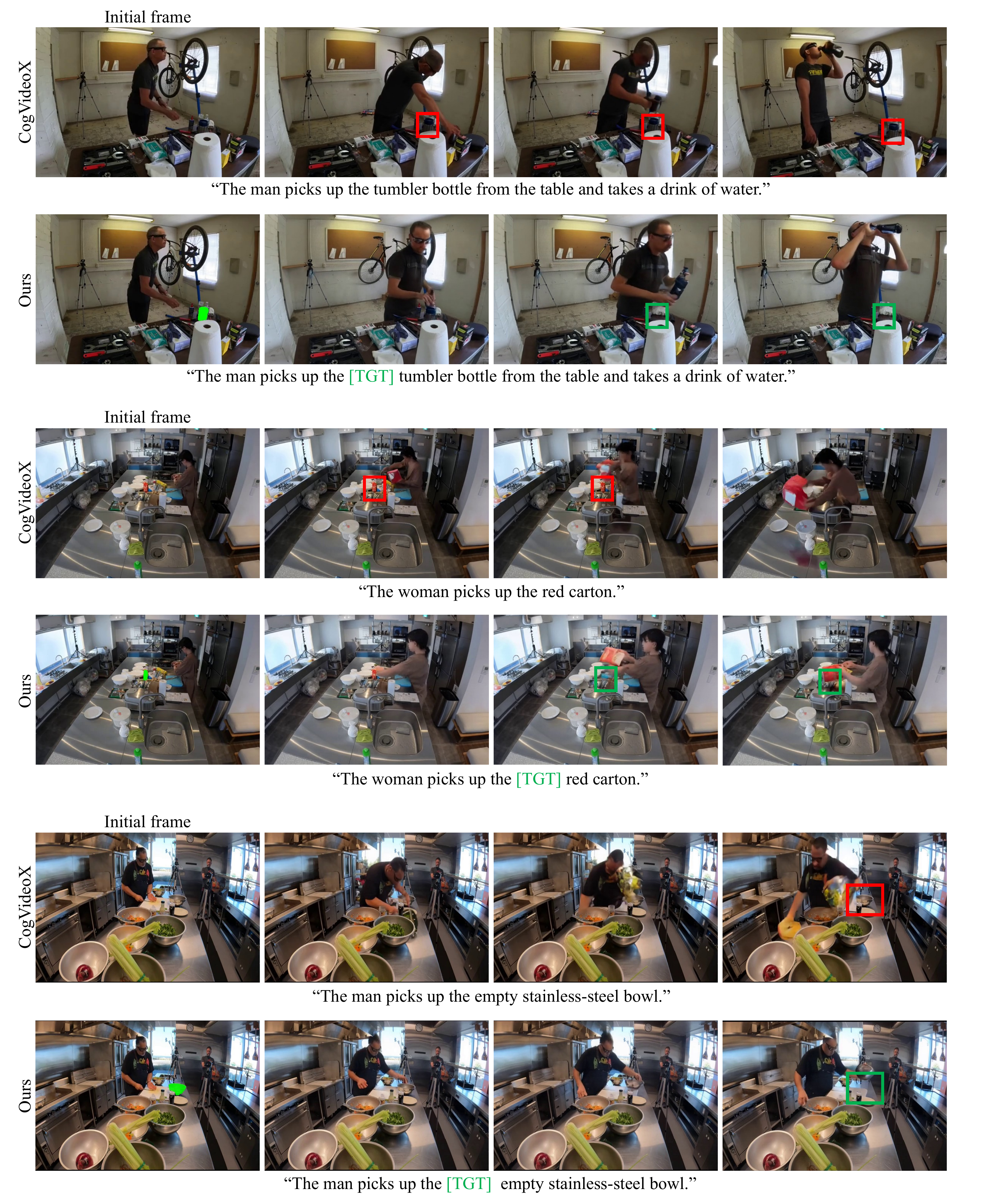}
\caption{\textbf{Targeting objects in complex scenes.} We compare results of original CogVideoX~\citep{yang2024cogvideox} and our target-aware model in complex scenes. Our model successfully generates target-aligned videos even when the target appears small in complex scenes. Best viewed with zoom.}
\label{fig:ego4d}
\end{figure}

\noindent \textbf{Targeting Objects in Complex Scenes.}
In ~\cref{fig:ego4d}, we present additional results of our target-aware video diffusion model applied to complex scenes such as a bike repair shop or kitchen, where describing the target with text prompts is challenging. We also present the results of the original CogVideoX~\citep{yang2024cogvideox}. Our model successfully generates videos that capture accurate interactions with the target object, even when the target occupies only a small portion of a complex scene. Note that the scene, sourced from the Ego-Exo4D dataset~\citep{grauman2024ego} is unseen during fine-tuning.

\begin{figure}
\centering
\vspace{-10px}
\includegraphics[width=1.0\columnwidth, trim={1cm 0cm 1cm 1cm}]{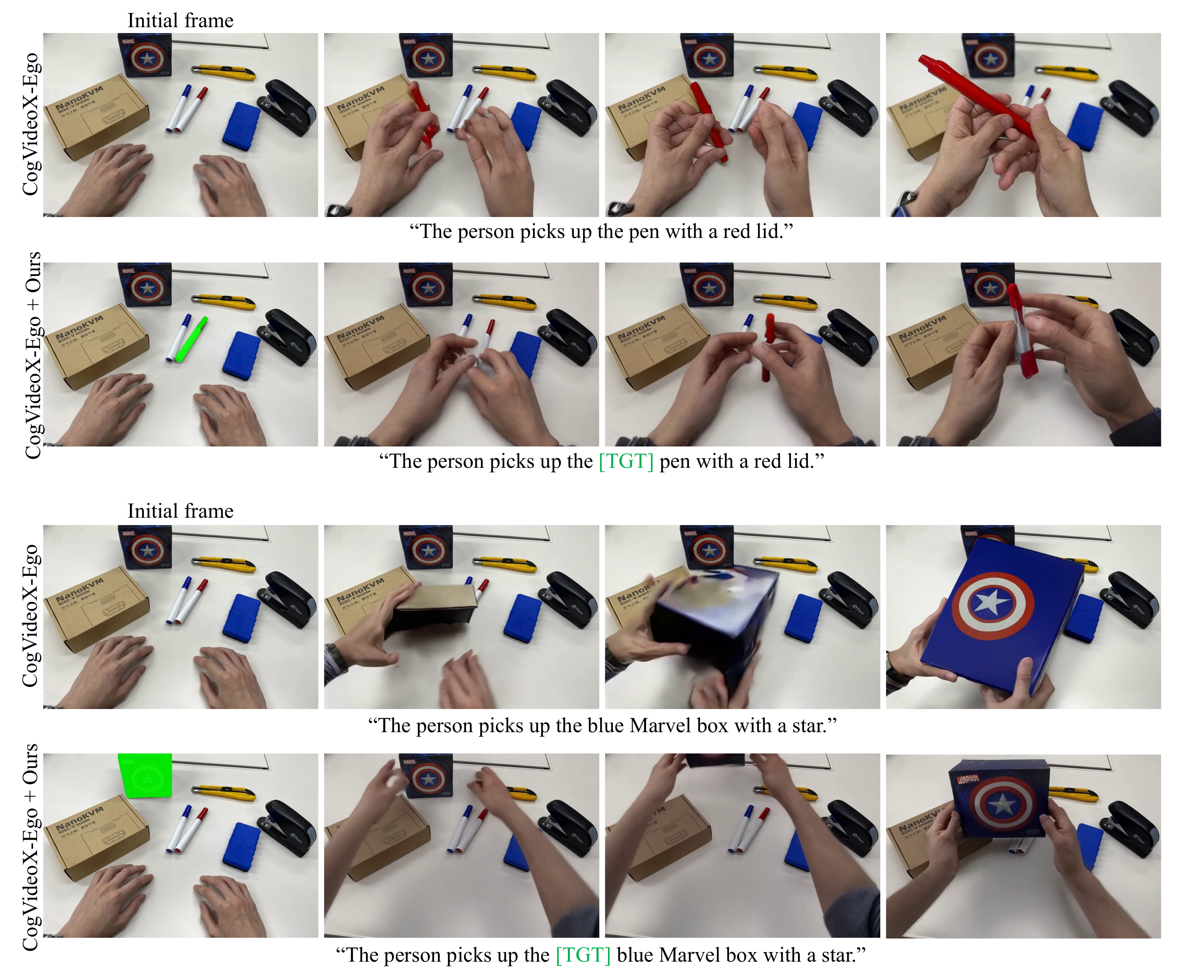}
\vspace{-10px}
\caption{\textbf{Targeting objects in egocentric view.} We fine-tune the base model on egocentric videos and then apply our target-aware LoRA module. The target awareness seamlessly generalizes to the egocentric setting without additional training, enabling precise interactions with the specified object from the actor's viewpoint.}
\label{fig:ego}
\end{figure}

\begin{figure}
\centering
\includegraphics[width=1.0\columnwidth, trim={1cm 0cm 1.5cm 1cm}]{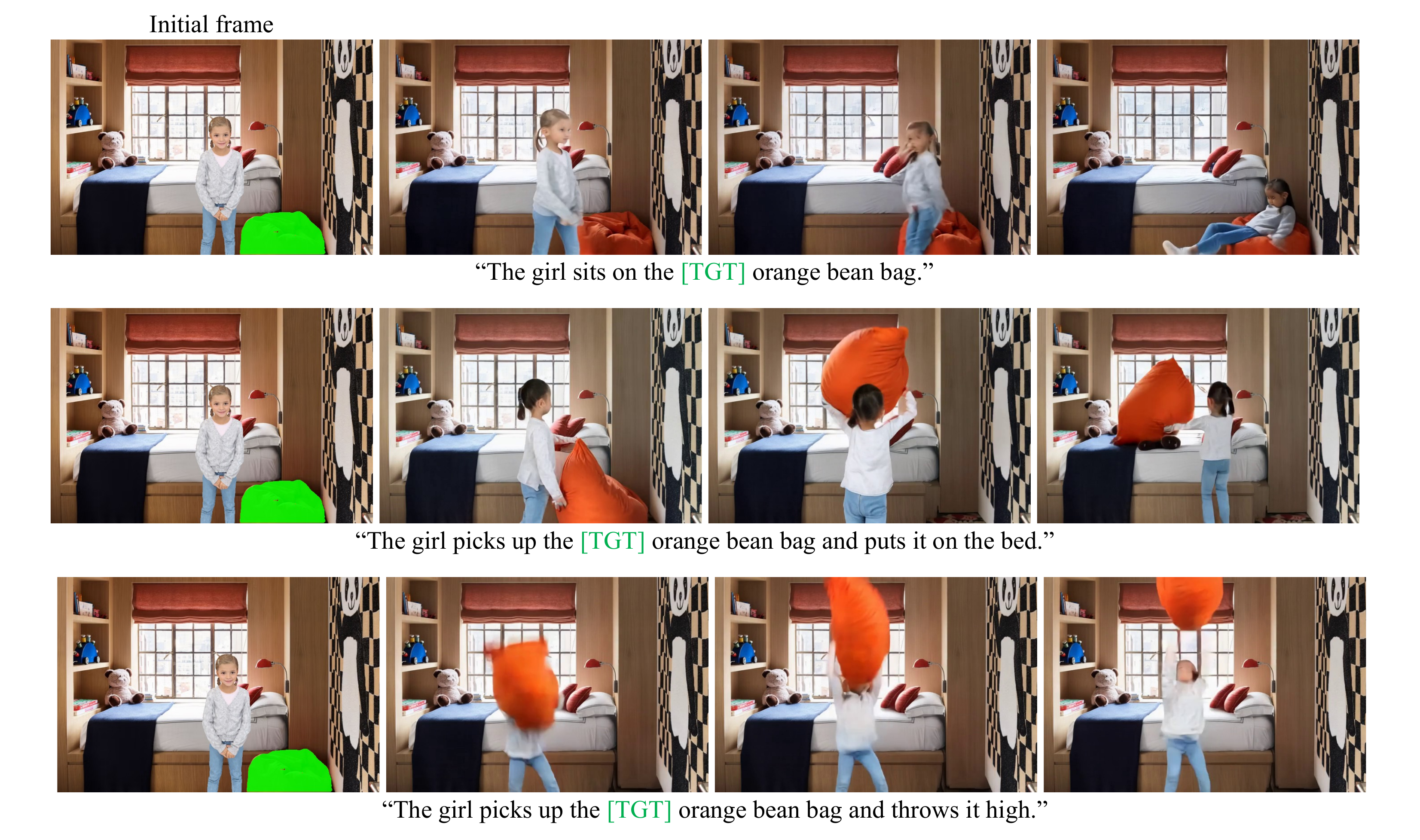}
\vspace{-10px}
\caption{\textbf{Diverse actions with the same target.} Our method can generate diverse actions with the same target using different prompts.}
\vspace{-15px}
\label{fig:various}
\end{figure}

\begin{figure}
\centering
\includegraphics[width=1.0\columnwidth, trim={0cm 0cm 0cm 0cm}]{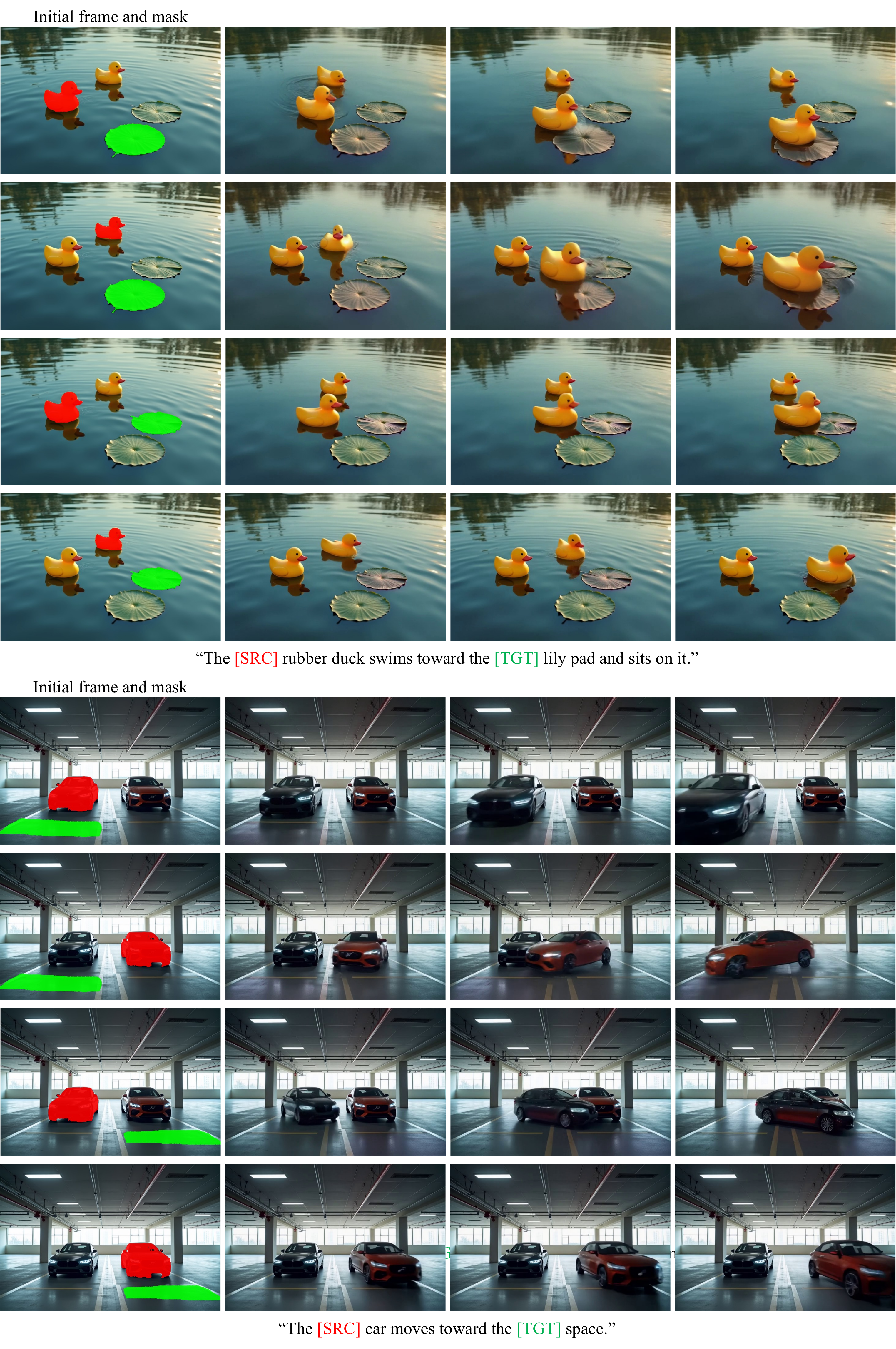}
\caption{\textbf{Control over multiple entities.} Our model can be extended to specify both the source actor and the target object using two masks.}
\label{fig:supp_twomasks}
\end{figure}

\begin{figure}
\centering
\includegraphics[width=1.0\columnwidth, trim={0cm 0cm 0cm 0cm}]{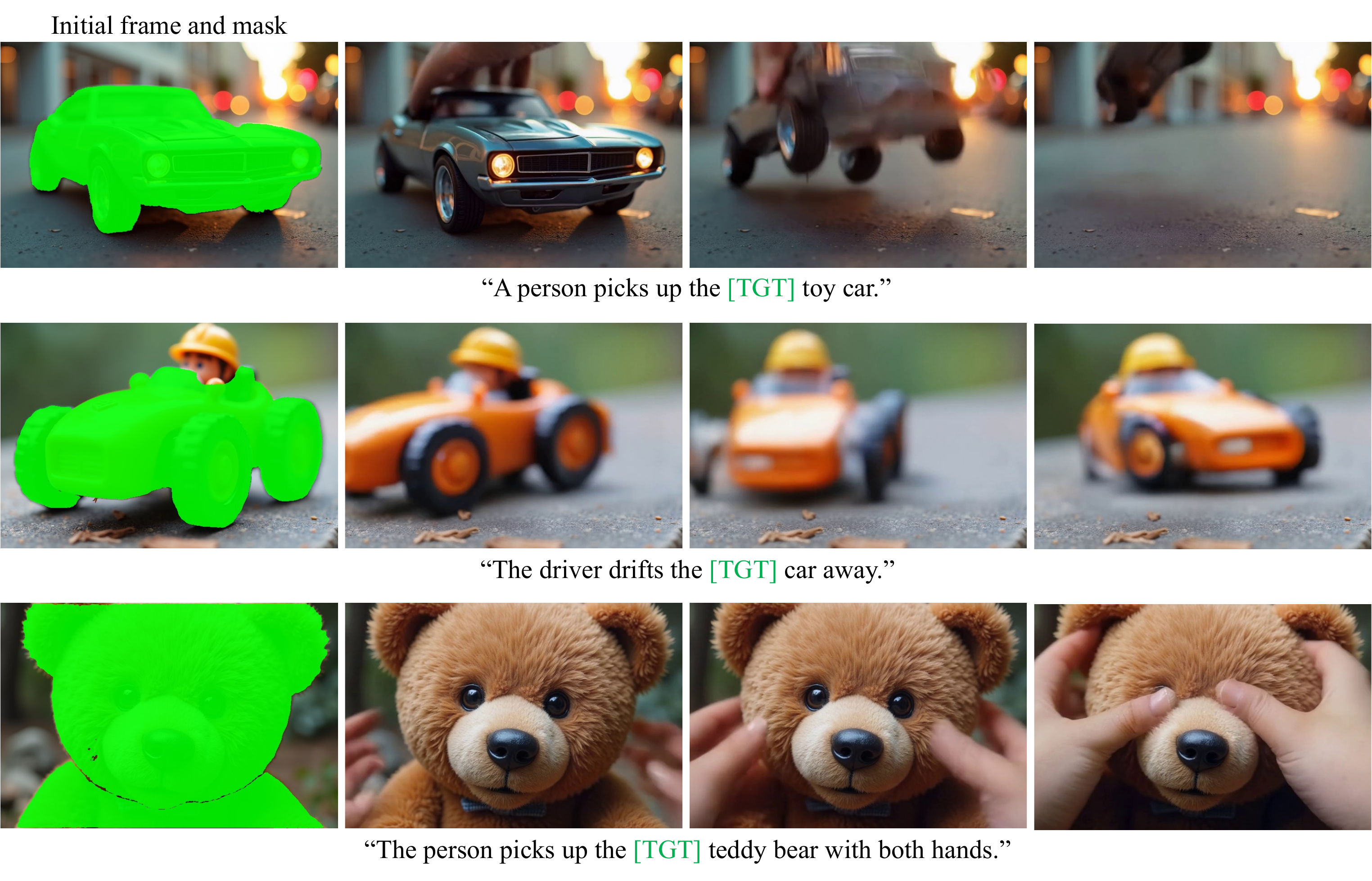}
\caption{\textbf{Large target in a scene.} Our method can handle targets that takes up a significant portion of the input frame.}
\label{fig:supp_closeups}
\end{figure}

\noindent \textbf{Egocentric View Generation.}
For robotics applications, generating videos from an egocentric perspective is particularly beneficial for capturing fine-grained actions. In~\cref{fig:ego}, we demonstrate egocentric video generation using our approach. To better adapt the base model for this setting, we first fine-tune it on egocentric videos of the EgoIT-99 dataset~\citep{yang2025egolifeegocentriclifeassistant}. We then apply our target-aware LoRA module to the fine-tuned model (CogVideoX-Ego) without any additional training. Notably, the target awareness generalizes seamlessly to the fine-tuned model.

\begin{figure*}
\centering
\includegraphics[width=1.0\linewidth, trim={0cm 0cm 0cm 1cm}]{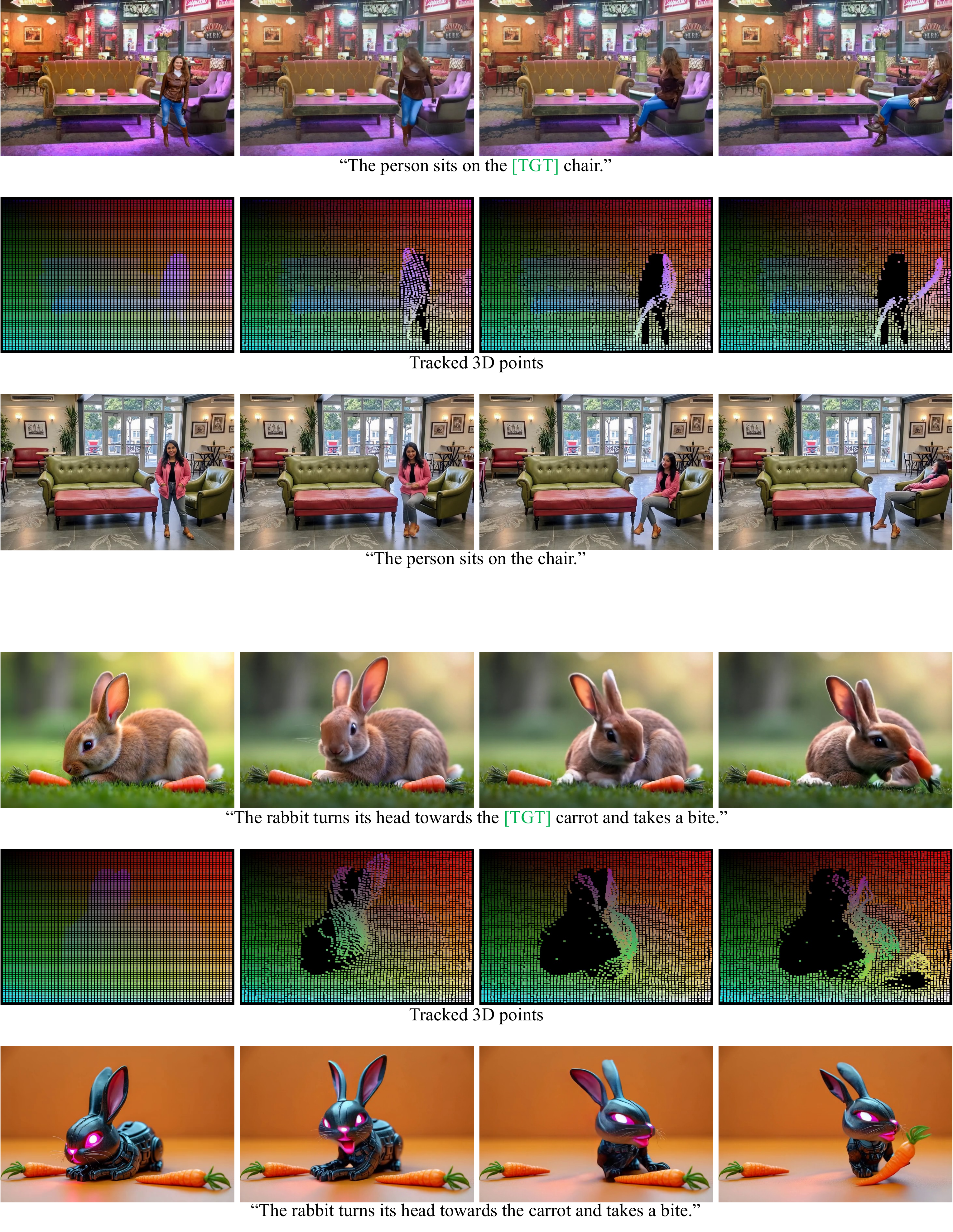}
\vspace{-10pt}
\caption{\textbf{Acting as a source video.} Our outputs can provide sufficient motion data for existing controllable video generation methods~\citep{gu2025diffusion} that require dense structural conditions over frames.}
\label{fig:das}
\end{figure*}

\noindent \textbf{Providing Motion.}
The output videos produced with our method, where the actor precisely interacts with the target, can serve as a source of motion data for existing controllable video generation approaches~\citep{gu2025diffusion}. Diffusion as Shader~\citep{gu2025diffusion} uses 3D tracking video~\citep{SpatialTracker} of a source clip to condition the motion in generated videos such that they follow the motion of the source. However, acquiring appropriate source videos for complex motions, such as human-object interactions or robot manipulations, is often challenging. In such cases, our method can generate the desired interactions and provide sufficient motion conditions, as demonstrated in ~\cref{fig:das}. We use Flux~\citep{Flux} with a canny-edge ControlNet~\citep{zhang2023controlnet} to generate the initial frames for running Diffusion as Shader.

\noindent \textbf{Diverse Actions with the Same Target.} We demonstrate in~\cref{fig:various} that our method can generate diverse actions with the same target by varying the prompt. The action quality depends on the base model, while the target awareness is applied via our method.

\noindent \textbf{Specifying Both the Actor and the Target.} ~\cref{fig:supp_twomasks} is an extended figure of ~\cref{fig:two_masks} in the main paper. As demonstrated, our model can be extended to specify both the source actor and the target object using two masks for interaction.

\noindent \textbf{Scenes with Large Targets.} We demonstrate in~\cref{fig:supp_closeups} that our method can handle targets that takes up a large portion of the input frame. Our model continues to interpret the mask with the [TGT] specification correctly and generates interactions focused on the large target region.

\begin{figure*}
\centering
\includegraphics[width=1.0\linewidth, trim={0cm 0cm 0cm 0cm}]{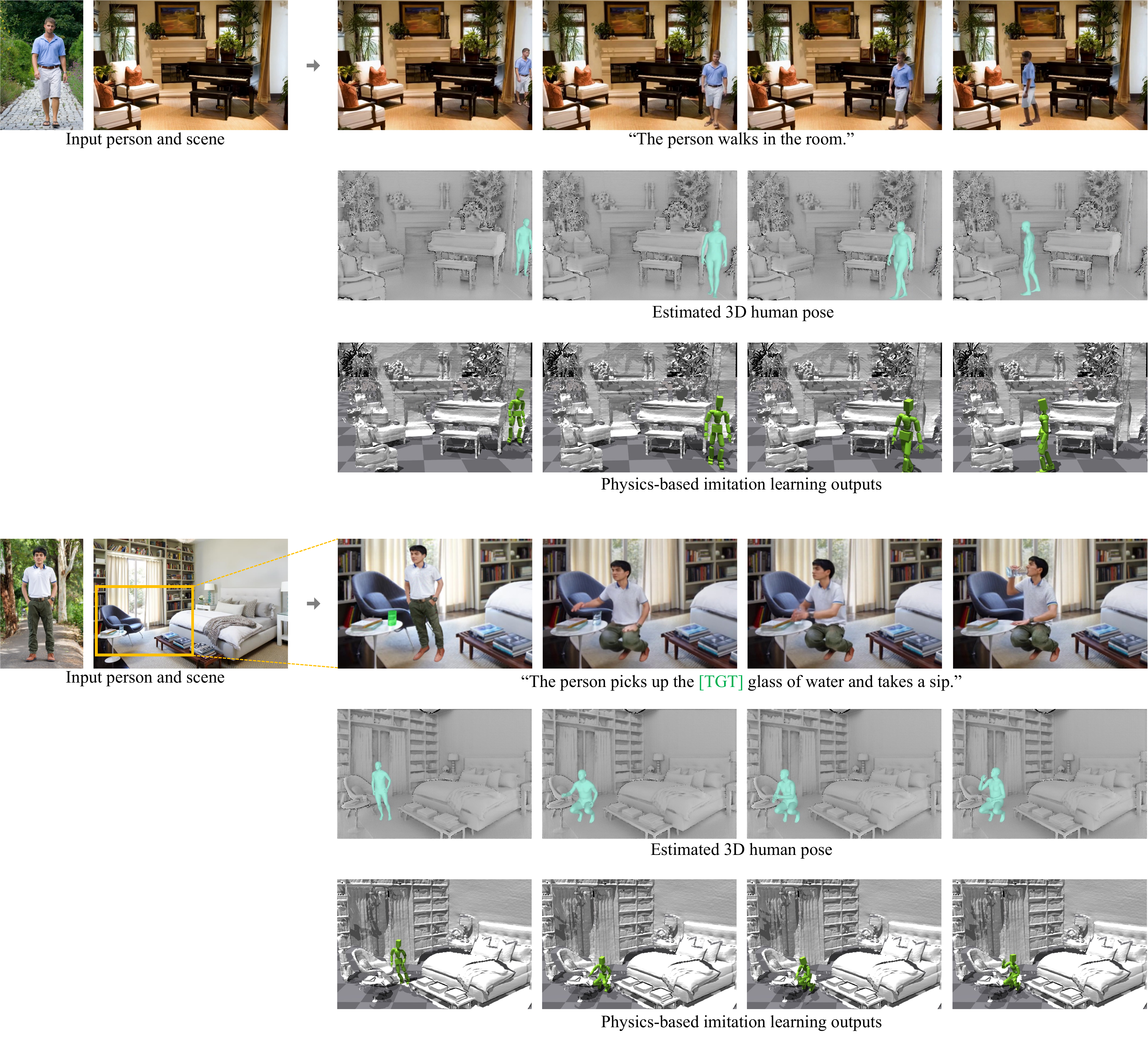}
\vspace{-10pt}
\caption{\textbf{Applications.} Given images of a person and a scene, we perform 3D insertion of the person into the scene and render them together to produce frames for video diffusion input. We interpolate synthesized initial and final frames to generate locomotion contents and utilize our Target-Specified video diffusion model to generate action and manipulation contents. We further demonstrate that extracted 3D human poses from our generated contents can be used as training data for physics-based imitation learning.}
\vspace{-10pt}
\label{fig:app_supp}
\end{figure*}

\noindent \textbf{Applications.}
~\cref{fig:app_supp} is an extended figure of ~\cref{fig:app2} in the main paper, demonstrating the downstream applications of our method. Given images of a person and a scene, we first synthesize human-inserted images as described in ~\cref{sec:app1}. As mentioned in the main paper, to achieve human navigation contents, we use a frame interpolation video diffusion model~\citep{CogvideoXInterpolation} to interpolate two synthesized images where the person is inserted in different positions of the scene. For human action or manipulation content, we similarly start from a human-inserted image and utilize our model to achieve precise interaction with the target. From the generated contents, we extract the 3D human pose sequences~\citep{shen2024gvhmr} and use them to learn a policy via physics-based imitation learning~\citep{wang2023physhoi} given a target motion.

\begin{figure}
\centering
\includegraphics[width=1.0\columnwidth, trim={1cm 0cm 1cm 0cm}]{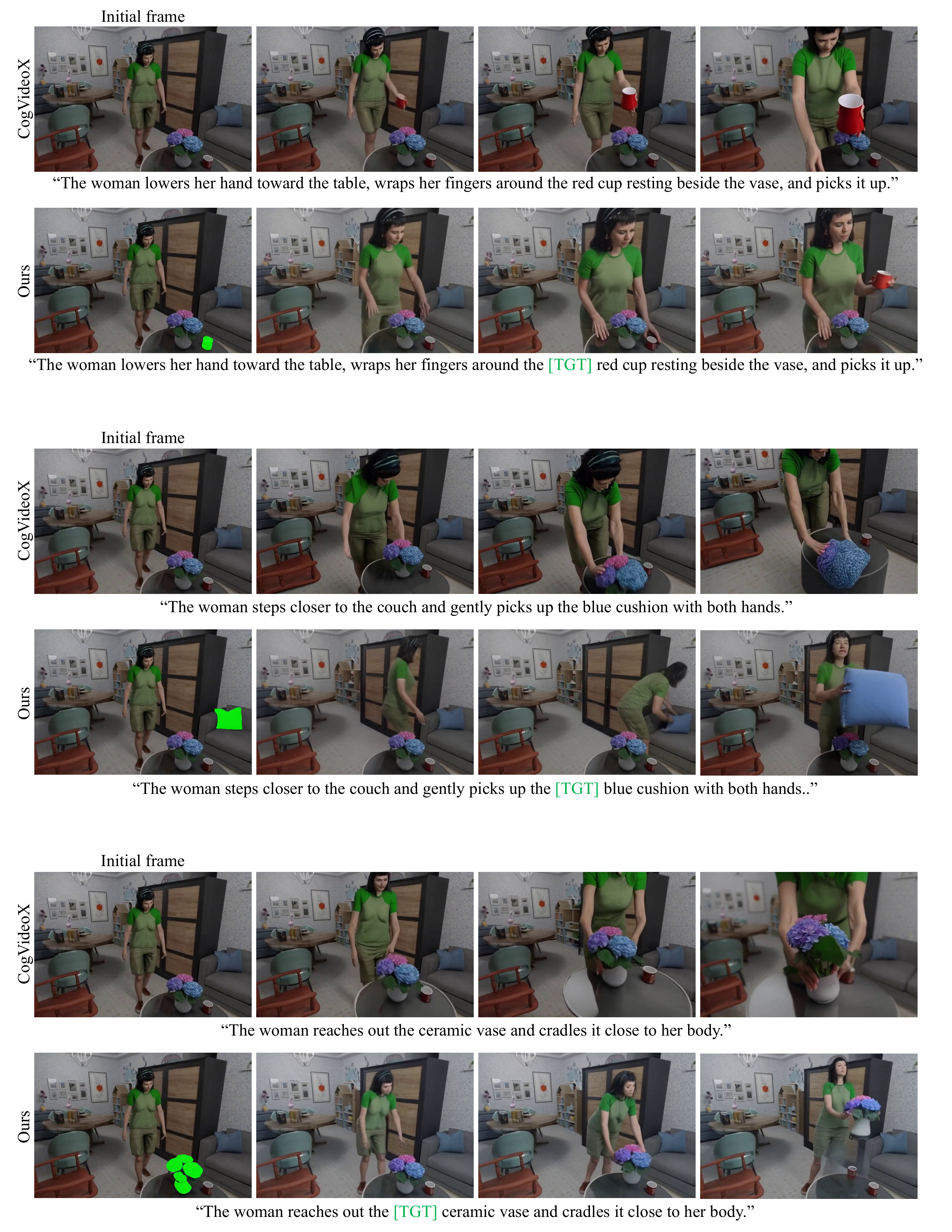}
\caption{\textbf{Additional qualitative comparison on target alignment.} We compare results of original CogVideoX~\citep{yang2024cogvideox} and our target-aware model. Our model successfully generates videos where the actor interacts accurately with the desired target. The target is colored in green every second row.}
\label{fig:extra_1}
\end{figure}

\begin{figure}
\centering
\includegraphics[width=1.0\columnwidth, trim={1cm 0cm 1cm 0cm}]{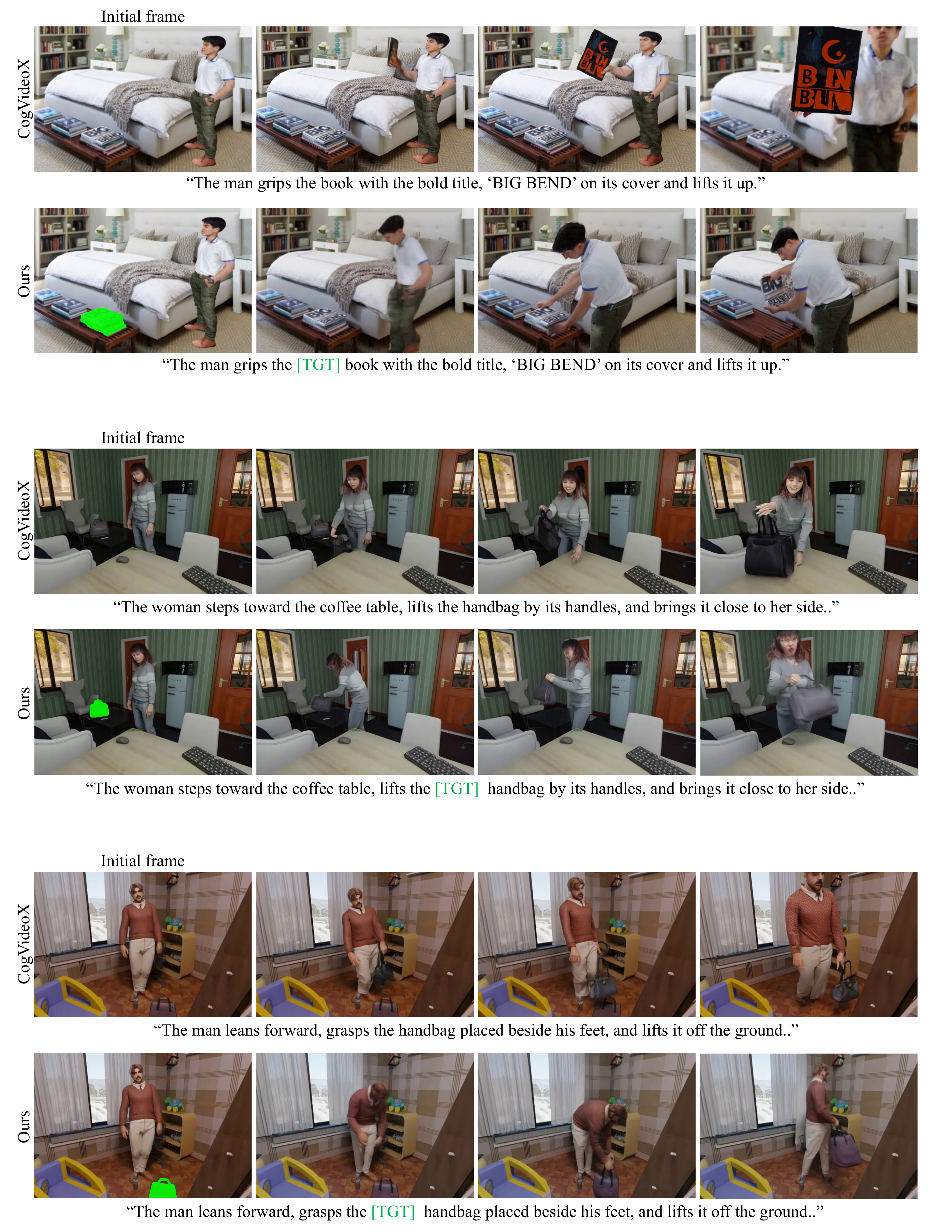}
\caption{\textbf{Additional qualitative comparison on target alignment.} We compare results of original CogVideoX~\citep{yang2024cogvideox} and our target-aware model. Our model successfully generates videos where the actor interacts accurately with the desired target. The target is colored in green every second row.}
\label{fig:extra_2}
\end{figure}

\begin{figure}
\centering
\includegraphics[width=1.0\columnwidth, trim={1cm 0cm 1cm 0cm}]{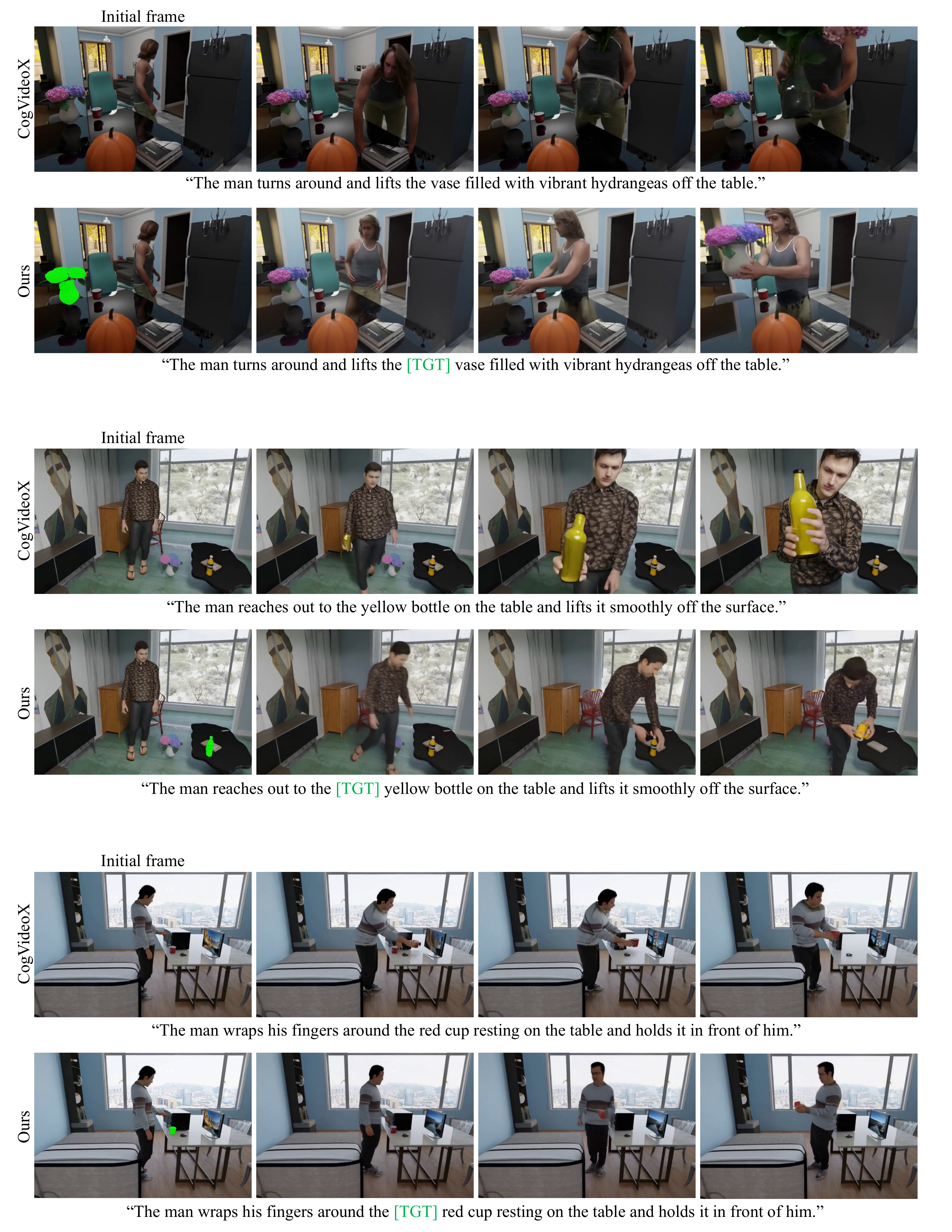}
\caption{\textbf{Additional qualitative comparison on target alignment.} We compare results of original CogVideoX~\citep{yang2024cogvideox} and our target-aware model. Our model successfully generates videos where the actor interacts accurately with the desired target. The target is colored in green every second row.}
\label{fig:extra_3}
\end{figure}

\begin{figure}
\centering
\includegraphics[width=1.0\columnwidth, trim={1cm 0cm 1cm 0cm}]{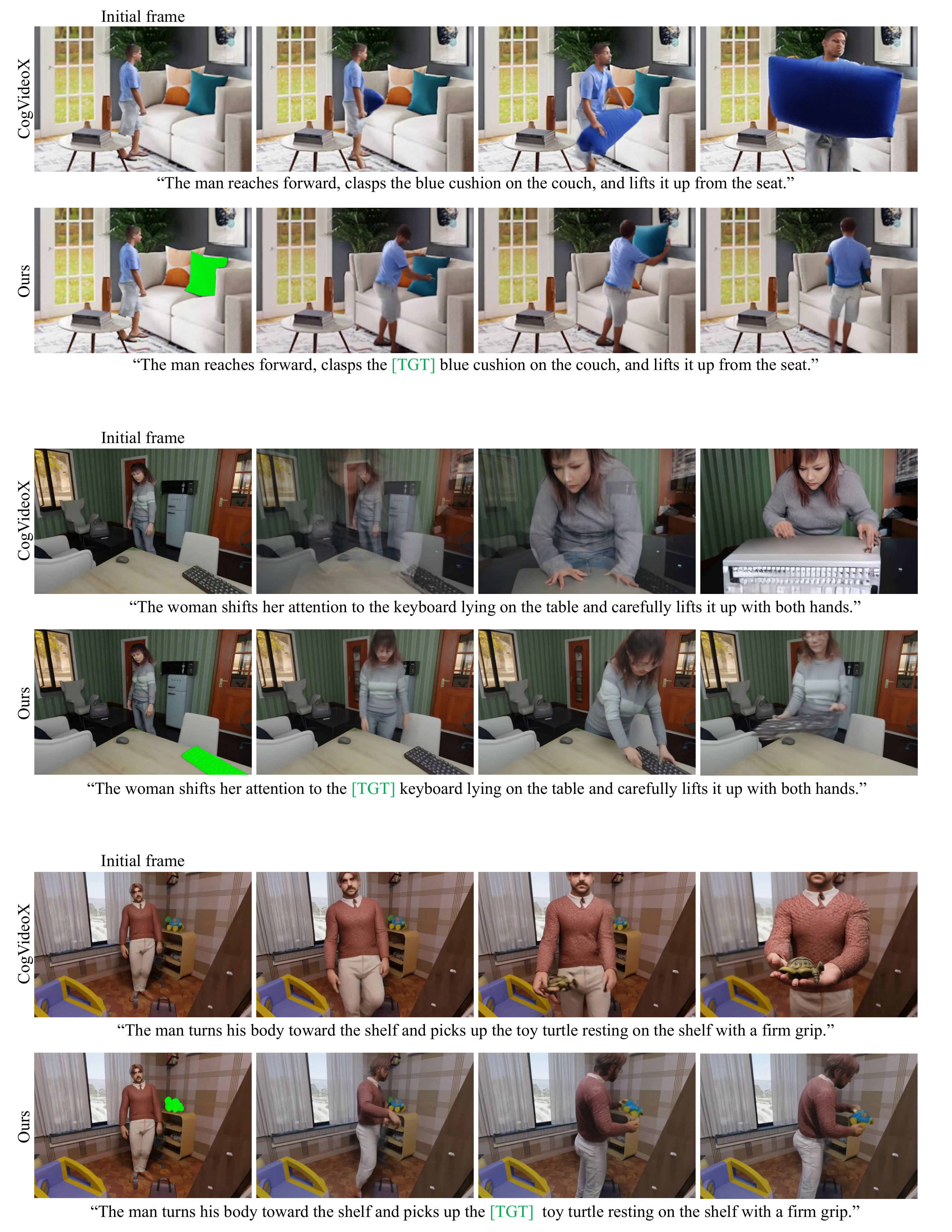}
\caption{\textbf{Additional qualitative comparison on target alignment.} We compare results of original CogVideoX~\citep{yang2024cogvideox} and our target-aware model. Our model successfully generates videos where the actor interacts accurately with the desired target. The target is colored in green every second row.}
\label{fig:extra_4}
\end{figure}

\begin{figure}
\centering
\includegraphics[width=1.0\columnwidth, trim={1cm 0cm 1cm 0cm}]{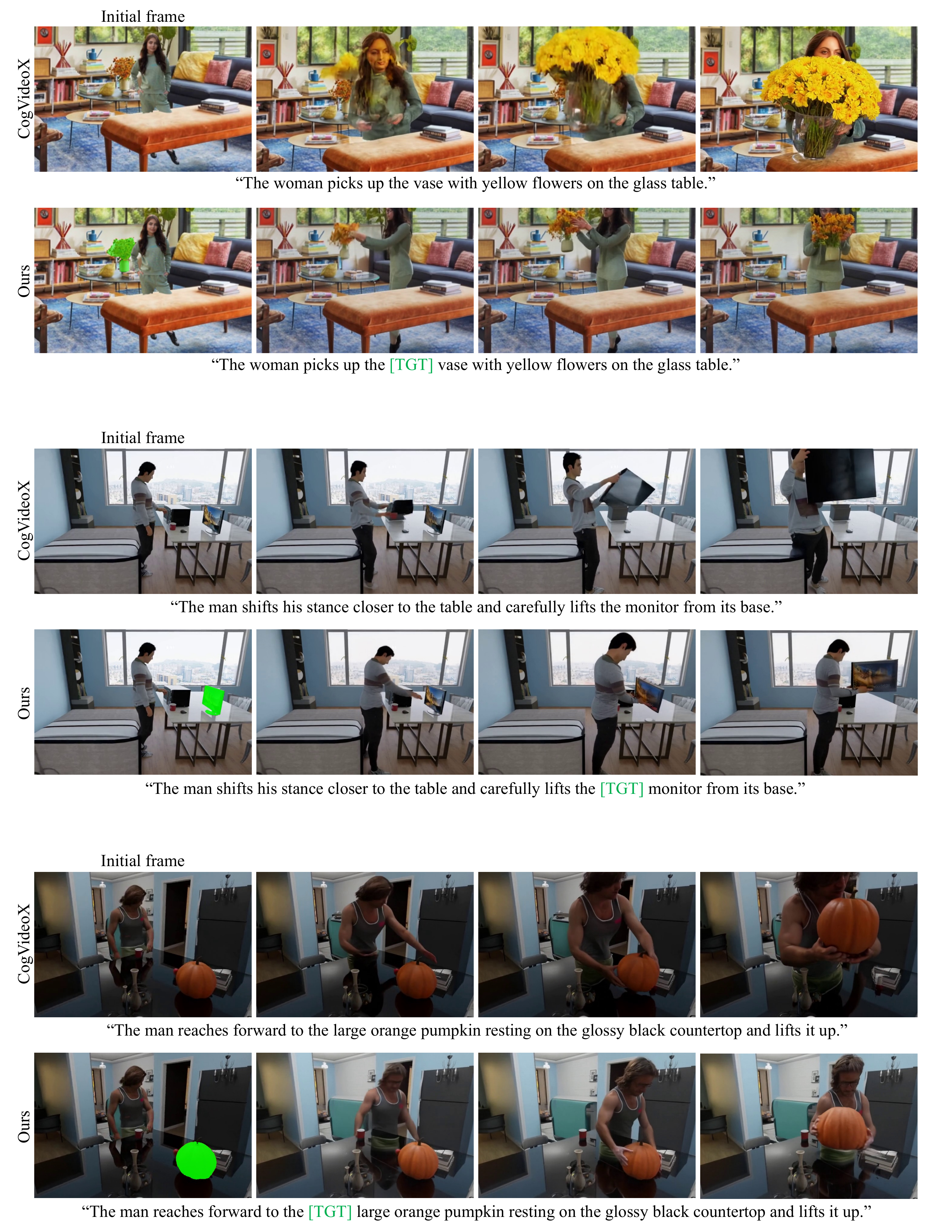}
\caption{\textbf{Additional qualitative comparison on target alignment.} We compare results of original CogVideoX~\citep{yang2024cogvideox} and our target-aware model. Our model successfully generates videos where the actor interacts accurately with the desired target. The target is colored in green every second row.}
\label{fig:extra_5}
\end{figure}

\begin{figure}
\centering
\includegraphics[width=1.0\columnwidth, trim={1cm 0cm 1cm 0cm}]{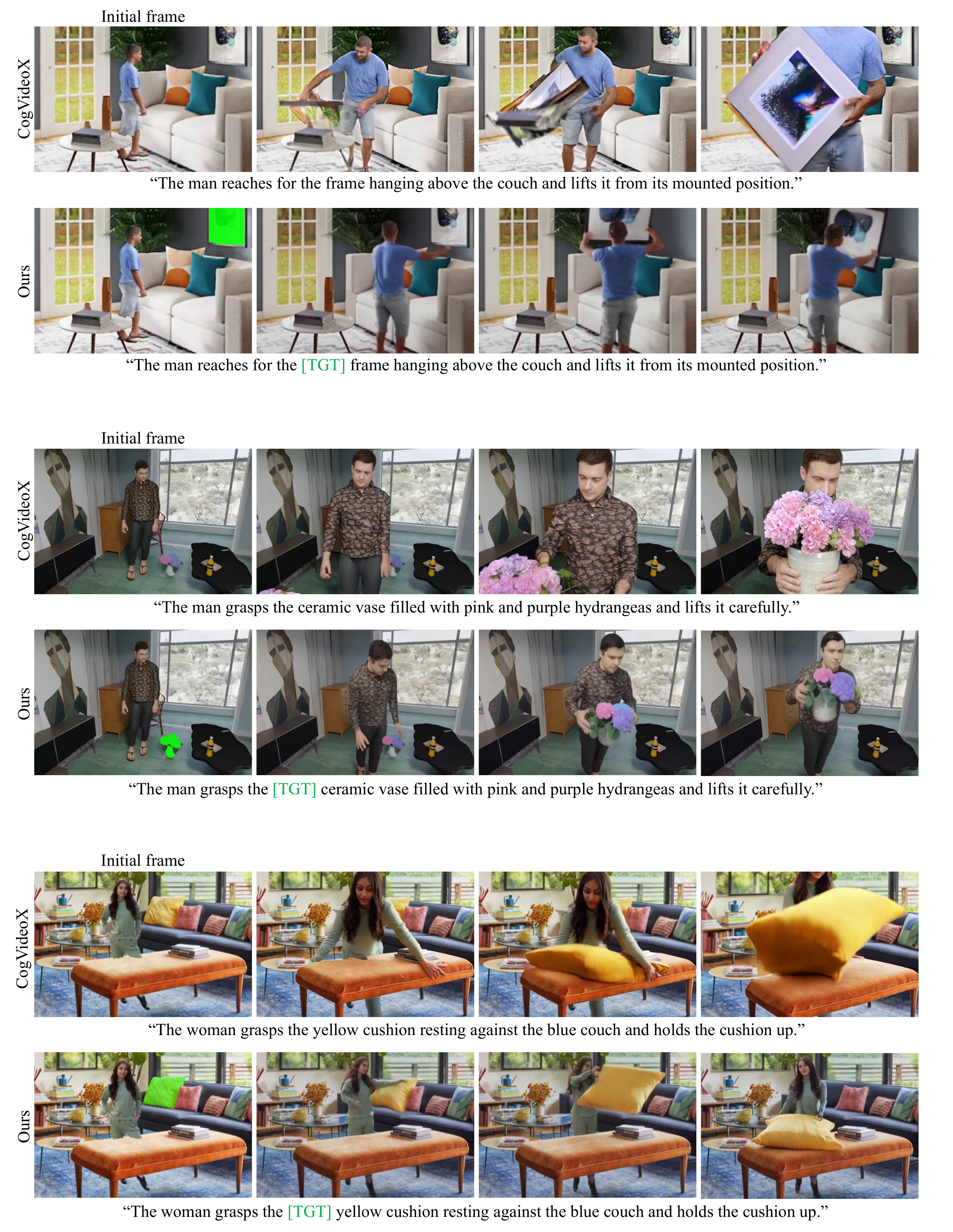}
\caption{\textbf{Additional qualitative comparison on target alignment.} We compare results of original CogVideoX~\citep{yang2024cogvideox} and our target-aware model. Our model successfully generates videos where the actor interacts accurately with the desired target.}
\label{fig:extra_6}
\end{figure}

\noindent \textbf{Target Alignment.}
~\crefrange{fig:extra_1}{fig:extra_6} are extended figures of ~\cref{fig:target}, demonstrating that our model enables accurate interactions between the actor and the target. We also provide generation outputs of the vanilla CogVideoX~\citep{yang2024cogvideox} for comparison.

\section{Discussions}
\label{sec:discussion_supp}

\subsection{Masks}
\noindent \textbf{Automatic Acquisition of Masks.} Although fully automating the inference pipeline is not the main focus of our work, we construct an automated pipeline to generate target masks from an input image and prompt, and evaluate its effectiveness. Specifically, we use GPT-4o~\citep{hurst2024gpt} to identify the target object noun from the input, and Grounded-SAM~\citep{ren2024grounded}, an open-vocabulary segmentation tool, to obtain the corresponding mask. The pipeline runs in approximately 15 seconds on a single RTX 3090 and achieves 93.14\% IoU accuracy on our evaluation set. We then use the automatically generated mask as input to our model and find that they achieve comparable performance, as presented in ~\cref{tab:mask_gsam}. For this experiment and the following mask ablations, we use a subset of 50 images from our evaluation set and similarly generate 5 videos per image.
\begin{table}

\centering
\small{

\begin{tabular}{lcc}
\toprule
       & Contact Score$\uparrow$ & Video Quality$\uparrow$ \\
\midrule
Original & 0.896 & 0.812  \\
Automatic & 0.864 & 0.810 \\

\bottomrule
\end{tabular}
}
\caption{\textbf{Automatic pipeline.} Our method stays robust even when target masks come from an automated pipeline, demonstrating that it does not rely on high-quality manual segmentation.}

\label{tab:mask_gsam}
\end{table}

\subsection{Robustness to Noisy Captions}
\noindent \textbf{Training.} Our training dataset contains captions that may incorrectly describe which object the actor is interacting with, due to limitations of current video captioning models~\citep{yang2024cogvideox}. To handle this, we prepend a simple, but always true sentence, ``The person interacts with [TGT] object.'' to the generated captions as mentioned in ~\cref{sec:dataset}. This guarantees that the [TGT] token is semantically linked to the object under interaction, while the generated part of the caption mainly provides information about the actor's appearance, the scene, and coarse motions in the video. These descriptive details help preserve the priors of the pre-trained backbone during fine-tuning. To explicitly assess the role of these descriptive, but noisy details, we fine-tune the model using only the general sentence, ``The person interacts with [TGT] object.'', completely removing the captioner-generated descriptions. As demonstrated in ~\cref{tab:noisy_caption}, the variant trained with the general sentence only, maintains target awareness to some extent but exhibits degraded target alignment and video quality compared to the full setting with descriptions. This indicates that, despite their noise, automatically generated captions still contribute for the model to be target-aware while maintaining its priors.
\begin{table}

\centering
\small{

\begin{tabular}{lcc}
\toprule
       & Contact Score$\uparrow$ & Video Quality$\uparrow$ \\
\midrule
General sentence only & 0.705 & 0.760  \\
General sentence with noisy descriptions & \textbf{0.878} & \textbf{0.807} \\

\bottomrule
\end{tabular}
}
\caption{\textbf{Effect of removing captioner-generated descriptions during training.} Despite their noise, automatically generated captions help preserve the priors of the pre-trained backbone during fine-tuning.}

\label{tab:noisy_caption}
\end{table}

\begin{figure}
\centering
\includegraphics[width=0.9\columnwidth, trim={0cm 0cm 0cm 0cm}]{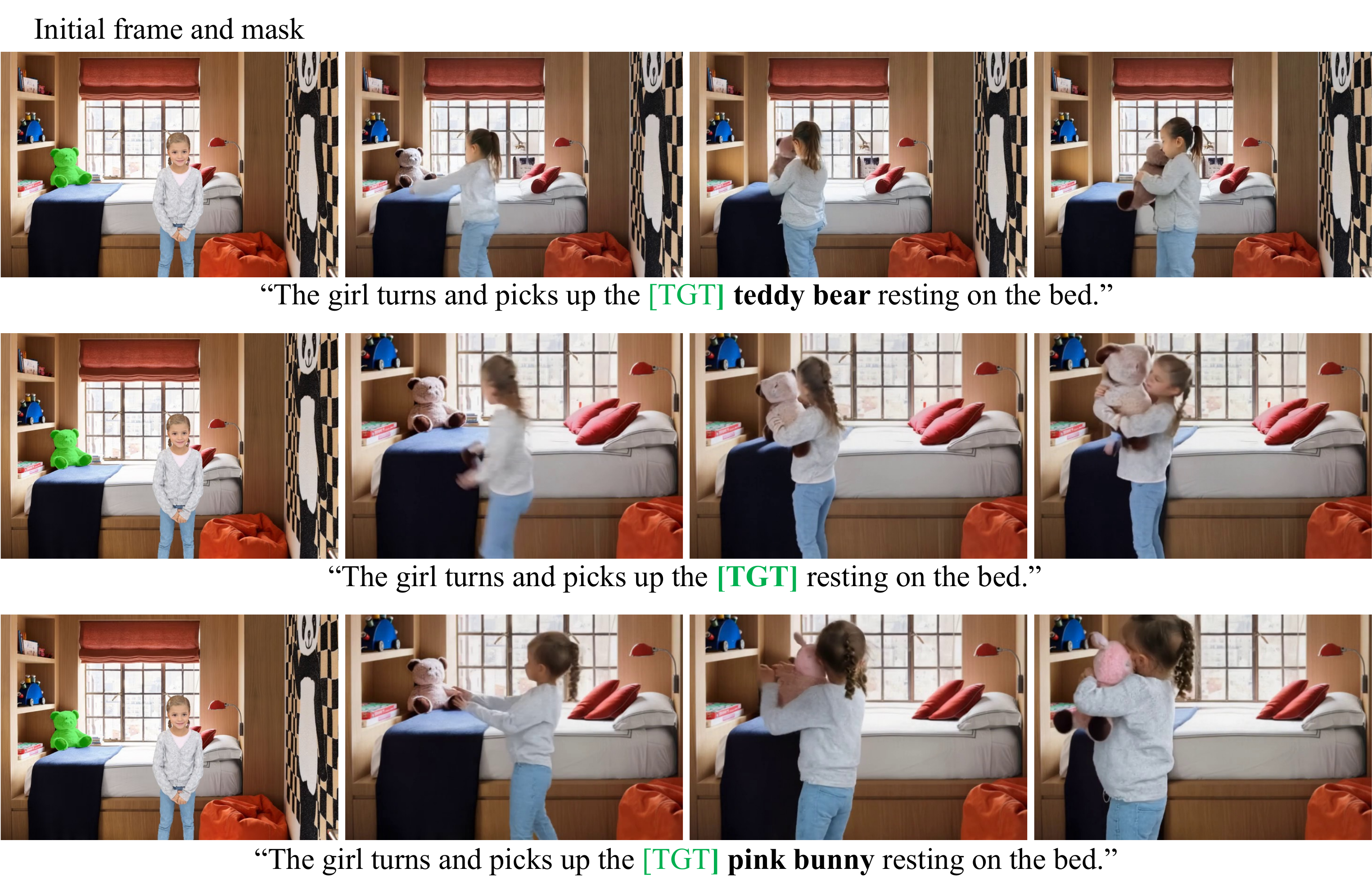}
\caption{\textbf{Inference with perturbed prompts.} Our model continues to generate correct interactions with the specified target even when the noun following [TGT] is removed or replaced, demonstrating robustness to noisy captions at inference.}
\label{fig:noisy_caption}
\end{figure}

\noindent \textbf{Inference.} We evaluate our model's robustness to noisy captions during inference by corrupting the textual descriptions while keeping the same target mask. In particular, we test (1) omitting the noun after [TGT], and (2) replacing the noun after [TGT] with an object name that does not correspond to the target. Qualitative results in \cref{fig:noisy_caption} show that the model continues to interact with the masked target region under both perturbations. This further confirms that spatial grounding primarily arises from the [TGT] token and its alignment with the mask, rather than the specific noun used in the prompt.

\subsection{Extended Contact Score}
We propose \textit{Contact Score} as our primary metric because it directly reflects the core objective of target-aware generation: whether the model correctly interacts with the specified target. In this task setting, even brief contact with the correct target is more important than a long, semantically rich interaction with a hallucinated object. At the same time, single-frame overlap can be overly permissive in rare cases (e.g., accidental or spurious touches). To provide a more thorough evaluation, we introduce two complementary metrics.

We define \textit{Contact Score (kf)} as a straightforward extension of Contact Score that counts an interaction as successful only if the detected contact region overlaps the target mask for at least $k$ consecutive frames (we report $k = 2$ and $k = 3$). This stricter criterion explicitly discounts accidental or transient touches. We note that \textit{Contact Score (kf)} can sometimes underestimate correct interactions when the target moves outside its initial mask region within a few frames: in such cases, the contact may naturally leave the original target mask even though the interaction is correct.

We further introduce \textit{Interaction Score} to require not only correct contact but also nontrivial target motion. Specifically, we track points sampled from the target-mask region of the first frame across the video using CoTracker~\citep{karaev23cotracker}. We count an interaction as accurate only if (1) the original Contact Score condition holds (the contact region overlaps the target mask in at least one frame) and (2) the mean displacement of the tracked points exceeds a threshold (10 pixels in our experiments; static videos typically yield mean displacement below 0.5 pixels). This jointly enforces that the actor contacts the correct target and that the target undergoes meaningful motion. It is important to retain the original Contact Score because displacement alone can be triggered without correct interaction (e.g., target motion without being touched). A limitation of \textit{Interaction Score} is that it can underestimate interactions where the target remains largely static (e.g., pushing on a fixed object or sliding a hand over a surface), since the displacement term remains small even when the interaction is correct. As demonstrated in ~\cref{tab:target_metrics_extended}, our method consistently outperforms baselines for all metrics.

\begin{table}[t]
\centering
\resizebox{\columnwidth}{!}{  
\begin{tabular}{lcccc}
\toprule
& \multicolumn{4}{c}{\textbf{Targeting Quality}} \\
\cmidrule(lr){2-5}
& Contact Score$\uparrow$ & Contact Score (2f)$\uparrow$ & Contact Score (3f)$\uparrow$ & Interaction Score$\uparrow$ \\
\midrule
CogVideoX & 0.560 & 0.446 & 0.377 & 0.474 \\
CogVideoX w.data & 0.638 & 0.504 & 0.393 & 0.540 \\
Attn. Mod. & 0.546 & 0.399 & 0.340 & 0.455 \\
Ours & \textbf{0.878} & \textbf{0.770} & \textbf{0.693} & \textbf{0.783} \\
\bottomrule
\end{tabular}
}
\caption{\textbf{Extended targeting metrics.} In addition to Contact Score, we report stricter temporal variants (Contact Score 2f/3f) and Interaction Score, which jointly requires correct target contact and nontrivial target motion.}
\label{tab:target_metrics_extended}
\end{table}

\subsection{Attention}
\label{sec:attention_supp}
\begin{figure}
\centering
\includegraphics[width=0.68\columnwidth, trim={0cm 0cm 0cm 0cm}]{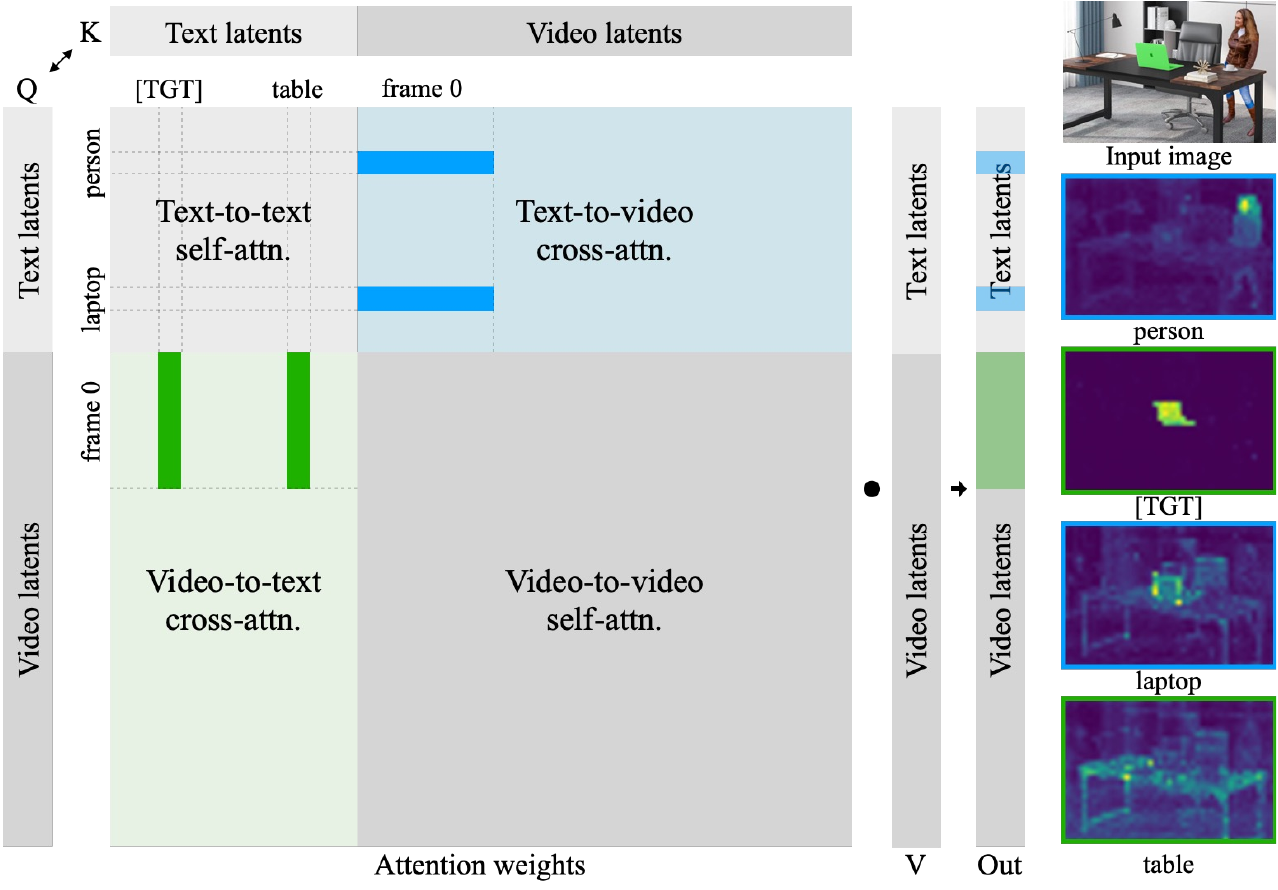}
\caption{\textbf{Attention mechanisms in MM-DiTs.} Attentions of MM-DiTs can be divided into text-to-text self-attention, text-to-video (T2V) cross-attention, video-to-text (V2T) cross-attention, and video-to-video self-attention. Since V2T cross-attention weights directly influence the values of video latents, we apply our cross-attention loss on V2T cross-attention regions.}
\vspace{10pt}
\label{fig:mmdit}
\end{figure}

\noindent \textbf{Attention Mechanisms in MM-DiTs.}
State-of-the-art diffusion models~\citep{Flux, esser2024scaling, kong2024hunyuanvideo} including our base model~\citep{yang2024cogvideox}, utilize multi-modal diffusion transformers (MM-DiTs)~\citep{esser2024scaling} for denoising. In MM-DiTs, text and video latents are concatenated into a single sequence, and attention computations are performed over the combined representation. Specifically, given query features $\mathbf{Q}$, key features $\mathbf{K}$ and value features $\mathbf{V}$, each obtained by passing the combined representation through separate linear layers, the attention in a transformer block is computed as,
\begin{equation}
\label{eq:attn_mech}
\begin{split}
    \text{Attn}(\mathbf{Q}, \mathbf{K}, \mathbf{V})
    = \text{Softmax}(\frac{\mathbf{QK}^T}{\sqrt{d}}) \mathbf{V},
\end{split}
\end{equation}
where $d$ is the channel dimension of $\mathbf{Q}$. The resulting $\text{Attn}(\mathbf{Q}, \mathbf{K}, \mathbf{V})$  is normalized, projected through linear layers, and used as the combined representation for the next transformer block. The attention weights are formed by $\text{Softmax}(\frac{\mathbf{QK}^T}{\sqrt{d}})$, where the value at index $[i, j]$ indicates the influence of the $i$-th token on the $j$-th token. As illustrated in ~\cref{fig:mmdit}, this process results in four distinct attention regions: text-to-text self-attention, text-to-video (T2V) cross-attention, video-to-text (V2T) cross-attention, and video-to-video self-attention.

As discussed in the main paper, while both T2V and V2T cross-attention maps encode semantic information, we find that applying our loss to the V2T cross-attention is more effective for enhancing target awareness. As demonstrated in ~\cref{fig:mmdit}, V2T cross-attention weights directly influence the video latents during the dot product computation of the attention weights and value features, whereas T2V cross-attention weights primarily affect the text latents. Although the influenced text latents can affect subsequent V2T cross-attention weights through $\mathbf{QK}^T$ computation, their impact is diminished, as shown in ~\cref{tab:ablation_area} of our main paper.

\begin{table}[t]
\vspace{10pt}
\centering
\small{

\begin{tabular}{lcccc}
\toprule
       & $\lambda=10$ & $\lambda=25$ & $\lambda=50$ & $\lambda=100$ \\
\midrule
$\tau=0.80T$ & 0.493 & 0.473 & 0.527 & 0.513 \\
$\tau=0.85T$ & 0.480 & 0.507 & 0.520 & 0.480 \\
$\tau=0.90T$ & 0.533 & 0.520 & 0.573 & 0.520 \\
$\tau=0.95T$ & 0.487 & 0.533 & \textbf{0.613} & 0.553 \\

\bottomrule
\end{tabular}
}
\caption{\textbf{Contact scores for attention modulation.} We report contact scores for different combinations of attention control weights and cut-off timesteps.}

\label{tab:ablation_attn}
\end{table}

\noindent \textbf{Attention Modulation.}
Prior work on controllable text-to-image generation~\citep{hertz2022prompt, kim2023dense, ma2023directed, chen2024training, xie2023boxdiff} demonstrates that by modifying cross-attention maps during inference, it is possible to control the placement of subjects in specific regions of the output image. Cross-attention modulation is applied as follows:
\begin{equation}
\label{eq:attn_mod}
\begin{split}
    \text{CrossAttnMod}(\mathbf{Q}, \mathbf{K}, \mathbf{V})
    = \text{Softmax}(\frac{\mathbf{QK}^T + \lambda \mathbf{S}}{\sqrt{d}}) \mathbf{V},
\end{split}
\end{equation}
where $\lambda$ denotes attention control weight, $\mathbf{S}$ is the modulation term with the same dimensions as the attention maps. The modulation term $\mathbf{S}$ takes positive values within the desired region for the subject and negative values outside that region. Formally, given a bounding box $\mathbf{B}$ that specifies the desired region for the object, $\mathbf{S}$ is defined as follows:
\begin{equation}
\label{eq:attn_mod_S}
\begin{split}
    \mathbf{S}[i, j] = 
    \begin{cases}
      1 - \frac{\lVert \mathbf{B} \rVert}{\lvert \mathbf{QK}^T \rvert}, & \text{if} \enspace i \in \mathbf{B} , j \in \mathbf{P}, 
      t \geq \tau \\
      0, & \text{if} \enspace i \in \mathbf{B} , j \in \mathbf{P} , t < \tau \\
      -\infty, & \text{otherwise}
    \end{cases}
\end{split}
\end{equation}
where $\lVert \mathbf{B} \rVert$ is the size of the bounding box, $\lvert \mathbf{QK}^T \rvert$ is the number of elements in $\mathbf{QK}^T$, $\mathbf{P}$ represents the indices of prompt tokens for subjects, and $\tau$ is a cut-off timestep. Since diffusion models form the subject layout in earlier steps~\citep{xie2023boxdiff, hertz2022prompt}, the amplification is only applied in the early stage. Recently, the technique has been extended to text-to-video diffusion models, enabling control over object trajectories in generated videos by modulating attention map weights of every frame~\citep{yang2024direct, wu2024motionbooth}.

As mentioned in the main paper, we adapt this attention modulation concept as a baseline. In our setting, since trajectories for actors and targets are not available, we add the word ``target'' to the object description in the prompt and amplify cross-attention weights in target mask regions for the new keyword. This modulation should mirror our approach without additional training. Since attention modulation modifies the internal attention computation during the denoising process, it is highly sensitive to hyperparameters such as the attention control weight $\lambda$ and the cut-off timestep $\tau$. We report the contact scores of each setting in ~\cref{tab:ablation_attn}, evaluated by generating three videos per image from our evaluation dataset. The results consistently show that the scores remain low even compared to the original CogVideoX~\citep{yang2024cogvideox} without any modification. This degradation stems from the attention mechanisms in MM-DiTs: since the attention computation contains a row-wise softmax operation, modulating the cross-attention values affects the self-attention values of the video, ultimately leading to degraded output video quality and low contact scores, highlighting the necessity of our method for building target-aware models.

\subsection{Limitations and Future Work}
The quality of our generated videos is inherently constrained by existing open-sourced video models, which often produce noticeable visual artifacts when synthesizing complex appearances. Given that closed-sourced commercial models~\citep{Kling, Veo2} yield more convincing results, we expect this limitation to be alleviated as more advanced open-sourced models become available. Nevertheless, enhancing video quality by incorporating interaction motion cues~\citep{jeong2024track4gen, chefer2025videojam} could be an interesting future work.

Also, due to the static camera setting of our dataset, videos generated by our model tend to exhibit fixed camera trajectories. Given the scarcity of interaction datasets with dynamic cameras, integrating camera control techniques~\citep{wang2024motionctrl, yang2024direct, he2024cameractrl} into our model could be a possible future direction.

Another current limitation is that our architecture requires adding an extra channel per specified target, which poses scalability challenges when dealing with a large number of targets. Designing a unified model that supports an arbitrary number of target specifications in a more memory-efficient manner presents an exciting future research direction.

\begin{figure}
\centering
\includegraphics[width=1.0\columnwidth, trim={0cm 0cm 0cm 1cm}]{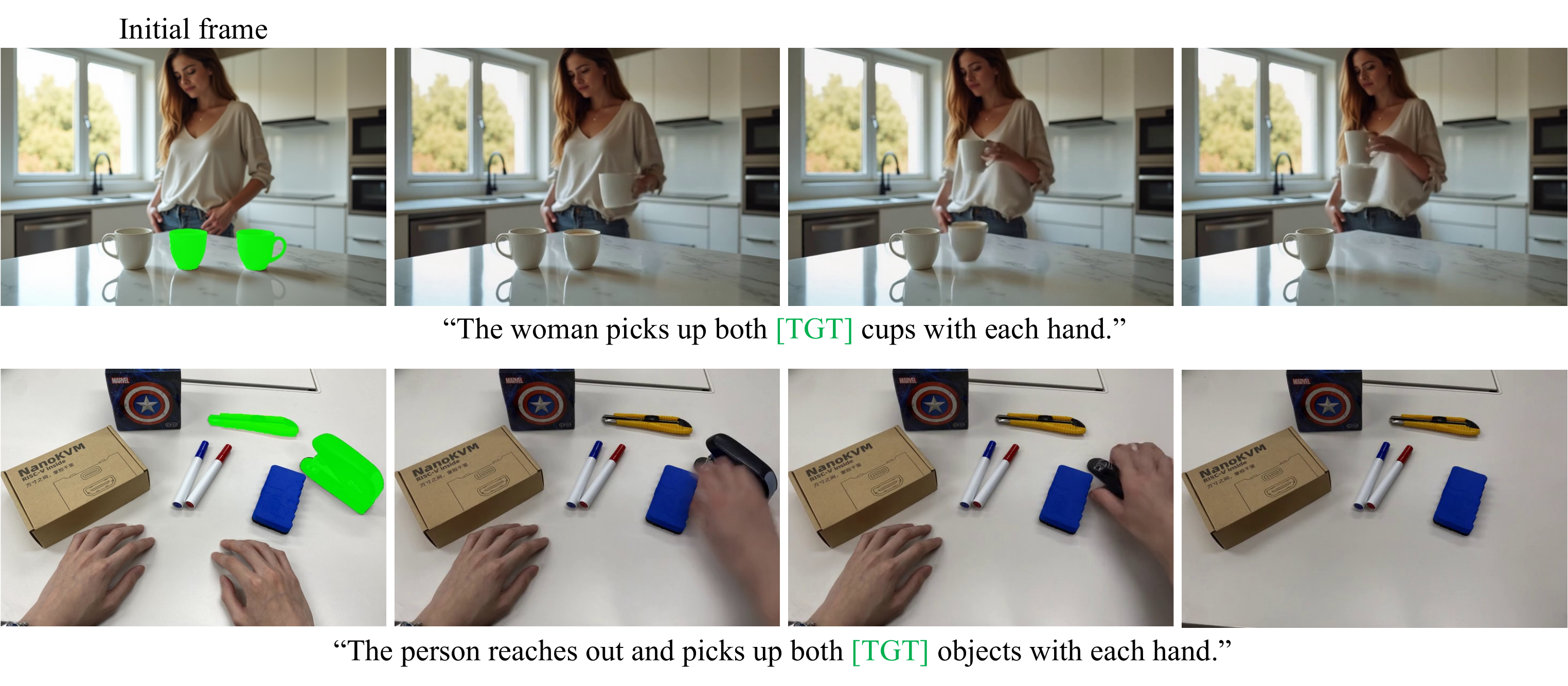}
\caption{\textbf{Failure case: targeting multiple objects with a single mask.} Since our model is trained on masks that contains a single object, it struggles when a mask spans multiple objects.}
\label{fig:limit_multi}
\end{figure}

Moreover, since our dataset contains masks that cover a single target object per video, our model mostly finds it difficult to handle cases where a single mask covers multiple objects (whether of the same or different categories), as demonstrated in ~\cref{fig:limit_multi}. Robust multi-instance handling under a single mask remains future work.

\begin{figure}
\centering
\includegraphics[width=1.0\columnwidth, trim={0cm 0cm 0cm 0cm}]{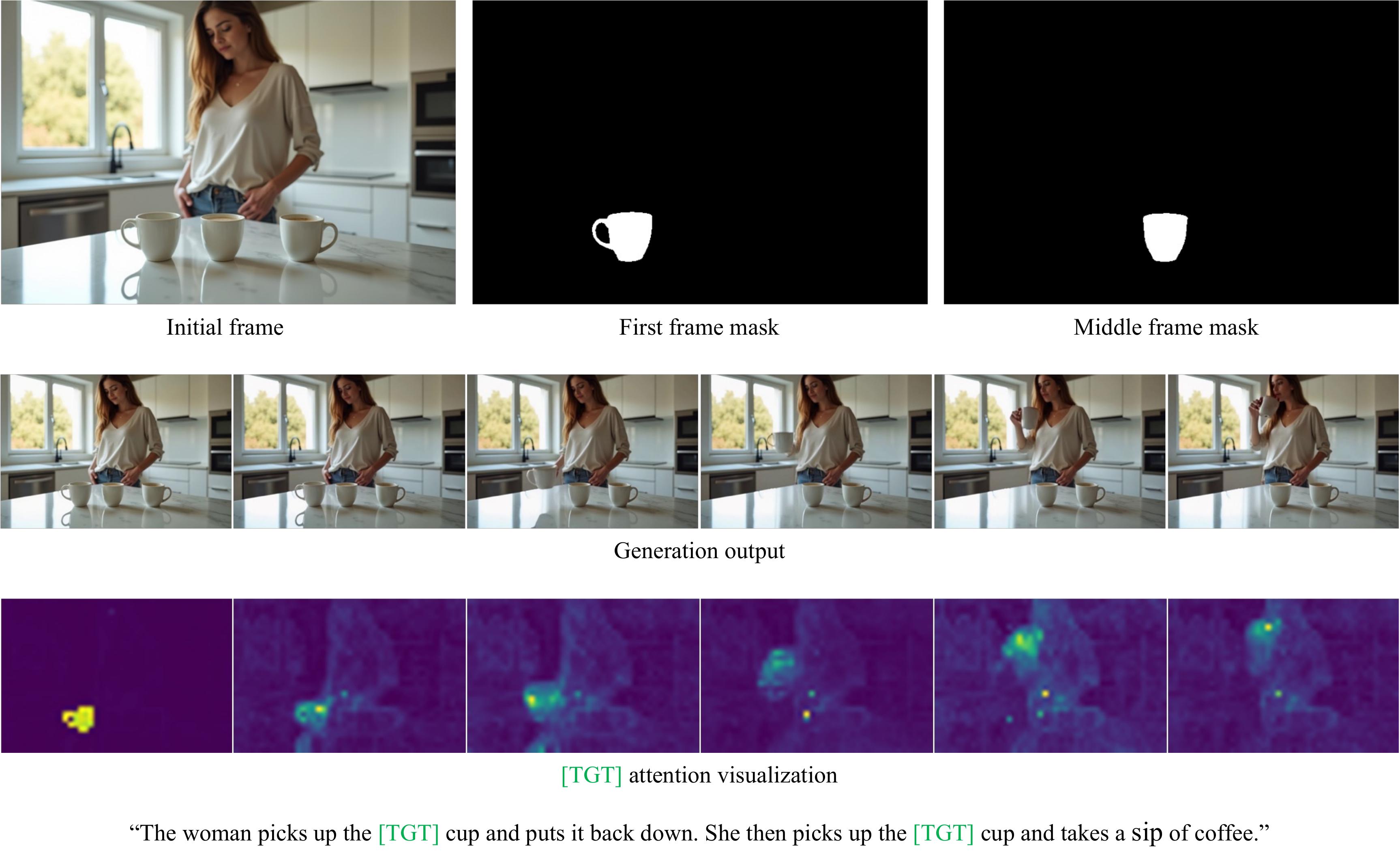}
\caption{\textbf{Failure case: targeting multiple objects over time.} Our framework assumes a single fixed target per generated video, preventing it from switching to new targets partway through the video. Even though an additional mask is provided at the middle frame, the model ignores it and continues to rely on the first-frame mask. As a result, it cannot handle sequential interactions with different targets within a single video. Nonetheless, target switching may still be achieved by generating a new video from the last frame of the previous one, enabling multi-target interactions through chained generation.}
\label{fig:limit_temp}
\end{figure}

Our current framework also assumes a fixed target specification throughout the whole video, which limits its ability to model complex interactions involving multiple objects over time ({\cref{fig:limit_temp}}). A natural next step would be to support temporal target switching, enabling fine-grained control over which object is interacted with at each point in time. This would open new possibilities for long-horizon planning and action sequencing, especially in domains like robotics or instructional video generation.

{Finally, our contact-based metrics rely on an off-the-shelf contact detector~\citep{contacthands_2020} that is designed specifically for human-object interactions and, in practice, only detects contacts involving human hands. As a result, these metrics cannot be directly applied to videos where the interacting agent is non-human (e.g., animals, robotic arms, tools). In such cases, the detector fails to produce meaningful contact predictions even when the generated interaction is qualitatively correct, leading to unreliable scores.}

\newpage



\begin{thebibliography}{117}
\providecommand{\natexlab}[1]{#1}
\providecommand{\url}[1]{\texttt{#1}}
\expandafter\ifx\csname urlstyle\endcsname\relax
  \providecommand{\doi}[1]{doi: #1}\else
  \providecommand{\doi}{doi: \begingroup \urlstyle{rm}\Url}\fi

\bibitem[Kli(2024)]{Kling}
Kling, 2024.
\newblock \url{https://kling.kuaishou.com/en}.

\bibitem[Veo(2024)]{Veo2}
Veo2, 2024.
\newblock \url{https://deepmind.google/technologies/veo/veo-2}.

\bibitem[Ajay et~al.(2023)Ajay, Han, Du, Li, Gupta, Jaakkola, Tenenbaum, Kaelbling, Srivastava, and Agrawal]{ajay2023compositional}
Anurag Ajay, Seungwook Han, Yilun Du, Shuang Li, Abhi Gupta, Tommi Jaakkola, Josh Tenenbaum, Leslie Kaelbling, Akash Srivastava, and Pulkit Agrawal.
\newblock Compositional foundation models for hierarchical planning.
\newblock \emph{NeurIPS}, 2023.

\bibitem[AlBahar et~al.(2023)AlBahar, Saito, Tseng, Kim, Kopf, and Huang]{albahar2023humansgd}
Badour AlBahar, Shunsuke Saito, Hung-Yu Tseng, Changil Kim, Johannes Kopf, and Jia-Bin Huang.
\newblock Single-image 3d human digitization with shape-guided diffusion.
\newblock In \emph{Proc. ACM SIGGRAPH Asia}, 2023.

\bibitem[Baik et~al.(2025)Baik, Kim, and Joo]{oor}
Sangwon Baik, Hyeonwoo Kim, and Hanbyul Joo.
\newblock Learning 3d object spatial relationships from pre-trained 2d diffusion models.
\newblock In \emph{Proc. ICCV}, 2025.

\bibitem[Bar et~al.(2024)Bar, Zhou, Tran, Darrell, and LeCun]{bar2024navigationworldmodels}
Amir Bar, Gaoyue Zhou, Danny Tran, Trevor Darrell, and Yann LeCun.
\newblock Navigation world models.
\newblock In \emph{Proc. CVPR}, 2024.

\bibitem[Bhatnagar et~al.(2022)Bhatnagar, Xie, Petrov, Sminchisescu, Theobalt, and Pons-Moll]{bhatnagar2022behave}
Bharat~Lal Bhatnagar, Xianghui Xie, Ilya~A Petrov, Cristian Sminchisescu, Christian Theobalt, and Gerard Pons-Moll.
\newblock Behave: Dataset and method for tracking human object interactions.
\newblock In \emph{Proc. CVPR}, 2022.

\bibitem[Black et~al.(2024)Black, Nakamoto, Atreya, Walke, Finn, Kumar, and Levine]{black2023zero}
Kevin Black, Mitsuhiko Nakamoto, Pranav Atreya, Homer Walke, Chelsea Finn, Aviral Kumar, and Sergey Levine.
\newblock Zero-shot robotic manipulation with pretrained image-editing diffusion models.
\newblock \emph{Proc. ICLR}, 2024.

\bibitem[{Black Forest Labs}(2023)]{Flux}
{Black Forest Labs}, 2023.
\newblock \url{https://blackforestlabs.ai/}.

\bibitem[Blattmann et~al.(2023)Blattmann, Dockhorn, Kulal, Mendelevitch, Kilian, Lorenz, Levi, English, Voleti, Letts, et~al.]{blattmann2023stable}
Andreas Blattmann, Tim Dockhorn, Sumith Kulal, Daniel Mendelevitch, Maciej Kilian, Dominik Lorenz, Yam Levi, Zion English, Vikram Voleti, Adam Letts, et~al.
\newblock Stable video diffusion: Scaling latent video diffusion models to large datasets.
\newblock \emph{arXiv preprint arXiv:2311.15127}, 2023.

\bibitem[Bochkovskii et~al.(2025)Bochkovskii, Delaunoy, Germain, Santos, Zhou, Richter, and Koltun]{Bochkovskii2024:arxiv}
Aleksei Bochkovskii, Ama\"{e}l Delaunoy, Hugo Germain, Marcel Santos, Yichao Zhou, Stephan~R. Richter, and Vladlen Koltun.
\newblock Depth pro: Sharp monocular metric depth in less than a second.
\newblock \emph{Proc. ICLR}, 2025.

\bibitem[Brooks et~al.(2023)Brooks, Holynski, and Efros]{brooks2022instructpix2pix}
Tim Brooks, Aleksander Holynski, and Alexei~A. Efros.
\newblock Instructpix2pix: Learning to follow image editing instructions.
\newblock In \emph{Proc. CVPR}, 2023.

\bibitem[Burgert et~al.(2025)Burgert, Xu, Xian, Pilarski, Clausen, He, Ma, Deng, Li, Mousavi, Ryoo, Debevec, and Yu]{burgert2025gowiththeflowmotioncontrollablevideodiffusion}
Ryan Burgert, Yuancheng Xu, Wenqi Xian, Oliver Pilarski, Pascal Clausen, Mingming He, Li~Ma, Yitong Deng, Lingxiao Li, Mohsen Mousavi, Michael Ryoo, Paul Debevec, and Ning Yu.
\newblock Go-with-the-flow: Motion-controllable video diffusion models using real-time warped noise.
\newblock In \emph{Proc. CVPR}, 2025.

\bibitem[Cha et~al.(2025)Cha, Lee, and Joo]{Cha_2025_CVPR}
Hyunsoo Cha, Inhee Lee, and Hanbyul Joo.
\newblock Perse: Personalized 3d generative avatars from a single portrait.
\newblock In \emph{Proc. CVPR}, 2025.

\bibitem[Cha et~al.(2026)Cha, Kim, and Joo]{cha2025durian}
Hyunsoo Cha, Byungjun Kim, and Hanbyul Joo.
\newblock Durian: Dual reference-guided portrait animation with attribute transfer.
\newblock \emph{Proc. ICLR}, 2026.

\bibitem[Chefer et~al.(2025)Chefer, Singer, Zohar, Kirstain, Polyak, Taigman, Wolf, and Sheynin]{chefer2025videojam}
Hila Chefer, Uriel Singer, Amit Zohar, Yuval Kirstain, Adam Polyak, Yaniv Taigman, Lior Wolf, and Shelly Sheynin.
\newblock Video{JAM}: Joint appearance-motion representations for enhanced motion generation in video models.
\newblock 2025.

\bibitem[Chen et~al.(2024)Chen, Laina, and Vedaldi]{chen2024training}
Minghao Chen, Iro Laina, and Andrea Vedaldi.
\newblock Training-free layout control with cross-attention guidance.
\newblock In \emph{Proc. WACV}, 2024.

\bibitem[Chen et~al.(2023)Chen, Wu, Xie, Wu, Li, Xia, Xiao, and Lin]{chen2023controlavideo}
Weifeng Chen, Jie Wu, Pan Xie, Hefeng Wu, Jiashi Li, Xin Xia, Xuefeng Xiao, and Liang Lin.
\newblock Control-a-video: Controllable text-to-video generation with diffusion models.
\newblock \emph{arXiv preprint arXiv:2305.13840}, 2023.

\bibitem[Deng et~al.(2024)Deng, Wang, Zhang, Tai, and Tang]{deng2023dragvideo}
Yufan Deng, Ruida Wang, Yuhao Zhang, Yu-Wing Tai, and Chi-Keung Tang.
\newblock Dragvideo: Interactive drag-style video editing.
\newblock \emph{Proc. ECCV}, 2024.

\bibitem[Du et~al.(2023)Du, Yang, Dai, Dai, Nachum, Tenenbaum, Schuurmans, and Abbeel]{du2023learning}
Yilun Du, Sherry Yang, Bo~Dai, Hanjun Dai, Ofir Nachum, Josh Tenenbaum, Dale Schuurmans, and Pieter Abbeel.
\newblock Learning universal policies via text-guided video generation.
\newblock \emph{NeurIPS}, 2023.

\bibitem[Esser et~al.(2023)Esser, Chiu, Atighehchian, Granskog, and Germanidis]{esser2023structure}
Patrick Esser, Johnathan Chiu, Parmida Atighehchian, Jonathan Granskog, and Anastasis Germanidis.
\newblock Structure and content-guided video synthesis with diffusion models.
\newblock In \emph{Proc. CVPR}, 2023.

\bibitem[Esser et~al.(2024)Esser, Kulal, Blattmann, Entezari, M{\"u}ller, Saini, Levi, Lorenz, Sauer, Boesel, et~al.]{esser2024scaling}
Patrick Esser, Sumith Kulal, Andreas Blattmann, Rahim Entezari, Jonas M{\"u}ller, Harry Saini, Yam Levi, Dominik Lorenz, Axel Sauer, Frederic Boesel, et~al.
\newblock Scaling rectified flow transformers for high-resolution image synthesis.
\newblock In \emph{Proc. ICML}, 2024.

\bibitem[{Fei, Zhengcong}(2024)]{CogvideoXInterpolation}
{Fei, Zhengcong}, 2024.
\newblock \url{https://github.com/feizc/CogvideX-Interpolation}.

\bibitem[Ghosh et~al.(2023)Ghosh, Dabral, Golyanik, Theobalt, and Slusallek]{ghosh2023imos}
Anindita Ghosh, Rishabh Dabral, Vladislav Golyanik, Christian Theobalt, and Philipp Slusallek.
\newblock Imos: Intent-driven full-body motion synthesis for human-object interactions.
\newblock In \emph{Proc. Eurographics}, 2023.

\bibitem[Grauman et~al.(2024)Grauman, Westbury, Torresani, Kitani, Malik, Afouras, Ashutosh, Baiyya, Bansal, Boote, et~al.]{grauman2024ego}
Kristen Grauman, Andrew Westbury, Lorenzo Torresani, Kris Kitani, Jitendra Malik, Triantafyllos Afouras, Kumar Ashutosh, Vijay Baiyya, Siddhant Bansal, Bikram Boote, et~al.
\newblock Ego-exo4d: Understanding skilled human activity from first-and third-person perspectives.
\newblock In \emph{Proc. CVPR}, 2024.

\bibitem[Gu et~al.(2025)Gu, Yan, Lu, Li, Dou, Si, Dong, Liu, Lin, Liu, et~al.]{gu2025diffusion}
Zekai Gu, Rui Yan, Jiahao Lu, Peng Li, Zhiyang Dou, Chenyang Si, Zhen Dong, Qifeng Liu, Cheng Lin, Ziwei Liu, et~al.
\newblock Diffusion as shader: 3d-aware video diffusion for versatile video generation control.
\newblock In \emph{Proc. ACM SIGGRAPH Asia}, 2025.

\bibitem[Guo et~al.(2024)Guo, Yang, Rao, Liang, Wang, Qiao, Agrawala, Lin, and Dai]{guo2023animatediff}
Yuwei Guo, Ceyuan Yang, Anyi Rao, Zhengyang Liang, Yaohui Wang, Yu~Qiao, Maneesh Agrawala, Dahua Lin, and Bo~Dai.
\newblock Animatediff: Animate your personalized text-to-image diffusion models without specific tuning.
\newblock \emph{Proc. ICLR}, 2024.

\bibitem[Ha \& Schmidhuber(2018)Ha and Schmidhuber]{ha2018world}
David Ha and J{\"u}rgen Schmidhuber.
\newblock World models.
\newblock In \emph{NeurIPS}, 2018.

\bibitem[HaCohen et~al.(2024)HaCohen, Chiprut, Brazowski, Shalem, Moshe, Richardson, Levin, Shiran, Zabari, Gordon, Panet, Weissbuch, Kulikov, Bitterman, Melumian, and Bibi]{HaCohen2024LTXVideo}
Yoav HaCohen, Nisan Chiprut, Benny Brazowski, Daniel Shalem, Dudu Moshe, Eitan Richardson, Eran Levin, Guy Shiran, Nir Zabari, Ori Gordon, Poriya Panet, Sapir Weissbuch, Victor Kulikov, Yaki Bitterman, Zeev Melumian, and Ofir Bibi.
\newblock Ltx-video: Realtime video latent diffusion.
\newblock \emph{arXiv preprint arXiv:2501.00103}, 2024.

\bibitem[Han \& Joo(2023)Han and Joo]{han2023chorus}
Sookwan Han and Hanbyul Joo.
\newblock Chorus: Learning canonicalized 3d human-object spatial relations from unbounded synthesized images.
\newblock In \emph{Proc. ICCV}, 2023.

\bibitem[Hassan et~al.(2019)Hassan, Choutas, Tzionas, and Black]{hassan2019resolving}
Mohamed Hassan, Vasileios Choutas, Dimitrios Tzionas, and Michael~J Black.
\newblock Resolving 3d human pose ambiguities with 3d scene constraints.
\newblock In \emph{Proc. CVPR}, 2019.

\bibitem[Hassan et~al.(2021{\natexlab{a}})Hassan, Ceylan, Villegas, Saito, Yang, Zhou, and Black]{hassan2021stochastic}
Mohamed Hassan, Duygu Ceylan, Ruben Villegas, Jun Saito, Jimei Yang, Yi~Zhou, and Michael~J Black.
\newblock Stochastic scene-aware motion prediction.
\newblock In \emph{Proc. ICCV}, 2021{\natexlab{a}}.

\bibitem[Hassan et~al.(2021{\natexlab{b}})Hassan, Ghosh, Tesch, Tzionas, and Black]{hassan2021populating}
Mohamed Hassan, Partha Ghosh, Joachim Tesch, Dimitrios Tzionas, and Michael~J Black.
\newblock Populating 3d scenes by learning human-scene interaction.
\newblock In \emph{Proc. CVPR}, 2021{\natexlab{b}}.

\bibitem[He et~al.(2025)He, Xu, Guo, Wetzstein, Dai, Li, and Yang]{he2024cameractrl}
Hao He, Yinghao Xu, Yuwei Guo, Gordon Wetzstein, Bo~Dai, Hongsheng Li, and Ceyuan Yang.
\newblock Cameractrl: Enabling camera control for text-to-video generation.
\newblock \emph{Proc. ICLR}, 2025.

\bibitem[Hertz et~al.(2022)Hertz, Mokady, Tenenbaum, Aberman, Pritch, and Cohen-Or]{hertz2022prompt}
Amir Hertz, Ron Mokady, Jay Tenenbaum, Kfir Aberman, Yael Pritch, and Daniel Cohen-Or.
\newblock Prompt-to-prompt image editing with cross attention control.
\newblock In \emph{Proc. ICLR}, 2022.

\bibitem[Ho \& Salimans(2022)Ho and Salimans]{ho2022classifier}
Jonathan Ho and Tim Salimans.
\newblock Classifier-free diffusion guidance.
\newblock In \emph{NeurIPS Workshop}, 2022.

\bibitem[Ho et~al.(2022)Ho, Salimans, Gritsenko, Chan, Norouzi, and Fleet]{ho2022video}
Jonathan Ho, Tim Salimans, Alexey Gritsenko, William Chan, Mohammad Norouzi, and David~J Fleet.
\newblock Video diffusion models.
\newblock \emph{NeurIPS}, 2022.

\bibitem[Hu et~al.(2022)Hu, Shen, Wallis, Allen-Zhu, Li, Wang, Wang, and Chen]{hu2021lora}
Edward~J Hu, Yelong Shen, Phillip Wallis, Zeyuan Allen-Zhu, Yuanzhi Li, Shean Wang, Lu~Wang, and Weizhu Chen.
\newblock Lora: Low-rank adaptation of large language models.
\newblock In \emph{Proc. ICLR}, 2022.

\bibitem[Huang et~al.(2022)Huang, Yi, H{\"o}schle, Safroshkin, Alexiadis, Polikovsky, Scharstein, and Black]{huang2022capturing}
Chun-Hao~P Huang, Hongwei Yi, Markus H{\"o}schle, Matvey Safroshkin, Tsvetelina Alexiadis, Senya Polikovsky, Daniel Scharstein, and Michael~J Black.
\newblock Capturing and inferring dense full-body human-scene contact.
\newblock In \emph{Proc. CVPR}, 2022.

\bibitem[Huang et~al.(2024)Huang, He, Yu, Zhang, Si, Jiang, Zhang, Wu, Jin, Chanpaisit, Wang, Chen, Wang, Lin, Qiao, and Liu]{huang2023vbench}
Ziqi Huang, Yinan He, Jiashuo Yu, Fan Zhang, Chenyang Si, Yuming Jiang, Yuanhan Zhang, Tianxing Wu, Qingyang Jin, Nattapol Chanpaisit, Yaohui Wang, Xinyuan Chen, Limin Wang, Dahua Lin, Yu~Qiao, and Ziwei Liu.
\newblock {VBench}: Comprehensive benchmark suite for video generative models.
\newblock In \emph{Proc. CVPR}, 2024.

\bibitem[Hurst et~al.(2024)Hurst, Lerer, Goucher, Perelman, Ramesh, Clark, Ostrow, Welihinda, Hayes, Radford, et~al.]{hurst2024gpt}
Aaron Hurst, Adam Lerer, Adam~P Goucher, Adam Perelman, Aditya Ramesh, Aidan Clark, AJ~Ostrow, Akila Welihinda, Alan Hayes, Alec Radford, et~al.
\newblock Gpt-4o system card.
\newblock \emph{arXiv preprint arXiv:2410.21276}, 2024.

\bibitem[Jain et~al.(2024)Jain, Nasery, Vineet, and Behl]{jain2024peekaboo}
Yash Jain, Anshul Nasery, Vibhav Vineet, and Harkirat Behl.
\newblock Peekaboo: Interactive video generation via masked-diffusion.
\newblock In \emph{Proc. CVPR}, 2024.

\bibitem[Jeong \& Ye(2024)Jeong and Ye]{jeong2023ground}
Hyeonho Jeong and Jong~Chul Ye.
\newblock Ground-a-video: Zero-shot grounded video editing using text-to-image diffusion models.
\newblock \emph{Proc. ICLR}, 2024.

\bibitem[Jeong et~al.(2025)Jeong, Huang, Ye, Mitra, and Ceylan]{jeong2024track4gen}
Hyeonho Jeong, Chun-Hao~Paul Huang, Jong~Chul Ye, Niloy Mitra, and Duygu Ceylan.
\newblock Track4gen: Teaching video diffusion models to track points improves video generation.
\newblock \emph{Proc. CVPR}, 2025.

\bibitem[Jiang et~al.(2023)Jiang, Liu, Cao, Cui, Zhang, Chen, Wang, Zhu, and Huang]{jiang2023full}
Nan Jiang, Tengyu Liu, Zhexuan Cao, Jieming Cui, Zhiyuan Zhang, Yixin Chen, He~Wang, Yixin Zhu, and Siyuan Huang.
\newblock Full-body articulated human-object interaction.
\newblock In \emph{Proc. ICCV}, 2023.

\bibitem[Jiang et~al.(2024{\natexlab{a}})Jiang, He, Wang, Li, Chen, Huang, and Zhu]{jiang2024autonomous}
Nan Jiang, Zimo He, Zi~Wang, Hongjie Li, Yixin Chen, Siyuan Huang, and Yixin Zhu.
\newblock Autonomous character-scene interaction synthesis from text instruction.
\newblock In \emph{Proc. ACM SIGGRAPH Asia}, 2024{\natexlab{a}}.

\bibitem[Jiang et~al.(2024{\natexlab{b}})Jiang, Zhang, Li, Ma, Wang, Chen, Liu, Zhu, and Huang]{jiang2024scaling}
Nan Jiang, Zhiyuan Zhang, Hongjie Li, Xiaoxuan Ma, Zan Wang, Yixin Chen, Tengyu Liu, Yixin Zhu, and Siyuan Huang.
\newblock Scaling up dynamic human-scene interaction modeling.
\newblock In \emph{Proc. CVPR}, 2024{\natexlab{b}}.

\bibitem[Karaev et~al.(2024)Karaev, Rocco, Graham, Neverova, Vedaldi, and Rupprecht]{karaev23cotracker}
Nikita Karaev, Ignacio Rocco, Benjamin Graham, Natalia Neverova, Andrea Vedaldi, and Christian Rupprecht.
\newblock Cotracker: It is better to track together.
\newblock In \emph{Proc. ECCV}, 2024.

\bibitem[Khachatryan et~al.(2023)Khachatryan, Movsisyan, Tadevosyan, Henschel, Wang, Navasardyan, and Shi]{khachatryan2023text2video}
Levon Khachatryan, Andranik Movsisyan, Vahram Tadevosyan, Roberto Henschel, Zhangyang Wang, Shant Navasardyan, and Humphrey Shi.
\newblock Text2video-zero: Text-to-image diffusion models are zero-shot video generators.
\newblock In \emph{Proc. CVPR}, 2023.

\bibitem[Kim et~al.(2025{\natexlab{a}})Kim, Kim, Lee, and Joo]{kim2025dwm}
Byungjun Kim, Taeksoo Kim, Junyoung Lee, and Hanbyul Joo.
\newblock Dexterous world models.
\newblock \emph{arXiv preprint arXiv:2512.17907}, 2025{\natexlab{a}}.

\bibitem[Kim et~al.(2024{\natexlab{a}})Kim, Han, Kwon, and Joo]{ComA}
Hyeonwoo Kim, Sookwan Han, Patrick Kwon, and Hanbyul Joo.
\newblock Beyond the contact: Discovering comprehensive affordance for 3d objects from pre-trained 2d diffusion models.
\newblock In \emph{Proc. ECCV}, 2024{\natexlab{a}}.

\bibitem[Kim et~al.(2025{\natexlab{b}})Kim, Beak, and Joo]{kim2025david}
Hyeonwoo Kim, Sangwon Beak, and Hanbyul Joo.
\newblock David: Modeling dynamic affordance of 3d objects using pre-trained video diffusion models.
\newblock In \emph{Proc. ICCV}, 2025{\natexlab{b}}.

\bibitem[Kim et~al.(2025{\natexlab{c}})Kim, Kim, Na, and Joo]{kim2024parahome}
Jeonghwan Kim, Jisoo Kim, Jeonghyeon Na, and Hanbyul Joo.
\newblock Parahome: Parameterizing everyday home activities towards 3d generative modeling of human-object interactions.
\newblock In \emph{Proc. CVPR}, 2025{\natexlab{c}}.

\bibitem[Kim et~al.(2023{\natexlab{a}})Kim, Saito, and Joo]{kim2023ncho}
Taeksoo Kim, Shunsuke Saito, and Hanbyul Joo.
\newblock Ncho: Unsupervised learning for neural 3d composition of humans and objects.
\newblock In \emph{Proc. ICCV}, 2023{\natexlab{a}}.

\bibitem[Kim et~al.(2024{\natexlab{b}})Kim, Kim, Saito, and Joo]{kim2024gala}
Taeksoo Kim, Byungjun Kim, Shunsuke Saito, and Hanbyul Joo.
\newblock Gala: Generating animatable layered assets from a single scan.
\newblock In \emph{Proc. CVPR}, 2024{\natexlab{b}}.

\bibitem[Kim et~al.(2014)Kim, Chaudhuri, Guibas, and Funkhouser]{kim2014shape2pose}
Vladimir~G Kim, Siddhartha Chaudhuri, Leonidas Guibas, and Thomas Funkhouser.
\newblock Shape2pose: Human-centric shape analysis.
\newblock \emph{ACM TOG}, 2014.

\bibitem[Kim et~al.(2023{\natexlab{b}})Kim, Lee, Kim, Ha, and Zhu]{kim2023dense}
Yunji Kim, Jiyoung Lee, Jin-Hwa Kim, Jung-Woo Ha, and Jun-Yan Zhu.
\newblock Dense text-to-image generation with attention modulation.
\newblock In \emph{Proc. ICCV}, 2023{\natexlab{b}}.

\bibitem[Kirillov et~al.(2023)Kirillov, Mintun, Ravi, Mao, Rolland, Gustafson, Xiao, Whitehead, Berg, Lo, Doll{\'a}r, and Girshick]{kirillov2023sam}
Alexander Kirillov, Eric Mintun, Nikhila Ravi, Hanzi Mao, Chloe Rolland, Laura Gustafson, Tete Xiao, Spencer Whitehead, Alexander~C. Berg, Wan-Yen Lo, Piotr Doll{\'a}r, and Ross Girshick.
\newblock Segment anything.
\newblock In \emph{Proc. ICCV}, 2023.

\bibitem[Kong et~al.(2024)Kong, Tian, Zhang, Min, Dai, Zhou, Xiong, Li, Wu, Zhang, Wu, Lin, Wang, Wang, Li, Huang, Yang, Tan, Wang, Song, Bai, Wu, Xue, Wang, Yuan, Wang, Liu, Li, Li, Wang, Yu, Deng, Li, Long, Chen, Cui, Peng, Yu, He, Xu, Zhou, Xu, Tao, Lu, Liu, Zhou, Wang, Yang, Wang, Liu, Jiang, and Zhong]{kong2024hunyuanvideo}
Weijie Kong, Qi~Tian, Zijian Zhang, Rox Min, Zuozhuo Dai, Jin Zhou, Jiangfeng Xiong, Xin Li, Bo~Wu, Jianwei Zhang, Kathrina Wu, Qin Lin, Aladdin Wang, Andong Wang, Changlin Li, Duojun Huang, Fang Yang, Hao Tan, Hongmei Wang, Jacob Song, Jiawang Bai, Jianbing Wu, Jinbao Xue, Joey Wang, Junkun Yuan, Kai Wang, Mengyang Liu, Pengyu Li, Shuai Li, Weiyan Wang, Wenqing Yu, Xinchi Deng, Yang Li, Yanxin Long, Yi~Chen, Yutao Cui, Yuanbo Peng, Zhentao Yu, Zhiyu He, Zhiyong Xu, Zixiang Zhou, Zunnan Xu, Yangyu Tao, Qinglin Lu, Songtao Liu, Dax Zhou, Hongfa Wang, Yong Yang, Di~Wang, Yuhong Liu, Jie Jiang, and Caesar Zhong.
\newblock Hunyuanvideo: A systematic framework for large video generative models.
\newblock \emph{arXiv preprint arXiv:2412.03603}, 2024.

\bibitem[Lee \& Joo(2023)Lee and Joo]{lee2023locomotion}
Jiye Lee and Hanbyul Joo.
\newblock Locomotion-action-manipulation: Synthesizing human-scene interactions in complex 3d environments.
\newblock In \emph{Proc. CVPR}, 2023.

\bibitem[Li et~al.(2023{\natexlab{a}})Li, Clegg, Mottaghi, Wu, Puig, and Liu]{li2023controllable}
Jiaman Li, Alexander Clegg, Roozbeh Mottaghi, Jiajun Wu, Xavier Puig, and C~Karen Liu.
\newblock Controllable human-object interaction synthesis.
\newblock In \emph{Proc. ECCV}, 2023{\natexlab{a}}.

\bibitem[Li et~al.(2023{\natexlab{b}})Li, Wu, and Liu]{li2023object}
Jiaman Li, Jiajun Wu, and C~Karen Liu.
\newblock Object motion guided human motion synthesis.
\newblock \emph{ACM TOG}, 2023{\natexlab{b}}.

\bibitem[Li \& Dai(2024)Li and Dai]{li2024genzi}
Lei Li and Angela Dai.
\newblock Genzi: Zero-shot 3d human-scene interaction generation.
\newblock In \emph{Proc. CVPR}, 2024.

\bibitem[Li et~al.(2019)Li, Liu, Kim, Wang, Yang, and Kautz]{li2019putting}
Xueting Li, Sifei Liu, Kihwan Kim, Xiaolong Wang, Ming-Hsuan Yang, and Jan Kautz.
\newblock Putting humans in a scene: Learning affordance in 3d indoor environments.
\newblock In \emph{Proc. CVPR}, 2019.

\bibitem[Liu et~al.(2024)Liu, Zhang, Li, Lin, and Jia]{liu2023videop2p}
Shaoteng Liu, Yuechen Zhang, Wenbo Li, Zhe Lin, and Jiaya Jia.
\newblock Video-p2p: Video editing with cross-attention control.
\newblock \emph{Proc. CVPR}, 2024.

\bibitem[Lu et~al.(2022)Lu, Zhou, Bao, Chen, Li, and Zhu]{lu2022dpm}
Cheng Lu, Yuhao Zhou, Fan Bao, Jianfei Chen, Chongxuan Li, and Jun Zhu.
\newblock Dpm-solver: A fast ode solver for diffusion probabilistic model sampling in around 10 steps.
\newblock In \emph{NeurIPS}, 2022.

\bibitem[Ma et~al.(2023)Ma, Lewis, Lahiri, Leung, and Kleijn]{ma2023directed}
Wan-Duo~Kurt Ma, J.~P. Lewis, Avisek Lahiri, Thomas Leung, and W.~Bastiaan Kleijn.
\newblock Directed diffusion: Direct control of object placement through attention guidance, 2023.

\bibitem[Makoviychuk et~al.(2021)Makoviychuk, Wawrzyniak, Guo, Lu, Storey, Macklin, Hoeller, Rudin, Allshire, Handa, et~al.]{makoviychuk2021isaac}
Viktor Makoviychuk, Lukasz Wawrzyniak, Yunrong Guo, Michelle Lu, Kier Storey, Miles Macklin, David Hoeller, Nikita Rudin, Arthur Allshire, Ankur Handa, et~al.
\newblock Isaac gym: High performance gpu-based physics simulation for robot learning.
\newblock \emph{arXiv preprint arXiv:2108.10470}, 2021.

\bibitem[Monszpart et~al.(2019)Monszpart, Guerrero, Ceylan, Yumer, and Mitra]{monszpart2019imapper}
Aron Monszpart, Paul Guerrero, Duygu Ceylan, Ersin Yumer, and Niloy~J Mitra.
\newblock imapper: interaction-guided scene mapping from monocular videos.
\newblock \emph{ACM TOG}, 2019.

\bibitem[Narasimhaswamy et~al.(2020)Narasimhaswamy, Nguyen, and Hoai]{contacthands_2020}
Supreeth Narasimhaswamy, Trung Nguyen, and Minh Hoai.
\newblock Detecting hands and recognizing physical contact in the wild.
\newblock In \emph{NeurIPS}, 2020.

\bibitem[Ni et~al.(2024)Ni, Hao, Wu, Kou, Liu, Zheng, Wang, and Zhuang]{ni2024generate}
Fei Ni, Jianye Hao, Shiguang Wu, Longxin Kou, Jiashun Liu, Yan Zheng, Bin Wang, and Yuzheng Zhuang.
\newblock Generate subgoal images before act: Unlocking the chain-of-thought reasoning in diffusion model for robot manipulation with multimodal prompts.
\newblock In \emph{Proc. CVPR}, 2024.

\bibitem[Ni et~al.(2023)Ni, Shi, Li, Huang, and Min]{ni2023conditional}
Haomiao Ni, Changhao Shi, Kai Li, Sharon~X Huang, and Martin~Renqiang Min.
\newblock Conditional image-to-video generation with latent flow diffusion models.
\newblock In \emph{Proc. CVPR}, 2023.

\bibitem[Pan et~al.(2023)Pan, Tewari, Leimk{\"u}hler, Liu, Meka, and Theobalt]{pan2023drag}
Xingang Pan, Ayush Tewari, Thomas Leimk{\"u}hler, Lingjie Liu, Abhimitra Meka, and Christian Theobalt.
\newblock Drag your gan: Interactive point-based manipulation on the generative image manifold.
\newblock In \emph{Proc. ACM SIGGRAPH}, 2023.

\bibitem[Park et~al.(2026)Park, Bharadhwaj, and Tulsiani]{park2025demodiffusiononeshothumanimitation}
Sungjae Park, Homanga Bharadhwaj, and Shubham Tulsiani.
\newblock Demodiffusion: One-shot human imitation using pre-trained diffusion policy.
\newblock In \emph{Proc. ICRA}, 2026.

\bibitem[Peebles \& Xie(2023)Peebles and Xie]{peebles2023scalable}
William Peebles and Saining Xie.
\newblock Scalable diffusion models with transformers.
\newblock In \emph{Proc. ICCV}, 2023.

\bibitem[Ren et~al.(2024)Ren, Liu, Zeng, Lin, Li, Cao, Chen, Huang, Chen, Yan, Zeng, Zhang, Li, Yang, Li, Jiang, and Zhang]{ren2024grounded}
Tianhe Ren, Shilong Liu, Ailing Zeng, Jing Lin, Kunchang Li, He~Cao, Jiayu Chen, Xinyu Huang, Yukang Chen, Feng Yan, Zhaoyang Zeng, Hao Zhang, Feng Li, Jie Yang, Hongyang Li, Qing Jiang, and Lei Zhang.
\newblock Grounded sam: Assembling open-world models for diverse visual tasks, 2024.

\bibitem[Rombach et~al.(2022{\natexlab{a}})Rombach, Blattmann, Lorenz, Esser, and Ommer]{rombach2022high}
Robin Rombach, Andreas Blattmann, Dominik Lorenz, Patrick Esser, and Bj{\"o}rn Ommer.
\newblock High-resolution image synthesis with latent diffusion models.
\newblock In \emph{Proc. CVPR}, 2022{\natexlab{a}}.

\bibitem[Rombach et~al.(2022{\natexlab{b}})Rombach, Blattmann, Lorenz, Esser, and Ommer]{rombach2022latent}
Robin Rombach, Andreas Blattmann, Dominik Lorenz, Patrick Esser, and Bj{\"o}rn Ommer.
\newblock High-resolution image synthesis with latent diffusion models.
\newblock In \emph{Proc. CVPR}, 2022{\natexlab{b}}.

\bibitem[Savva et~al.(2016)Savva, Chang, Hanrahan, Fisher, and Nie{\ss}ner]{savva2016pigraphs}
Manolis Savva, Angel~X Chang, Pat Hanrahan, Matthew Fisher, and Matthias Nie{\ss}ner.
\newblock Pigraphs: learning interaction snapshots from observations.
\newblock \emph{ACM TOG}, 2016.

\bibitem[Shen et~al.(2024)Shen, Pi, Xia, Cen, Peng, Hu, Bao, Hu, and Zhou]{shen2024gvhmr}
Zehong Shen, Huaijin Pi, Yan Xia, Zhi Cen, Sida Peng, Zechen Hu, Hujun Bao, Ruizhen Hu, and Xiaowei Zhou.
\newblock World-grounded human motion recovery via gravity-view coordinates.
\newblock In \emph{Proc. ACM SIGGRAPH Asia}, 2024.

\bibitem[Shi et~al.(2024{\natexlab{a}})Shi, Huang, Wang, Bian, Li, Zhang, Zhang, Cheung, See, Qin, et~al.]{shi2024motion}
Xiaoyu Shi, Zhaoyang Huang, Fu-Yun Wang, Weikang Bian, Dasong Li, Yi~Zhang, Manyuan Zhang, Ka~Chun Cheung, Simon See, Hongwei Qin, et~al.
\newblock Motion-i2v: Consistent and controllable image-to-video generation with explicit motion modeling.
\newblock In \emph{Proc. ACM SIGGRAPH}, 2024{\natexlab{a}}.

\bibitem[Shi et~al.(2024{\natexlab{b}})Shi, Xue, Liew, Pan, Yan, Zhang, Tan, and Bai]{shi2024dragdiffusion}
Yujun Shi, Chuhui Xue, Jun~Hao Liew, Jiachun Pan, Hanshu Yan, Wenqing Zhang, Vincent~YF Tan, and Song Bai.
\newblock Dragdiffusion: Harnessing diffusion models for interactive point-based image editing.
\newblock In \emph{Proc. CVPR}, 2024{\natexlab{b}}.

\bibitem[Shin et~al.(2024)Shin, Choi, and Park]{shin2024instantdrag}
Joonghyuk Shin, Daehyeon Choi, and Jaesik Park.
\newblock Instantdrag: Improving interactivity in drag-based image editing.
\newblock In \emph{Proc. ACM SIGGRAPH Asia}, 2024.

\bibitem[Starke et~al.(2019)Starke, Zhang, Komura, and Saito]{starke2019neural}
Sebastian Starke, He~Zhang, Taku Komura, and Jun Saito.
\newblock Neural state machine for character-scene interactions.
\newblock \emph{ACM TOG}, 2019.

\bibitem[Taheri et~al.(2020)Taheri, Ghorbani, Black, and Tzionas]{taheri2020grab}
Omid Taheri, Nima Ghorbani, Michael~J Black, and Dimitrios Tzionas.
\newblock Grab: A dataset of whole-body human grasping of objects.
\newblock In \emph{Proc. ECCV}, 2020.

\bibitem[Taheri et~al.(2022)Taheri, Choutas, Black, and Tzionas]{taheri2022goal}
Omid Taheri, Vasileios Choutas, Michael~J Black, and Dimitrios Tzionas.
\newblock Goal: Generating 4d whole-body motion for hand-object grasping.
\newblock In \emph{Proc. CVPR}, 2022.

\bibitem[Teng et~al.(2023)Teng, Xie, Wu, Han, Li, and Liu]{teng2023drag}
Yao Teng, Enze Xie, Yue Wu, Haoyu Han, Zhenguo Li, and Xihui Liu.
\newblock Drag-a-video: Non-rigid video editing with point-based interaction.
\newblock \emph{arXiv preprint arXiv:2312.02936}, 2023.

\bibitem[Wang et~al.(2021{\natexlab{a}})Wang, Xu, Xu, Liu, and Wang]{wang2021synthesizing}
Jiashun Wang, Huazhe Xu, Jingwei Xu, Sifei Liu, and Xiaolong Wang.
\newblock Synthesizing long-term 3d human motion and interaction in 3d scenes.
\newblock In \emph{Proc. CVPR}, 2021{\natexlab{a}}.

\bibitem[Wang et~al.(2021{\natexlab{b}})Wang, Yan, Dai, and Lin]{wang2021scene}
Jingbo Wang, Sijie Yan, Bo~Dai, and Dahua Lin.
\newblock Scene-aware generative network for human motion synthesis.
\newblock In \emph{Proc. CVPR}, 2021{\natexlab{b}}.

\bibitem[Wang et~al.(2023{\natexlab{a}})Wang, Xie, Liu, Chen, Cao, Wang, and Shen]{vid2vid-zero}
Wen Wang, kangyang Xie, Zide Liu, Hao Chen, Yue Cao, Xinlong Wang, and Chunhua Shen.
\newblock Zero-shot video editing using off-the-shelf image diffusion models.
\newblock \emph{arXiv preprint arXiv:2303.17599}, 2023{\natexlab{a}}.

\bibitem[Wang et~al.(2023{\natexlab{b}})Wang, Lin, Zeng, Luo, Zhang, and Zhang]{wang2023physhoi}
Yinhuai Wang, Jing Lin, Ailing Zeng, Zhengyi Luo, Jian Zhang, and Lei Zhang.
\newblock Physhoi: Physics-based imitation of dynamic human-object interaction.
\newblock \emph{arXiv preprint arXiv:2312.04393}, 2023{\natexlab{b}}.

\bibitem[Wang et~al.(2024)Wang, Yuan, Wang, Li, Chen, Xia, Luo, and Shan]{wang2024motionctrl}
Zhouxia Wang, Ziyang Yuan, Xintao Wang, Yaowei Li, Tianshui Chen, Menghan Xia, Ping Luo, and Ying Shan.
\newblock Motionctrl: A unified and flexible motion controller for video generation.
\newblock In \emph{Proc. ACM SIGGRAPH}, 2024.

\bibitem[Wen et~al.(2024)Wen, Yang, Kautz, and Birchfield]{wen2024foundationpose}
Bowen Wen, Wei Yang, Jan Kautz, and Stan Birchfield.
\newblock Foundationpose: Unified 6d pose estimation and tracking of novel objects.
\newblock In \emph{Proc. CVPR}, 2024.

\bibitem[Wu et~al.(2024{\natexlab{a}})Wu, Li, Zeng, Zhang, Zhou, Li, Tong, and Chen]{wu2024motionbooth}
Jianzong Wu, Xiangtai Li, Yanhong Zeng, Jiangning Zhang, Qianyu Zhou, Yining Li, Yunhai Tong, and Kai Chen.
\newblock Motionbooth: Motion-aware customized text-to-video generation.
\newblock \emph{NeurIPS}, 2024{\natexlab{a}}.

\bibitem[Wu et~al.(2024{\natexlab{b}})Wu, Li, Gu, Zhao, He, Zhang, Shou, Li, Gao, and Zhang]{wu2024draganything}
Weijia Wu, Zhuang Li, Yuchao Gu, Rui Zhao, Yefei He, David~Junhao Zhang, Mike~Zheng Shou, Yan Li, Tingting Gao, and Di~Zhang.
\newblock Draganything: Motion control for anything using entity representation.
\newblock In \emph{Proc. ECCV}, 2024{\natexlab{b}}.

\bibitem[Xiao et~al.(2024)Xiao, Wang, Zhang, Xue, Peng, Shen, and Zhou]{SpatialTracker}
Yuxi Xiao, Qianqian Wang, Shangzhan Zhang, Nan Xue, Sida Peng, Yujun Shen, and Xiaowei Zhou.
\newblock Spatialtracker: Tracking any 2d pixels in 3d space.
\newblock In \emph{Proc. CVPR}, 2024.

\bibitem[Xie et~al.(2023)Xie, Li, Huang, Liu, Zhang, Zheng, and Shou]{xie2023boxdiff}
Jinheng Xie, Yuexiang Li, Yawen Huang, Haozhe Liu, Wentian Zhang, Yefeng Zheng, and Mike~Zheng Shou.
\newblock Boxdiff: Text-to-image synthesis with training-free box-constrained diffusion.
\newblock In \emph{Proc. ICCV}, 2023.

\bibitem[Xie et~al.(2024)Xie, Bhatnagar, Lenssen, and Pons-Moll]{xie2024template}
Xianghui Xie, Bharat~Lal Bhatnagar, Jan~Eric Lenssen, and Gerard Pons-Moll.
\newblock Template free reconstruction of human-object interaction with procedural interaction generation.
\newblock In \emph{Proc. CVPR}, 2024.

\bibitem[Xing et~al.(2024)Xing, Xia, Zhang, Chen, Yu, Liu, Wang, Wong, and Shan]{xing2023dynamicrafter}
Jinbo Xing, Menghan Xia, Yong Zhang, Haoxin Chen, Wangbo Yu, Hanyuan Liu, Xintao Wang, Tien-Tsin Wong, and Ying Shan.
\newblock Dynamicrafter: Animating open-domain images with video diffusion priors.
\newblock \emph{Proc. ECCV}, 2024.

\bibitem[Xu et~al.(2023)Xu, Li, Wang, and Gui]{xu2023interdiff}
Sirui Xu, Zhengyuan Li, Yu-Xiong Wang, and Liang-Yan Gui.
\newblock Interdiff: Generating 3d human-object interactions with physics-informed diffusion.
\newblock In \emph{Proc. CVPR}, 2023.

\bibitem[Xu et~al.(2024)Xu, Wang, Wang, and Gui]{xu2024interdreamer}
Sirui Xu, Ziyin Wang, Yu-Xiong Wang, and Liang-Yan Gui.
\newblock Interdreamer: Zero-shot text to 3d dynamic human-object interaction.
\newblock In \emph{NeurIPS}, 2024.

\bibitem[Xu et~al.(2025)Xu, Ling, Wang, and Gui]{xu2025intermimic}
Sirui Xu, Hung~Yu Ling, Yu-Xiong Wang, and Liangyan Gui.
\newblock Intermimic: Towards universal whole-body control for physics-based human-object interactions.
\newblock In \emph{Proc. CVPR}, 2025.

\bibitem[Yan et~al.(2023)Yan, Brown, Abbeel, Girdhar, and Azadi]{yan2023motion}
Wilson Yan, Andrew Brown, Pieter Abbeel, Rohit Girdhar, and Samaneh Azadi.
\newblock Motion-conditioned image animation for video editing.
\newblock \emph{arXiv preprint arXiv:2311.18827}, 2023.

\bibitem[Yang et~al.(2025{\natexlab{a}})Yang, Liu, Guo, Dong, Zhang, Zhang, Wang, Zhou, Xie, Wang, Ouyang, Lin, Cominelli, Cai, Zhang, Zhang, Hong, Widmer, Gringoli, Yang, Li, and Liu]{yang2025egolifeegocentriclifeassistant}
Jingkang Yang, Shuai Liu, Hongming Guo, Yuhao Dong, Xiamengwei Zhang, Sicheng Zhang, Pengyun Wang, Zitang Zhou, Binzhu Xie, Ziyue Wang, Bei Ouyang, Zhengyu Lin, Marco Cominelli, Zhongang Cai, Yuanhan Zhang, Peiyuan Zhang, Fangzhou Hong, Joerg Widmer, Francesco Gringoli, Lei Yang, Bo~Li, and Ziwei Liu.
\newblock Egolife: Towards egocentric life assistant.
\newblock In \emph{Proc. CVPR}, 2025{\natexlab{a}}.

\bibitem[Yang et~al.(2024)Yang, Hou, Huang, Ma, Wan, Zhang, Chen, and Liao]{yang2024direct}
Shiyuan Yang, Liang Hou, Haibin Huang, Chongyang Ma, Pengfei Wan, Di~Zhang, Xiaodong Chen, and Jing Liao.
\newblock Direct-a-video: Customized video generation with user-directed camera movement and object motion.
\newblock In \emph{ACM TOG}, 2024.

\bibitem[Yang et~al.(2025{\natexlab{b}})Yang, Teng, Zheng, Ding, Huang, Xu, Yang, Hong, Zhang, Feng, et~al.]{yang2024cogvideox}
Zhuoyi Yang, Jiayan Teng, Wendi Zheng, Ming Ding, Shiyu Huang, Jiazheng Xu, Yuanming Yang, Wenyi Hong, Xiaohan Zhang, Guanyu Feng, et~al.
\newblock Cogvideox: Text-to-video diffusion models with an expert transformer.
\newblock In \emph{Proc. ICLR}, 2025{\natexlab{b}}.

\bibitem[Ye et~al.(2023)Ye, Zhang, Liu, Han, and Yang]{ye2023ip}
Hu~Ye, Jun Zhang, Sibo Liu, Xiao Han, and Wei Yang.
\newblock Ip-adapter: Text compatible image prompt adapter for text-to-image diffusion models.
\newblock \emph{arXiv preprint arXiv:2308.06721}, 2023.

\bibitem[Yi et~al.(2024)Yi, Thies, Black, Peng, and Rempe]{yi2024generating}
Hongwei Yi, Justus Thies, Michael~J Black, Xue~Bin Peng, and Davis Rempe.
\newblock Generating human interaction motions in scenes with text control.
\newblock In \emph{Proc. ECCV}, 2024.

\bibitem[Yin et~al.(2023)Yin, Wu, Liang, Shi, Li, Ming, and Duan]{yin2023dragnuwa}
Shengming Yin, Chenfei Wu, Jian Liang, Jie Shi, Houqiang Li, Gong Ming, and Nan Duan.
\newblock Dragnuwa: Fine-grained control in video generation by integrating text, image, and trajectory.
\newblock \emph{arXiv preprint arXiv:2308.08089}, 2023.

\bibitem[Zhang et~al.(2023{\natexlab{a}})Zhang, Wu, and Dong]{zhang2023genpose}
Jiyao Zhang, Mingdong Wu, and Hao Dong.
\newblock Genpose: Generative category-level object pose estimation via diffusion models.
\newblock \emph{NeurIPS}, 2023{\natexlab{a}}.

\bibitem[Zhang et~al.(2023{\natexlab{b}})Zhang, Rao, and Agrawala]{zhang2023controlnet}
Lvmin Zhang, Anyi Rao, and Maneesh Agrawala.
\newblock Adding conditional control to text-to-image diffusion models.
\newblock In \emph{Proc. ICCV}, 2023{\natexlab{b}}.

\bibitem[Zhang et~al.(2021)Zhang, Ye, Shiratori, and Komura]{zhang2021manipnet}
Menghe Zhang, Yuting Ye, Takaaki Shiratori, and Taku Komura.
\newblock Manipnet: neural manipulation synthesis with a hand-object spatial representation.
\newblock \emph{ACM TOG}, 2021.

\bibitem[Zhang et~al.(2020{\natexlab{a}})Zhang, Zhang, Ma, Black, and Tang]{zhang2020place}
Siwei Zhang, Yan Zhang, Qianli Ma, Michael~J Black, and Siyu Tang.
\newblock Place: Proximity learning of articulation and contact in 3d environments.
\newblock In \emph{Proc. 3{DV}}, 2020{\natexlab{a}}.

\bibitem[Zhang et~al.(2022)Zhang, Bhatnagar, Starke, Guzov, and Pons-Moll]{zhang2022couch}
Xiaohan Zhang, Bharat~Lal Bhatnagar, Sebastian Starke, Vladimir Guzov, and Gerard Pons-Moll.
\newblock Couch: Towards controllable human-chair interactions.
\newblock In \emph{Proc. ECCV}, 2022.

\bibitem[Zhang et~al.(2024)Zhang, Wei, Jiang, Zhang, Zuo, and Tian]{zhang2023controlvideo}
Yabo Zhang, Yuxiang Wei, Dongsheng Jiang, Xiaopeng Zhang, Wangmeng Zuo, and Qi~Tian.
\newblock Controlvideo: Training-free controllable text-to-video generation.
\newblock \emph{Proc. ICLR}, 2024.

\bibitem[Zhang et~al.(2020{\natexlab{b}})Zhang, Hassan, Neumann, Black, and Tang]{zhang2020generating}
Yan Zhang, Mohamed Hassan, Heiko Neumann, Michael~J Black, and Siyu Tang.
\newblock Generating 3d people in scenes without people.
\newblock In \emph{Proc. CVPR}, 2020{\natexlab{b}}.

\bibitem[Zhao et~al.(2022)Zhao, Wang, Zhang, Beeler, and Tang]{Zhao:ECCV:2022}
Kaifeng Zhao, Shaofei Wang, Yan Zhang, Thabo Beeler, and Siyu Tang.
\newblock Compositional human-scene interaction synthesis with semantic control.
\newblock In \emph{Proc. ECCV}, 2022.

\end{thebibliography}
\end{document}